\documentclass{article}

\PassOptionsToPackage{numbers}{natbib}
\usepackage[eandd,preprint]{neurips_2026}

\usepackage[utf8]{inputenc} % allow utf-8 input
\usepackage[T1]{fontenc}    % use 8-bit T1 fonts
\usepackage{hyperref}       % hyperlinks
\usepackage{url}            % simple URL typesetting
\usepackage{booktabs}       % professional-quality tables
\usepackage{amsfonts}       % blackboard math symbols
\usepackage{nicefrac}       % compact symbols for 1/2, etc.
\usepackage{microtype}      % microtypography
\usepackage{xcolor}         % colors
\usepackage{amsmath}
\usepackage{float}
\usepackage{graphicx}
\usepackage[font=footnotesize]{subcaption}
\usepackage{placeins}
\usepackage{multirow}
\usepackage[font=footnotesize]{caption}
\title{ContextShift: A Controlled Benchmark for Context Dependence in Object Detection}

\author{
    Dan Zlotnikov \quad Alex Lazarovich \quad Ohad Ben-Shahar \\
    Stein Faculty of Computer and Information Science \\
    Ben-Gurion University of the Negev, Israel \\
    {\tt\small \{danzlot, alexlaz\}@post.bgu.ac.il, ben-shahar@cs.bgu.ac.il}
}

\begin{document}

\maketitle
\begin{abstract}
Modern object detectors achieve strong performance on standard benchmarks, yet their robustness to contextual variation remains insufficiently investigated. Prior evaluations largely rely on aggregate metrics such as AP on uncontrolled distribution shifts, which conflate multiple factors and can obscure how performance degrades under context change. We introduce \textsc{ContextShift}, a controlled benchmark for evaluating context dependence by systematically manipulating object--context relationships while preserving object appearance. Built on COCO 2017, it isolates context as an independent variable both implicitly, through geometric transformations and explicitly via synthetic and natural background substitutions - with a compatibility axis based on normalized pointwise mutual information (NPMI) for continuous evaluation.
Across diverse detector architectures, we observe a consistent degradation pattern: false negatives increase up to 227\% and prediction volume decreases down to -44\%, while false positives remain relatively stable or decline. This observed suppression behavior is not revealed by aggregate metrics such as AP, which can mask substantial recall loss and changes in prediction dynamics. Further analysis suggests that degradation is not primarily explained by reduced confidence, but is instead consistent with a reduced formation of valid detection candidates. Moreover, we find that performance along the statistical compatibility axis is non-monotonic, peaking at intermediate NPMI and degrading toward both extremes, indicating that statistical co-occurrence does not correlate linearly with effective visual context. 
Finally, we show that a training methodology based on context-aware augmentations rather than the dataset alone, which yields overall better models: every augmented variant outperforms the dataset-only baseline on both original (unmanipulated) test images as well as manipulated images, indicating that data-oriented augmentation can partially recover performance lost to prediction-suppression failures by exposing the model to object--context decoupling during training.

\end{abstract}

\section{Introduction}
Modern object detectors achieve strong performance on large-scale benchmarks such as COCO~\cite{lin2014coco}, driven by advances in training methodology, dataset scale, and optimization. However, visual inputs inherently entangle objects with their surrounding context — e.g. background, co-occurring objects or illumination — allowing detectors to exploit contextual regularities that are predictive in training data but not causally tied to object identity. This can lead to brittle behavior under distribution shift~\cite{koh2021wilds,fabbrizzi2022survey,yao2022wildtime}, an important failure mode in various use-cases, e.g. autonomous vehicles~\cite{bojarski2016end} and detection-based security and surveillance systems~\cite{sultani2018real, viola2004robust}. Existing robustness benchmarks rely on natural, uncontrolled variation~\cite{mao2023cocoo} or image-level corruptions~\cite{michaelis2019benchmarking}; while valuable, they confound multiple factors and do not provide a controlled axis of object--background compatibility, making it difficult to isolate and quantify context dependence.

In this work, we study context dependence by \emph{treating context as an independent variable}. Starting from COCO 2017~\cite{lin2014coco}, we introduce two manipulation families: \emph{implicit} context manipulation based on transformations applied to objects within the original context, and \emph{explicit} context manipulations that alter the surrounding context through synthetic and natural background substitution. 
Importantly, the core of the evaluation setting for natural background substitution is driven by a concept of \textit{Compatibility}. 
For natural backgrounds, we define an object--background compatibility framework based on normalized pointwise mutual information (NPMI)~\cite{bouma2009normalized}, enabling continuous evaluation across graded context variation. This axis captures statistical object--background association rather than perceptual realism, allowing controlled and objective study of detector behavior as compatibility varies.
Across diverse detector architectures, we observe a consistent prediction suppression pattern (false negatives increase, total predictions decrease, false positives stay relatively stable or decrease). 

Moreover, careful examination of detector performance along the NPMI-based compatibility axis offers possible insight into overfit-like behavior across all models. And, showing that it is not exclusive to manipulated images, but is apparent also when tested on the non-manipulated original image dataset -- indicates that it is not an artifact of the manipulation pipeline, but rather a phenomenon that can be attributed to object-context coupling. To explain this behavior, we analyze prediction dynamics. We reveal that missed detections are not primarily low-confidence predictions, but instead indicate reduced prediction candidate formation. We find that this suppression mechanism is consistent across manipulations and object-background composition settings, and present a simple mitigation technique which demonstrates that simple pre-processing techniques that introduce context shifts during training can help models decouple objects from their surrounding context and improve performance across the board.

\paragraph{Contributions.}
\begin{itemize}
\item We introduce \textsc{ContextShift}, a controlled, fully reproducible benchmark, along with the \textsc{ContextShift} Dataset, which allows isolation of context as an independent variable via systematic object--background manipulation with a continuous compatibility axis.
\item We identify a consistent failure mode under context variation: \emph{prediction suppression} consistent with reduced candidate formation rather than confidence degradation.
\item We show that performance degrades non-monotonically with compatibility. It peaks at mid-range NPMI and degrades toward both extremes via distinct mechanisms: context deficit at low compatibility and unmet specificity at high compatibility. 
\item We show that both behaviors generalize across controlled manipulations and unmodified real images, reflecting intrinsic detector dynamics rather than artifacts of the evaluation pipeline, and present a proof-of-concept mitigation approach using augmented training data which partially recovers performance across context manipulations.
\end{itemize}

\section{Related Work}
\label{sec:related_work}

Context plays a central role in visual recognition, improving detection and segmentation~\cite{mottaghi2014role}, but also introducing \emph{contextual bias} through spurious object–environment correlations~\cite{singh2020dont,wang2022clad}. While well studied in classification, it remains less systematically explored in object detection.

Detectors often rely on context beyond the object region. D-RISE~\cite{petsiuk2021drise} showed that models attend to background regions when forming predictions, while Dvornik et al.~\cite{dvornik2018modeling} demonstrated that mismatched object--background pairings degrade performance. Son et al.~\cite{son2024quantifying} further established a causal link between co-occurrence and degradation under distribution shift. Related work on out-of-context behavior and spurious correlations has primarily focused on classification or simplified settings: Acharya et al.~\cite{acharya2022detecting} and Lynch et al.~\cite{lynch2023spawrious} analyze spurious correlations, and Xiao et al.~\cite{xiao2021noise} show that models can classify from background alone, motivating the need for detection-level analysis. 
Adversarial manipulation studies provide complementary evidence of strong context sensitivity. Saha et al.~\cite{saha2020spatial} show that modifying the background alone can suppress detections, highlighting the extent to which predictions depend on contextual cues. At the benchmark level, COCO-C~\cite{michaelis2019benchmarking} evaluates robustness to synthetic corruptions such as noise, blur, and weather, exposing sensitivity to low-level perturbations but missing more sophisticated context changes. COCO-O~\cite{mao2023cocoo} extends this to natural distribution shifts across domains (e.g., painting, cartoon, sketch) and introduces the notion of \emph{effective robustness}, showing that improvements on COCO do not transfer reliably. However, both benchmarks vary multiple factors simultaneously---including object appearance, scene composition, and background---making it difficult to isolate the specific role of context in performance degradation, and by mostly focusing on AP metrics, they miss out on specific mechanisms of failure.
Several mitigation strategies have been proposed, including augmentation and contrastive approaches~\cite{wang2022clad,noohdani2024decompose,ghiasi2021copypaste}, but these lack controlled benchmarks for systematically measuring and analyzing context dependence.

Inspired by such prior considerations, in this paper we introduce a controlled benchmark that systematically manipulates object--context relationships while preserving object identity, enabling analysis of context dependence and distinguishing reduced candidate formation from confidence degradation. We show that the same suppression behavior appears under COCO-O, highlighting their complementarity: existing benchmarks quantify how much performance drops, while \textsc{ContextShift} reveals how and why it degrades, allowing for deeper understanding of degradation mechanics and opening new avenues of discussion and mitigation of this phenomena.

\section{Benchmark Design}
\label{sec:benchmark_design}

\subsection{Problem Definition}
We consider standard object detection, where a model maps an image $x$ to detections $\mathcal{D}(x)$. We represent each image as $x = (\mathcal{O}, c)$, where $\mathcal{O} = {(o, a)}$ is the set of object--annotation pairs in the image and $c$ denotes the context. To study context sensitivity, we define manipulation functions $ f_{\text{manip}}(x) = x' = (\mathcal{O}', c')$, where $\mathcal{O}' \subseteq \mathcal{O}$ and $c'$ is the transformed context. We consider two types: \emph{implicit} manipulations, which alter geometric relationships between a given object and the rest of the scene, and \emph{explicit} manipulations, which modify background context directly. In geometric and natural background manipulations, both of these manipulations are restricted to at most one object--annotation pair to isolate context effects on individual object detection.

\subsection{Object-Context Compatibility Modeling}
\label{sec:compatibility_modeling}

To tackle the object-context definition problem, one must ask - how can  compatibility between an object and its context be numerically defined? More importantly -- how do \emph{detection models} infer this relationship during training? To answer this we use a data-driven prior derived from large-scale image statistics. Specifically, we compute an object--context compatibility matrix over COCO object classes and Places365 scene categories using Normalised Pointwise Mutual Information (NPMI)~\cite{bouma2009normalized}, estimated from image-level co-occurrence.
NPMI provides a bounded, continuous axis ($[-1,1]$) that orders object--background pairs from statistically mismatched to highly associated. We use this axis purely as a \emph{structured evaluation variable} to control natural background substitutions, rather than as a measure of perceptual realism.
The compatibility matrix is precomputed by scanning Places365~\cite{zhou2017places} with an open-vocabulary detector (YOLO-World~\cite{cheng2024yolo}), enabling consistent and detector-agnostic evaluation, to be used in future work. We perform correlation analysis with the 5 models used for evaluation in this paper, which confirms that this prior positively correlates with detector-specific co-occurrence patterns Table~\ref{tab:compatibility_corr}. Full construction details are provided in Appendix~\ref{sec:app_compatibility_modeling}.

\begin{table}[t]
\centering
\caption{Correlation analysis with the five models used for evaluation in this paper --- Faster R-CNN~\cite{ren2015fasterrcnn}, YOLO26M~\cite{redmon2016yolo,jocher2023ultralytics}, Deformable DETR~\cite{zhu2021deformabledetr}, RF-DETR~\cite{roboflow_rf_detr}, and D-FINE~\cite{dfine2024} (Appendix~\ref{sec:appendix_models}). Pearson correlation between the precomputed NPMI compatibility prior and detector-specific estimates.}
\label{tab:compatibility_corr}
\small
\begin{tabular}{lccccc}
\toprule
 & Def. DETR & D-FINE & Faster R-CNN & RF-DETR & YOLO26M \\
\midrule
NPMI Corr. (\%) & 67.7 & 68.4 & 67.7 & 78.1 & 77.8 \\
\bottomrule
\end{tabular}
\end{table}

\subsection{Context Manipulations}
\label{sec:implicit_manipulations}
To isolate context as an independent variable, we systematically manipulate object--context relationships while preserving object identity and appearance. This section describes the manipulation families and the compatibility modeling used to study object--background interactions in a controlled and interpretable manner.

\subsubsection{Implicit Manipulations}

Implicit manipulations, named this way since they alter the object’s spatial relationship to the existing context without explicitly modifying the background context itself, apply geometric transformations to a single object $(o,a)$ while preserving appearance features, i.e., $\mathcal(o,a)' = \mathcal(o, a')$ and $c' = c$. We consider similarity transformations (rotation, scaling, and translation) at predefined severity levels $s \in \mathcal{S}_m$.
Although the background is unchanged, these transformations alter spatial configuration (position, scale, orientation), probing sensitivity to the relationship with the context. Invalid cases (e.g., boundary overflow, overlaps) are filtered out (Appendix~\ref{sec:appendix_filtering_criteria}). Only one object is modified in each image to isolate manipulation effects. The object's original footprint is reconstructed using LaMa~\cite{suvorov2022resolution}, a Fourier-convolution network trained for object removal that fills holes with globally coherent background texture rather than boundary-propagated smearing.

\subsubsection{Explicit Context Manipulations}

Explicit manipulations alter the background while preserving the object count and appearance. In other words, explicit manipulations preserve $\mathcal{O}' = \mathcal{O}$. We allow two type of explicit manipulations:

\begin{description}

\item{Synthetic backgrounds:}
We replace $c$ with simple patterns (solid color, gradient, low-frequency noise), preserving object features while removing semantic context. 

\item{Natural backgrounds:}
We replace $c$ with a natural background $c'$ sampled from Places365\cite{zhou2017places}, yielding $x' = (\mathcal{O}, c')$. Backgrounds are filtered to contain no detectable objects, so that detections in the composited image can be attributed to the original object set $\mathcal{O}$ rather than unlabeled scene content. Compatibility is defined in an object-centric manner, evaluating each $(o, c')$ pair independently, after realizing the
composition using a blending pipeline (Appendix~\ref{sec:appendix_bg_blending}). The NPMI-based compatibility function $\kappa(o, c')$ is computed using the process in Section~\ref{sec:compatibility_modeling}.

\end{description}

\begin{figure*}[htbp]
\centering
\begin{subfigure}[t]{0.32\textwidth}
\includegraphics[width=\linewidth]{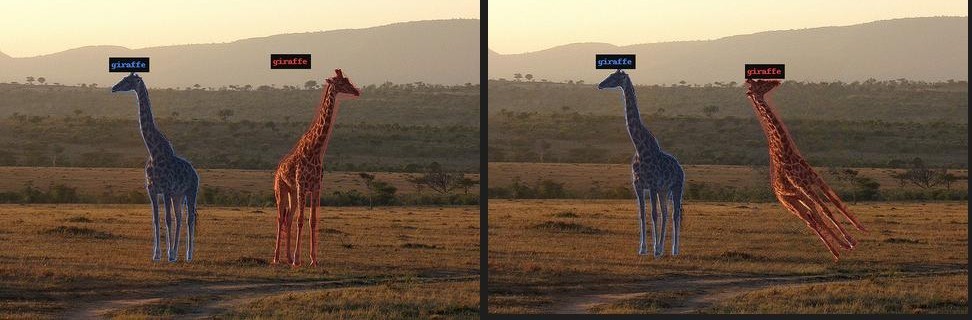}
\caption{Implicit (Rotation)}
\end{subfigure}
\hfill
\begin{subfigure}[t]{0.32\textwidth}
\includegraphics[width=\linewidth]{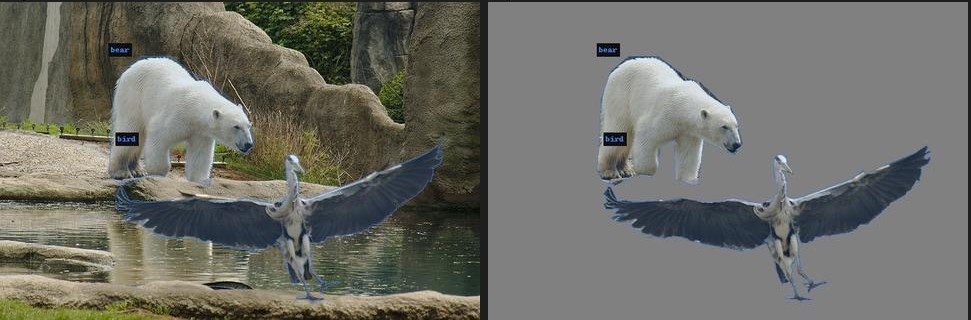}
\caption{Explicit (synthetic background)}
\end{subfigure}
\hfill
\begin{subfigure}[t]{0.32\textwidth}
\includegraphics[width=\linewidth]{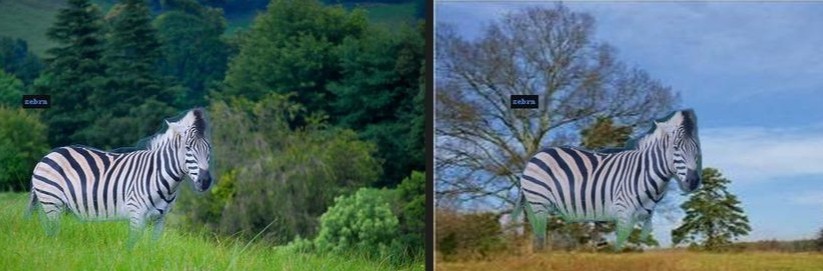}
\caption{Explicit (Natural background)}
\end{subfigure}

\caption{Representative implicit (rotation) and explicit (synthetic and natural) context manipulations. Additional full examples with all variants are provided in Figure~\ref{fig:manip_examples_appendix}.}
\label{fig:manip_examples}
\end{figure*}

Figure~\ref{fig:manip_examples} illustrates representative examples of implicit and explicit manipulations.

\subsection{Constructed Dataset}
To materialize our benchmark design, we create the \textsc{ContextShift} dataset.
It is constructed from the COCO 2017 validation set~\cite{lin2014coco} using the proposed manipulation framework. Starting from $\mathbf{4{,}952}$ base images (out of $\mathbf{5{,}000}$ total COCO images, having filtered out 48 images with no annotated objects), we generate $\mathbf{131{,}885}$ samples across implicit and explicit context manipulations. Implicit and synthetic variants are materialized as discrete datasets for reuse and benchmarking, while compatibility-based natural background substitution is performed in-pipeline to avoid combinatorial explosion, enabling fine-grained evaluation over a continuous compatibility space. Table~\ref{tab:dataset_stats} summarizes the dataset composition; full per-manipulation dataset details are provided in Appendix~\ref{sec:appendix_dataset}.

\section{Evaluation}

We evaluate five detectors spanning two-stage, one-stage, and transformer-based paradigms (cf. Table~\ref{tab:compatibility_corr}) on the COCO 2017 validation set under two manipulation classes: implicit (geometric) and explicit (synthetic backgrounds, compatibility-driven natural backgrounds). All results are reported relative to the clean baseline; For discrete manipulations (geometric and synthetic backgrounds), metrics are averaged over manipulation severity levels $s \in \mathcal{S}_m$; for continous compatibility-based background manipulations, they are aggregated along the compatibility axis (Appendix~\ref{sec:appendix_metrics}). Unless stated otherwise, inference uses a fixed confidence threshold $\tau=0.25$, model-default post-processing, fixed input resolution, and no test-time augmentation. We report AP@0.5, relative area under the severity curve (rAUC), and behavioral metrics $\Delta$FN/img, $\Delta$FP/img, and $\Delta$pred/img.  Full protocol details are provided in Appendix~\ref{sec:appendix_evaluation_protocol}.

\section{Results}
\FloatBarrier
We report cross-model summaries here, with full results in Appendix~\ref{sec:appendix_detailed_results}.
Across all settings, degradation follows a consistent pattern: false negatives increase and prediction volume decreases, while false positives remain relatively stable (compared to false negatives). As illustrated in Section~\ref{sec:prediction_candidate}, this performance degradation is likely explained by reduced formation of valid detection candidates rather than incorrect detections.

\subsection{Implicit Context Manipulation}

All geometric manipulations exhibit prediction suppression (FN$\uparrow$, predictions$\downarrow$, FP$\approx$), with severity primarily driven by transformations that alter object scale or canonical appearance.
Table~\ref{tab:robustness_summary} reports robustness metrics averaged across models, computed only on manipulated objects to isolate each transformation's direct effect. Full per-model breakdowns and full-object-set results are provided in Appendix~\ref{sec:appendix_results_implicit}. Translation produces the mildest degradation (rAUC $=0.94$) with only a moderate reduction in prediction volume ($\Delta$pred/img $=-9.3\%$), indicating limited sensitivity to local spatial-context shifts.
Scaling manipulations reveal a predictable asymmetry. Shrinking substantially degrades robustness (rAUC $=0.81$) and produces the largest FN increase ($\Delta$FN/img $=+106.2\%$), suggesting that down-scaling causes objects to frequently fail in generating valid detection hypotheses (plausibly as a result of feature degradation). In contrast, enlarging slightly improves performance across models.
Rotation produces the strongest overall degradation despite preserving scale and location, yielding the lowest robustness (rAUC $=0.57$) and the largest reduction in prediction volume ($\Delta$pred/img $=-30.2\%$). This suggests strong sensitivity to canonical orientation within the context and axis-aligned structure learned during training.
Across all degrading transformations, the dominant failure mode is suppression rather than confusion: detections are removed rather than replaced with incorrect predictions. 

\begin{table*}[htbp]
\centering
\caption{Cross-model summary across implicit context manipulations, computed on manipulated objects only. Values report the mean across the five evaluated detectors, with the minimum and maximum across models in parentheses. Full per-model results are provided in Table~\ref{tab:implicit_models_mega_obj}.}
\label{tab:robustness_summary}
\footnotesize
\begin{tabular}{lcccc}
\toprule
 & \textbf{Rotation} & \textbf{Shrinkage} & \textbf{Enlargement} & \textbf{Translation} \\
\midrule
AP rAUC & \textbf{0.57 [0.51, 0.62]} & 0.81 [0.77, 0.83] & 1.05 [1.02, 1.06] & 0.94 [0.89, 1.02] \\
$\Delta$FN/img (\%) & +93.5 [61.9, 147.0] & \textbf{+106.2 [41.4, 168.1]} & $-$8.5 [$-$12.3, $-$5.4] & $-$16.4 [$-$0.2, 40.7] \\
$\Delta$FP/img (\%) & \textbf{+6.1 [1.0, 11.7]} & $-$17.9 [$-$36.2, 5.4] & +4.7 [$-$2.9, 8.9] & $-$19.8 [$-$28.9, $-$2.5] \\
$\Delta$pred/img (\%) & \textbf{$-$30.2 [$-$41.3, $-$21.1]} & $-$18.9 [$-$27.1, $-$12.2] & +4.6 [3.0, 6.5] & $-$9.3 [$-$17.7, $-$5.7] \\
\bottomrule
\end{tabular}
\par\vspace{4pt}
\begin{minipage}{\linewidth}
\footnotesize
Parenthetical values report the model-wise range [min,max] across detectors. Metrics are averaged over images ($N_{\text{rotate}}{=}1{,}582$, $N_{\text{shrink}}{=}4{,}952$, $N_{\text{enlarge}}{=}1{,}376$, $N_{\text{translation}}{=}1{,}355$). Under a conservative bound ($\mathrm{CV}{=}1$), the 95\% confidence interval half-width is at most ${\pm}4.6\%$ (Rotate) and ${\pm}3.0\%$ (Shrink), substantially smaller than observed effects. Results are therefore statistically reliable.
\end{minipage}
\end{table*}

\subsection{Explicit Context Manipulation}

\subsubsection{Synthetic Backgrounds Result Summary}

Synthetic background manipulations reinforce the behavior observed under implicit transformations, consistently inducing prediction suppression. As global image statistics diverge from natural images, suppression becomes more pronounced, with low-frequency noise producing the strongest degradation, followed by gradient and solid color backgrounds (Table~\ref{tab:robustness_summary_synthetic_bg}). We note that in one case (YOLO), AP is increased ($AP rAUC=1.05$), namely in the least degrading manipulation - solid color. However, even in that particular case, prediction suppression is still demonstrated quite clearly. This complementary setting reiterates the implicit-manipulation findings: degradation is driven primarily by missed detections and reduced prediction volume.

\begin{table*}[htbp]
\centering
\caption{Cross-model summary across synthetic background manipulations. Values report the mean across the five evaluated detectors, with the minimum and maximum across models in parentheses. Full per-model results are provided in Table~\ref{tab:bg_models_mega_transposed}.}
\label{tab:robustness_summary_synthetic_bg}
\footnotesize
\begin{tabular}{lccc}
\toprule
 & \textbf{Low Freq Noise} & \textbf{Smooth Gradient} & \textbf{Solid Color} \\
\midrule
AP rAUC & \textbf{0.82 [0.68, 0.96]} & 0.91 [0.83, 0.95] & 0.95 [0.87, 1.05] \\
$\Delta$FN/img (\%) & \textbf{+32.4 [10.8, 67.1]} & +19.6 [6.2, 51.3] & +13.3 [$-$2.7, 42.6] \\
$\Delta$FP/img (\%) & $-$52.2 [$-$62.5, $-$43.9] & \textbf{$-$36.8} [$-$51.5, $-$25.6] & $-$41.4 [$-$53.1, $-$35.0] \\
$\Delta$pred/img (\%) & \textbf{$-$27.6 [$-$41.5, $-$17.0]} & $-$17.2 [$-$33.4, $-$9.7] & $-$15.8 [$-$32.4, $-$9.2] \\
\bottomrule
\end{tabular}
\par\vspace{4pt}
\begin{minipage}{\linewidth}
\footnotesize
Parenthetical values show model-wise range [min,max]. Metrics are computed over COCO\,2017 val ($N{=}4{,}952$). Under a conservative bound, 95\% CI half-widths are $\leq{\pm}1.5\%$, far smaller than observed effects, indicating statistical reliability.
\end{minipage}
\end{table*}

\subsubsection{Natural Backgrounds Result Summary}
\label{sec:natural_bg_results}

To isolate the effect of object--context compatibility without interference from co-occurring objects, we evaluate natural background substitution in a \emph{single-object setting}. Specifically, we choose to analyze images with $|\mathcal{O}| = 1$.
We restrict evaluation to 30 COCO classes (Table~\ref{tab:coco30}) that cover diverse scene-grounded categories (vehicles, animals, furniture, outdoor objects), deliberately excluding food items and small accessories, ensuring that NPMI scores reflect meaningful object--background compatibility rather than local feature degradation.
A full-object-set evaluation is provided in Appendix~\ref{sec:appendix_natural_bg_full_dataset}.

\begin{figure}[ht]
\centering
\includegraphics[width=\textwidth]{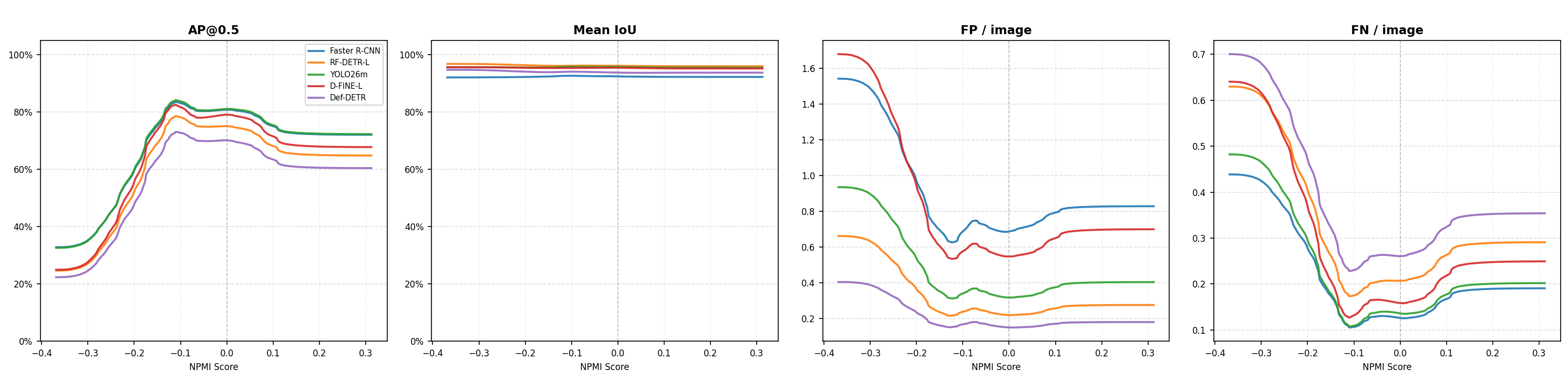}
\footnotesize
\caption{Detection performance as a function of object--background NPMI compatibility ($N{=}1{,}079$ images, sliding window $W{=}500$, stride $50$, Gaussian smoothing $\sigma{=}50$).}

\label{fig:npmi_curves_focal}
\end{figure}

Table~\ref{tab:robustness_summary_natural_bg} demonstrates consistent suppression across models: FN increases while FP declines, with Figure~\ref{fig:npmi_curves_focal} showing all metrics behaving non-monotonically. We discuss this behavior in Section~\ref{sec:comp_detection_behavior}.

\begin{table}[htbp]
\centering
\caption{Robustness summary under natural background substitution (single-object setting, NPMI-based compatibility).}
\label{tab:robustness_summary_natural_bg}
\footnotesize
\begin{tabular}{lccccc}
\toprule
 & \textbf{Def-DETR} & \textbf{D-FINE-L} & \textbf{Faster R-CNN} & \textbf{RF-DETR-L} & \textbf{YOLO26M} \\
\midrule
AP rAUC & 0.81 & 0.87 & 0.89 & \textbf{0.78} & 0.93 \\
$\Delta$FN/img & $+90.2\%$ & $+151.5\%$ & $\mathbf{+227.1\%}$ & $+183.9\%$ & $+158.6\%$ \\
$\Delta$FP/img & $-60.8\%$ & $\mathbf{-33.2}\%$ & $-41.8\%$ & $-53.2\%$ & $-47.8\%$ \\
$\Delta$pred/img & $\mathbf{-44.2\%}$ & $-31.2\%$ & $-34.9\%$ & $-42.9\%$ & $-36.3\%$ \\
\bottomrule
\end{tabular}
\end{table}

\subsubsection{Real-Image Validation of Context-Dependent Suppression}
\label{sec:real_image_npmi_results}

\begin{figure}[ht]
\centering
\includegraphics[width=\textwidth]{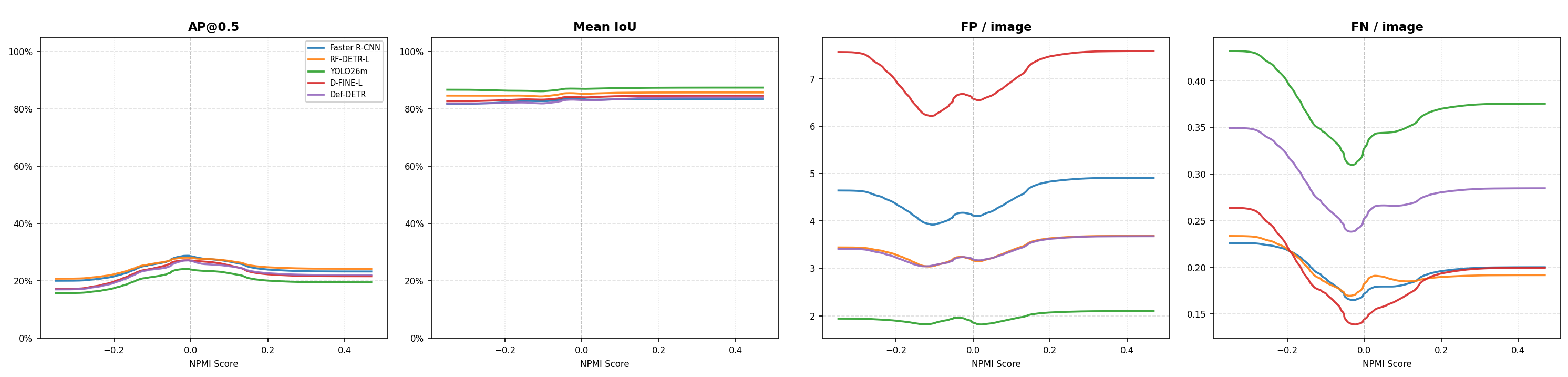}
\caption{Detection performance as a function of NPMI compatibility on real images (focal object evaluation).}
\label{fig:real_npmi_curves}
\end{figure}

To verify that the observed behavior is not a composition artifact, we evaluate
unmodified COCO images by assigning each object instance a compatibility score
derived from its image's scene label, obtained via a ResNet50\cite{he2016resnet}
classifier trained on Places365~\cite{zhou2017places} and applied to the COCO images,
which yields $N{=}36{,}335$ object--context pairs ($4{,}952$ images).
Instances are ranked by NPMI score and divided into five equal-sized quintiles
($n{\approx}7{,}267$ each), from least compatible
(Q1, NPMI$\,{\in}\,[-0.42,\,-0.14]$) to most compatible
(Q5, NPMI$\,{\in}\,[0.04,\,0.70]$). Naturally, middle quintiles are far denser than the extremes.
Table~\ref{tab:npmi_quintiles} shows the same non-monotonic pattern, albeit less
pronounced: performance peaks at intermediate compatibility (Q3,
NPMI$\,{\approx}\,-0.019$) and degrades at both extremes.
The effect is asymmetric---Q1 impairs detection ($-23$ to $-35$\% AP, $+29$ to
$+66$\% FN) relative to Q3, while Q5 shows a smaller but consistent degradation
($-6$ to $-14$\% AP, $+8$ to $+23$\% FN).
Figure~\ref{fig:real_npmi_curves} corroborates these trends, again demonstrating
a U-shaped behavior.
These results are consistent with the behavior observed in
Section~\ref{sec:natural_bg_results}, indicating that it is not solely explained
by the composition pipeline.

\begin{table}[t]
\centering
\caption{Per-model performance across NPMI quintiles on real COCO val2017 images (no compositing or manipulation). Entries are sorted by NPMI score into five equal-sized quintiles ($n \approx 7{,}267$ per bin).}
\label{tab:npmi_quintiles}
\small
\setlength{\tabcolsep}{4pt}
\begin{tabular}{lcccc|cc|cc}
\toprule
& \multicolumn{4}{c|}{AP@0.5} & \multicolumn{2}{c|}{$\Delta$FN/img (\%)} & \multicolumn{2}{c}{$\Delta$FP/img (\%)} \\
\cmidrule(lr){2-5}\cmidrule(lr){6-7}\cmidrule(lr){8-9}
Model & Overall & Q3$^*$ & $\Delta$Q1 & $\Delta$Q5 & $\Delta$Q1 & $\Delta$Q5 & $\Delta$Q1 & $\Delta$Q5 \\
\midrule
Def-DETR       & 0.222 & 0.268 & $-$32.3\% & $-$13.9\% & $+$32.8\% & $+$10.9\% & $+$5.3\% & $+$12.4\% \\
D-FINE-L       & 0.227 & 0.269 & $-$35.1\% & $-$12.3\% & $+$66.0\% & $+$23.0\% & $+$12.2\% & $+$13.3\% \\
Faster R-CNN   & 0.246 & 0.285 & $-$27.7\% & $-$10.7\% & $+$28.7\% & $+$8.4\%  & $+$8.3\% & $+$12.6\% \\
RF-DETR-L      & 0.241 & 0.273 & $-$23.2\% & $-$6.7\%  & $+$31.0\% & $+$8.0\%  & $+$6.5\% & $+$12.5\% \\
YOLO26m        & 0.201 & 0.236 & $-$29.0\% & $-$10.0\% & $+$29.4\% & $+$12.5\% & $+$1.4\% & $+$6.9\%  \\
\bottomrule
\end{tabular}
\vspace{2pt}
\begin{minipage}{0.95\linewidth}
\footnotesize
$^*$Q3 is the best-performing quintile across all models, corresponding to object--scene pairs that co-occur near-chance. All $\Delta$ values are relative to Q3: $\Delta = (Q_k - Q_3)/Q_3$. Q1 contains the least compatible pairs ; Q5 contains the most compatible pairs.
\end{minipage}
\end{table}

\subsubsection{Prediction Candidate Analysis}
\label{sec:prediction_candidate}

We analyze \emph{prediction candidate existence} (IoU $\geq 0.5$, independent of confidence) to distinguish missed detections from low-confidence ones (Appendix~\ref{sec:appendix_candidate_analysis}). Across all manipulations, excluding, solid-color, prediction candidate existence decreases more severely than mean confidence, indicating a failure to form valid detection hypotheses rather than simple score degradation. For natural backgrounds, following the same U-shaped performance trends, the effect varies across the compatibility axis. Table~\ref{tab:npmi_quintiles_focal} shows that existence drops most strongly in highly incompatible contexts (Q1: $\Delta\mathcal{E}=-22.5\%$), improves toward intermediate compatibility, with minimal $\Delta\mathcal{E}$ in Q3, before degrading again in higher-compatibility regimes. This non-monotonic behavior mirrors the U-shaped trend observed at the detection level. Importantly, existence drops remain substantially larger than score changes across all bins, indicating that context primarily affects candidate formation. Suppression persists even in compatible contexts (Q5: $\Delta\mathcal{E}=-9.9\%$), confirming that the failure mode is not limited to strongly mismatched scenes.

\begin{table}[htbp]
\centering
\caption{Candidate-level effect of natural background substitution on the focal object (NPMI-based compatibility quintiles).}
\label{tab:npmi_quintiles_focal}
\footnotesize
\begin{tabular}{lcc}
\toprule
\textbf{Bin} & $\Delta\mathcal{E}$ (mean) & $\Delta\mathcal{S}$ (mean) \\
\midrule
Q1 [-0.40, -0.19) & $-22.5\%$ & $-3.34\%$ \\
Q2 [-0.19, -0.11) & $-7.3\%$  & $+0.26\%$ \\
\textbf{Q3 [-0.11, -0.05) }& $\mathbf{-3.6\%}$  & $\mathbf{+0.84\%}$ \\
Q4 [-0.05, 0.04)  & $-5.9\%$  & $+0.66\%$ \\
Q5 [0.04, 0.74]   & $-9.9\%$  & $-0.57\%$ \\
\bottomrule
\end{tabular}
\vspace{2pt}
\begin{minipage}{\linewidth}
\footnotesize
$\Delta\mathcal{E}$: relative change in detection existence rate (fraction of focal instances receiving at least one candidate with IoU$\,{\geq}\,0.5$) vs.\ the original image.
$\Delta\mathcal{S}$: relative change in mean peak confidence score among matched candidates, computed only when a candidate exists in both conditions.
\end{minipage}
\end{table}

\subsubsection{Effect of object--context compatibility on detection performance}
\label{sec:comp_detection_behavior}
Sections~\ref{sec:natural_bg_results}--\ref{sec:real_image_npmi_results}, have demonstrated a  U-shaped non-monotonic behavior along the NPMI axis: AP and recall peak at intermediate compatibility and degrade toward both low- and high-compatibility conditions. This behavior may be explained by addressing training-induced priors: firstly, low-compatibility contexts provide few contextual cues supporting candidate formation, naturally leading to suppressed detection (Q1: $\Delta\mathcal{E} = -22.5\%$). At intermediate compatibility, where object--context co-occurrence is near chance, detectors operate with minimal contextual bias and achieve the strongest performance. In highly compatible contexts, however, composited objects may violate the specific contextual patterns learned during training (e.g., lighting, scale, or spatial arrangement), again reducing candidate formation (Q5: $\Delta\mathcal{E} = -9.9\%$). This suggests an overfitting-like phenomenon with regard to object--context relationship, where detectors implicitly learn narrow object--context associations and become less reliable when these expectations are only partially matched.

\subsubsection{Validation and Robustness}
A natural question is how much of the observed suppression arises from thresholding, the compatibility definition, or composition artifacts. We address each in turn. Confidence sweeps (Appendix~\ref{sec:appendix_confidence_sweeps}) recover only part of the lost recall, indicating that many missed detections are absent rather than thresholded. Alternative compatibility measures (Appendix~\ref{sec:compatibility_robustness}) reproduce the suppression effect but lack a structured relationship, supporting the need for a meaningful compatibility axis. Finally, composition blending ablations (Appendix~\ref{sec:results_blending_ablation}) preserve the same trends, confirming that suppression and inversion are not artifacts of the composition pipeline.

\section{Mitigation: Context-Aware Augmentation}
\label{sec:mitigation}

Having identified prediction suppression under context shift, it is natural to hypothesize
that training-time exposure to object--context decoupling can mitigate this degrading effect.
We select YOLO to be our representative model for this mitigation proof-of-concept, retraining
it on COCO~train2017, with augmented dataset variants extended with $\sim$11k manipulated
images ($\sim$9\%). While repeating training images is known to reduce data efficiency and degrade detection
performance~\cite{aghabagherloo2025impact, singh2024benchmarking}, any gains
from adding manipulated images drawn from the same dataset must stem from the nature of the manipulation
rather than data volume alone.
These images created by applying compatibility-based background swap (\emph{bg\_swap}) and four
geometric transformations (\emph{shrink}, \emph{enlarge}, \emph{rotate}, \emph{translate}) on
randomly selected images from the original training dataset.

\begin{figure}[htbp]
\centering
\includegraphics[width=\linewidth]{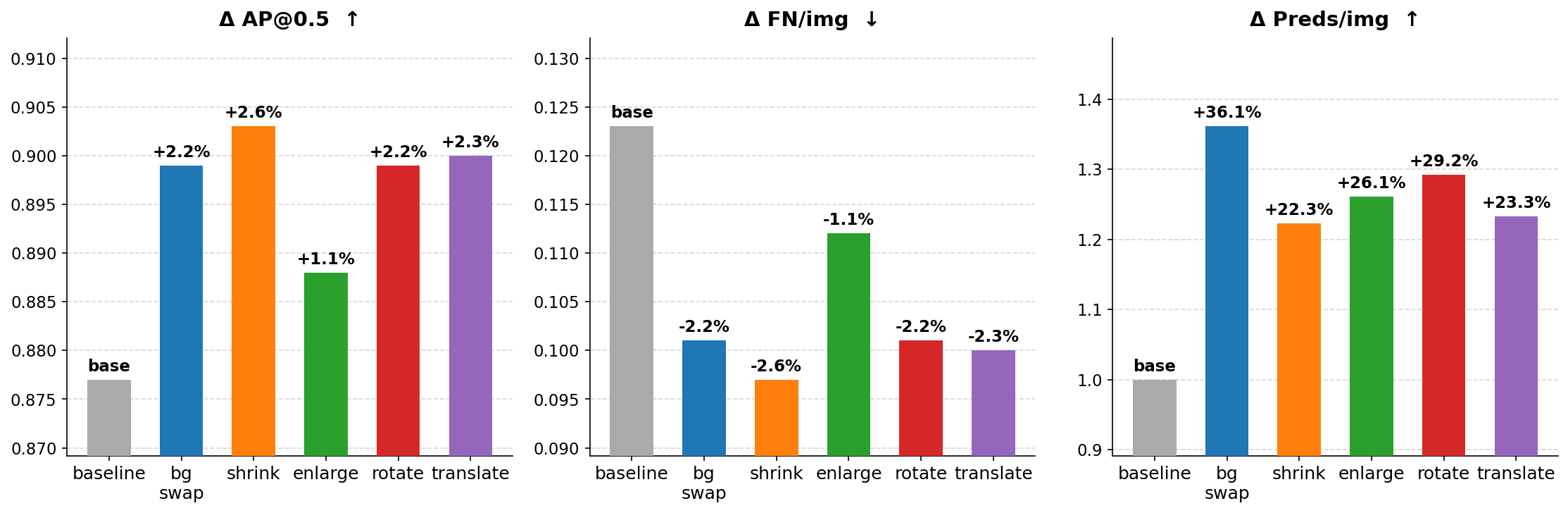}
\caption{
Detection improvements from context-augmentation strategies over an unaugmented baseline.
All five strategies improve AP@0.5 by 1.1–2.6\% and reduce false negatives by 8.9–21.1\%, while increasing total prediction volume by 22–36\%. Results are evaluated on a 990-image held-out split.
}
\label{fig:aug_delta_mitigation}
\end{figure}

Training setup and initialization are identical across variants.
Evaluation was performed on a held-out 990-image split of COCO val2017, under all five context
shifts.
We report performance on clean (unmanipulated) image datasets, with full mitigation pipeline
and evaluation results on clean and manipulated datasets are reported in
Appendix~\ref{sec:appendix_mitigation}.
We find that all augmented variants outperform the baseline on both clean and manipulated data:
clean AP@0.5 improves, FN/img drops, and total prediction volume increases
(Figure~\ref{fig:aug_delta_mitigation}).
These results indicate that prediction suppression is largely data-driven and can be mitigated
via object--context decoupling without harming clean performance.

\section{Impacts, Limitations, and Future Work}

This work introduces a controlled benchmark that isolates context in object detection. By identifying prediction suppression as a consistent failure mode, \textsc{ContextShift} enables analysis beyond aggregate metrics and provides a reproducible framework for comparing models. These findings support the development of context-robust detectors, particularly in settings where missed detections are critical, while also highlighting potential context-based vulnerabilities that could be exploited to confuse detection systems.
The controlled manipulations used in this work enable controlled analysis of object--context interactions while also providing a foundation for future extensions toward more complex real-world distributions. However, one limitation is the current compatibility axis, based on statistical co-occurrence and therefore opens a natural direction toward richer compatibility models that incorporate visual cues such as geometry, lighting, and perspective. Similarly, the object-centric formulation isolates the direct effect of context on individual instances, providing a tractable starting point for future work on multi-object and relational settings involving occlusion, interaction, and contextual reinforcement.
More broadly, this paper focuses on COCO2017 and Places365 as its main evaluated datasets, and extending evaluation beyond these datasets, domains, and modalities will help characterize the generality of prediction suppression under context shift. Finally, applying the framework to related tasks such as segmentation, tracking, and video may reveal whether suppression-based failures represent a broader property of visual recognition systems and could support the development of context-robust training and evaluation strategies.

\section{Discussion \& Conclusions}

\begin{figure}[htbp]
\centering
\includegraphics[width=0.5\linewidth]{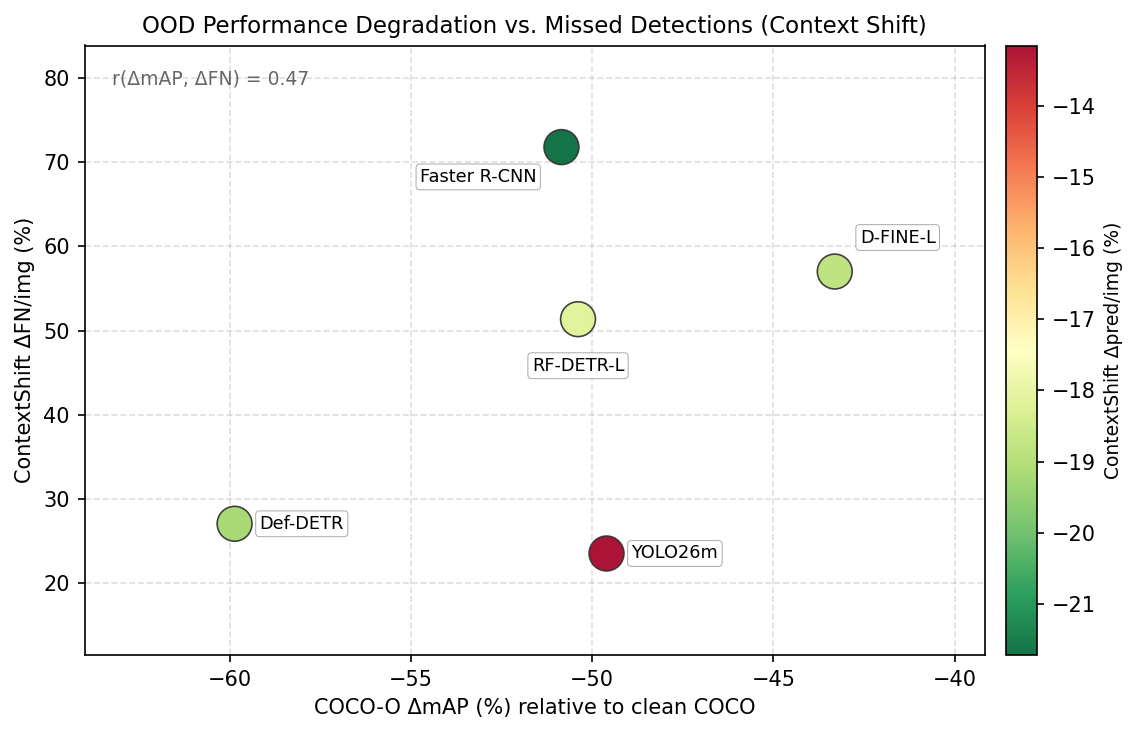}
\caption{
Relationship between COCO-O degradation and context-dependent missed detections.
Each point corresponds to a detector. The x-axis shows relative performance drop ($\Delta$mAP) between COCO and COCO-O, and the y-axis shows \textsc{ContextShift} $\Delta$FN/img (increase in missed detections). Color indicates \textsc{ContextShift} |$\Delta$pred/img|. 
}
\label{fig:cocoo_vs_context}
\end{figure}

We present the \textsc{ContextShift} framework for controlled evaluation and benchmarking of object detection models, aiming to provide a controlled, reproducible protocol for context manipulation, and to serve as a robustness benchmark for existing models and facilitating the development of more robust models, capable of withstanding context shifts with minimal degradation.
To put our work in context with historical approaches, we compare detector behavior on COCO-O (cf. Section~\ref{sec:related_work}). As shown in Figure~\ref{fig:cocoo_vs_context}, despite similar COCO-O $\Delta$mAP degradation under COCO-O-style context variation (between $-49.6\%$ and $-50.8\%$), detectors exhibit large differences in $\Delta$FN/img (between $23.5\%$ and $71.8\%$) and more moderate but still relatively distinguishable differences in $\Delta$pred/img (between $-13.2\%$ and $-21.7\%$) under controlled context-shift, making them easier to differentiate in terms of robustness to context variation. 

Inspired by existing robustness benchmarks, \textsc{ContextShift} is an effort at isolating objects from their context in order to disentangle some of the conflating parameters caused by uncontrolled variation. While some degradation effects may seem obvious at first glance, through a series of structured manipulations we manage to diagnose five modern object detection models with the same failure mode - prediction suppression through candidate existence decrease. We illustrate these effects using the NPMI metric as our compatibility axis, showing that even clean, unmanipulated images, induce the same non-monotonic performance degradation on these models, as our context-shifted images. Finally, we provide a basis for developing new data-centered training methodologies, as demonstrated with a simple, proof-of-concept mitigation strategy.

\newpage
\bibliographystyle{plainnat}
\bibliography{references}

\newpage
\appendix

\section{Manipulation Framework}
\label{sec:appendix_implicit_manipulations}

\subsection{Object-Background Compatibility Modeling}
\label{sec:app_compatibility_modeling}

\subsubsection{Compatibility Metric}

We construct an $|\mathcal{O}| \times |\mathcal{P}|$ compatibility matrix between object classes $\mathcal{O}$ and scene (context) categories $\mathcal{P}$ using \textbf{Normalised Pointwise Mutual Information (NPMI)}~\cite{bouma2009normalized}. NPMI measures the statistical association between an object $o \in \mathcal{O}$ and a scene category $p \in \mathcal{P}$ based on image-level co-occurrence:
\begin{equation}
\mathrm{NPMI}(o, p) = \frac{\log_2 P(o,p) - \log_2 \big(P(o)P(p)\big)}{-\log_2 P(o,p)}.
\end{equation}

Here, $P(o,p)$ denotes the fraction of images assigned to scene category $p$ that contain at least one instance of object $o$. The marginal probabilities are defined as $P(p) = N_p / N$, where $N_p$ is the number of images assigned to scene $p$ and $N$ is the total number of images, and $P(o)$ is the fraction of images in the dataset containing object $o$.

Each scene category $p$ corresponds to a subset of images grouped by their predicted Places365 label~\cite{zhou2017places}. All images assigned to the same category are treated as belonging to the same “place.”

NPMI is bounded in $[-1,1]$, where positive values indicate positive association, values near zero indicate independence, and negative values indicate mutual exclusivity. Importantly, NPMI captures \emph{association strength} rather than raw frequency, allowing a continuous ordering of object--background compatibility independent of dataset imbalance.

We emphasize that NPMI reflects statistical co-occurrence patterns rather than perceptual realism or task difficulty. In this work, it is used strictly as a \emph{controlled evaluation axis} to parameterize background substitutions.

\subsubsection{Scene Data Processing}

To estimate the compatibility matrix, we process the Places365 dataset~\cite{zhou2017places}, containing approximately $1.8$ million images spanning 365 scene categories. Each image is assigned to a scene category using the provided labels.

We scan all images using YOLO-World~\cite{cheng2024yolo}, an open-vocabulary detector, restricted to the 80-class COCO vocabulary~\cite{lin2014coco}. Detection is performed at a confidence threshold of $0.25$, yielding approximately $4.6$ million object detections across the dataset.

From these detections, we compute image-level presence indicators for each object class, ignoring bounding box counts to avoid biases from duplicate detections or object size. These presence indicators are aggregated per scene category to estimate $P(o,p)$, $P(o)$, and $P(p)$.

The resulting compatibility matrix has dimensions $365 \times 80$ (scene categories $\times$ object classes) and is precomputed once as a fixed prior. This enables efficient reuse across experiments and ensures consistency when evaluating different detection models.

\subsection{Examples and Interpretation}

The resulting NPMI values reflect intuitive object--scene associations. For example, \textit{pizza} exhibits high compatibility with \textit{pizzeria} ($\mathrm{NPMI} = +0.87$), near-zero compatibility with \textit{pub-indoor} ($\mathrm{NPMI} \approx 0$), and negative compatibility with \textit{zen\_garden} ($\mathrm{NPMI} = -0.18$). These values define a continuous spectrum used to sample backgrounds at varying compatibility levels.

\subsection{Detector Consistency Analysis}

We evaluate whether the YOLO-World-derived compatibility matrix reflects detector-agnostic co-occurrence structure by comparing it to compatibility estimates computed independently from each detector.

For each detector, we construct a compatibility matrix using the same NPMI formulation applied to its own detections, and compute the Pearson correlation with the precomputed prior. Results (Table~\ref{tab:compatibility_corr}) show consistent positive correlation across all architectures (67.7\%–78.1\%).

This agreement indicates that while imperfect, the compatibility structure captured by the precomputed matrix is largely stable across detectors, supporting its use as a shared evaluation prior.

Additional examples and qualitative interpretations of compatibility values are provided below.

\subsection{Implicit Manipulation Filtering Criteria}
\label{sec:appendix_filtering_criteria}

To ensure validity and avoid degenerate cases, we applied manipulation-specific constraints, summarized in Table~\ref{tab:manipulation_filters}. 

\begin{table}[htbp]
  \caption{Manipulation-specific filtering criteria. Filters ensure transformed objects remain valid, visible, and geometrically feasible after manipulation.}
  \label{tab:manipulation_filters}
  \centering
  \small
  \begin{tabular*}{\columnwidth}{@{\extracolsep{\fill}} p{1cm} p{6cm} p{4cm} @{}}
    \toprule
    Manipulation & Overlap filters & Additional constraints \\
    \midrule
    Shrink &
    --- &
    --- \\
    Enlarge &
    no annotation overlap or image border overflow at any manipulated scale&
    bbox $< 50\%$ image  \\
    Rotate &
    no annotation overlap or image border overflow at any manipulated angle\textsuperscript{1} &
    bbox $< 25\%$ image \\
    Translate &
    no annotation overlap at any manipulated translate &
    enough image pad in a single direction (up, right, down or left) at 40\% image size \\
    \bottomrule
  \end{tabular*}

  \vspace{0.3em}
  \footnotesize{\textsuperscript{1} The rotation cap accounts for worst-case axis-aligned bounding-box expansion at $45^\circ$, preventing rotated boxes from exceeding image boundaries.}
\end{table}

\subsection{Additional Manipulation Examples}
\label{sec:appendix_manipulation_examples}

\begin{figure*}[htbp]
\centering

\begin{subfigure}[t]{\textwidth}
    \centering
    \includegraphics[width=\textwidth]{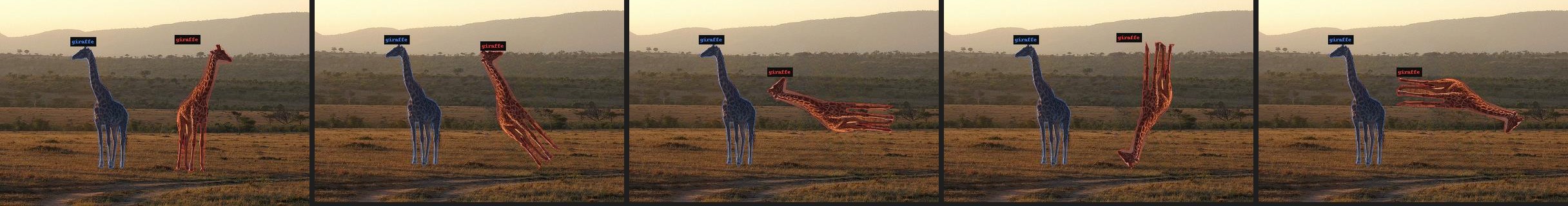}
    \caption{Implicit manipulation: rotation across all evaluated angles.}
\end{subfigure}

\vspace{0.8em}

\begin{subfigure}[t]{\textwidth}
    \centering
    \includegraphics[width=\textwidth]{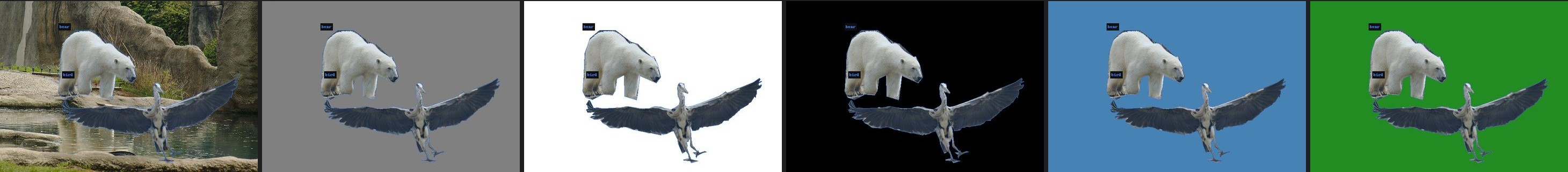}
    \caption{Explicit manipulation: synthetic background substitutions across representative variants.}
\end{subfigure}

\vspace{0.8em}

\begin{subfigure}[t]{\textwidth}
    \centering
    \includegraphics[width=\textwidth]{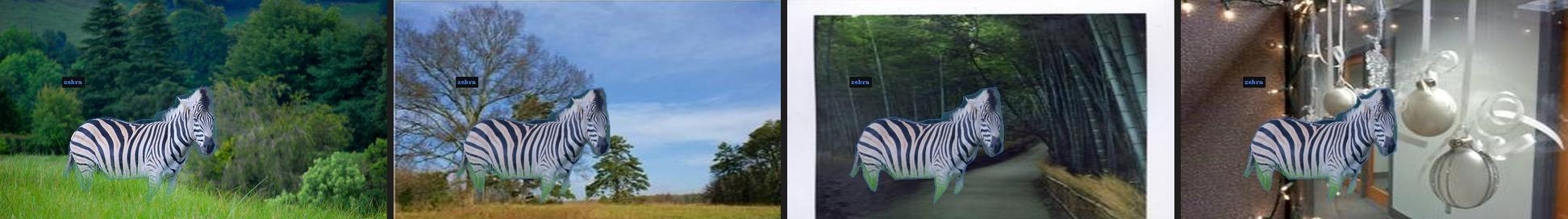}
    \caption{Explicit manipulation: natural background substitutions across varying compatibility levels.}
\end{subfigure}

\caption{Full manipulation examples with all variants for representative images. Top: implicit rotation variants. Middle: synthetic background variants. Bottom: natural background substitutions ordered by compatibility.}
\label{fig:manip_examples_appendix}
\end{figure*}

Figure~\ref{fig:manip_examples_appendix} provides full examples of the manipulation families used in \textsc{ContextShift}, showing the complete set of variants for one representative image from each family.

\subsection{Natural Backgrounds - Pipeline and Parameters}
\label{sec:appendix_bg_blending}

To composite segmented objects onto new backgrounds, we apply a four-stage blending pipeline. Each stage addresses a specific source of visual artifact introduced by naive copy-paste composition.

\paragraph{Stage 1 — Edge Decontamination (\texttt{decontaminate\_px}).}
The segmentation polygon extracted from the source image inevitably captures a thin fringe of source-background pixels along the object boundary. We remove this fringe by eroding the alpha mask with an elliptical structuring element of diameter $2r{+}1$ pixels, where $r$ is \texttt{decontaminate\_px}. Setting this parameter to zero disables the stage.

\paragraph{Stage 2 — Alpha Feathering (\texttt{feather\_px}).}
A hard polygon boundary produces a sharp, visually implausible edge in the composite. We replace it with a smooth alpha transition by applying a Gaussian blur to the eroded alpha channel, with $\sigma = \texttt{feather\_px}/2$. The blur radius effectively controls the width of the soft boundary. Setting this parameter to zero retains a hard edge.

\paragraph{Stage 3 — Local Luminance Matching (\texttt{luminance\_match}, \texttt{luminance\_max\_shift}).}
Objects extracted from brightly lit scenes can appear incongruous on darker backgrounds, and vice versa. When enabled, this stage computes the median luminance of the background patch beneath the object's bounding box and scales the object's RGB channels multiplicatively to match it, capped at $\pm\texttt{luminance\_max\_shift}$ (e.g., $0.15 \Rightarrow {\pm}15\%$). This preserves the object's relative texture and colour while harmonising its global brightness with the scene.

\paragraph{Stage 4 — Object Blur (\texttt{blur\_sigma}).}
High-frequency texture at the object boundary can remain conspicuous even after feathering. A mild isotropic Gaussian blur (standard deviation \texttt{blur\_sigma}) is applied to the object's RGB channels before composition to attenuate these artifacts. Setting \texttt{blur\_sigma} to zero disables this stage.

\begin{table}[htbp]
  \caption{Object blending pipeline parameters and default values. Each stage can be disabled independently by setting its parameter to zero or \texttt{False}.}
  \label{tab:blending_params}
  \centering
  \small
  \begin{tabular*}{\columnwidth}{@{\extracolsep{\fill}} p{2.5cm} p{1cm} p{9cm} @{}}
    \toprule
    Parameter & Default & Description \\
    \midrule
    \texttt{decontaminate\_px} & 1 px &
      Erodes the alpha mask by a $2r{+}1$ elliptical kernel before composition, removing source-background colour fringe at the object boundary. Set to 0 to disable. \\[4pt]
    \texttt{feather\_px} & 2 px &
      Gaussian blur radius ($\sigma = r/2$) applied to the alpha channel, replacing the hard polygon boundary with a smooth transition. Set to 0 to disable. \\[4pt]
    \texttt{luminance\_match} & \texttt{True} &
      If enabled, scales object brightness to match the local background luminance at the paste location. \\[4pt]
    \texttt{luminance\_max\_shift} & 0.15 &
      Maximum multiplicative brightness adjustment (${\pm}15\%$). Only applied when \texttt{luminance\_match} is \texttt{True}. \\[4pt]
    \texttt{blur\_sigma} & 0.5 &
      Sigma of a Gaussian blur applied to the object's RGB channels after composition, softening high-frequency boundary artifacts. Set to 0 to disable. \\
    \bottomrule
  \end{tabular*}
\end{table}

\subsection{Appendix: Dataset Construction and Statistics}
\label{sec:appendix_dataset}

\subsubsection{Dataset Construction}

The dataset is constructed from the COCO 2017 validation split~\cite{lin2014coco} by applying the manipulation framework described in Section~\ref{sec:benchmark_design}. Starting from $4{,}952$ base images, we generate $131{,}885$ samples across implicit and explicit context manipulations.

Implicit transformations (rotation, scaling, translation) and synthetic background substitutions (solid color, smooth gradient, low-frequency noise) are materialized as discrete datasets, enabling direct inspection, reproducibility, and benchmarking across models. 

In contrast, compatibility-driven natural background substitution is evaluated in-pipeline. Since compatibility defines a continuous space, explicitly materializing all object–background combinations would result in a combinatorial explosion. Instead, backgrounds are sampled at evaluation time, allowing controlled, fine-grained analysis across the compatibility spectrum without increasing dataset size.

\paragraph{Dataset Statistics}
Table~\ref{tab:dataset_stats} summarizes the dataset composition across manipulation families. Counts reflect the number of generated image samples per manipulation type.

\begin{table}[htbp]
\centering
\caption{Dataset composition across manipulation families.}
\label{tab:dataset_stats}
\small
\begin{tabular}{lccc}
\toprule
\textbf{Manipulation} & \textbf{Folders} & \textbf{Images/folder} & \textbf{Subtotal} \\
\midrule
background/low\_freq\_noise & 5 (1 original + 4 scales) & 4,952 & 24,760 \\
background/smooth\_gradient & 5 (1 original + 4 variants) & 4,952 & 24,760 \\
background/solid\_color     & 6 (1 original + 5 colors)   & 4,952 & 29,712 \\
geometric/enlarged          & 6 (1 original + 5 scales)   & 1,376 & 8,256 \\
geometric/offset            & 5 (1 original + 4 variants) & 1,355 & 6,775 \\
geometric/rotated           & 5 (1 original + 4 angles)   & 1,582 & 7,910 \\
geometric/shrinked          & 6 (1 original + 5 scales)   & 4,952 & 29,712 \\
\midrule
\textbf{Total} & 38 &  & 131,885 \\
\bottomrule
\end{tabular}
\end{table}

\FloatBarrier

\FloatBarrier
\section{Metric Definitions}
\label{sec:appendix_metrics}

\subsection{Relative Area under the Severity Curve (rAUC).}
\label{sec:appendix_metrics_rauc}
Let $P_m(s)$ denote AP@0.5 under manipulation family $m$ at severity level $s$,
where $s \in \mathcal{S}_m = \{s_1, \ldots, s_{|\mathcal{S}_m|}\}$ corresponds to
increasing transformation magnitude, and $P_{\mathrm{clean}}$ denotes performance on
the unmanipulated baseline.
Robustness is defined as the trapezoidal area under the performance curve across
manipulated severity levels, normalised by clean performance:
\begin{equation}
\mathrm{rAUC}^{(m)} \approx
\frac{1}{P_{\mathrm{clean}}\,(s_{|\mathcal{S}_m|} - s_1)}
\sum_{i=1}^{|\mathcal{S}_m|-1}
\frac{P_m(s_i) + P_m(s_{i+1})}{2}\,(s_{i+1} - s_i).
\end{equation}
$P_{\mathrm{clean}}$ enters only as a normalisation constant, so
$\mathrm{rAUC} = 1$ indicates performance matching the clean baseline on average,
values below 1 indicate degradation, and values above 1 indicate improvement.
For multi-family settings, rAUC is averaged across qualifying families.

\subsection{Behavioral Analysis Metrics.}
\label{sec:appendix_metrics_behavior}
We measure relative differences in false positives (FP), false negatives (FN), and
total predictions per image with respect to the clean baseline.
The per-image change in false negatives is defined as:
\begin{equation}
\delta_s = \frac{\overline{\mathrm{FN}}_s - \overline{\mathrm{FN}}_{\mathrm{clean}}}
{\overline{\mathrm{FN}}_{\mathrm{clean}}} \times 100,
\end{equation}
where $\overline{\mathrm{FN}}_s$ is the mean FN count per image at severity $s$.

For \textbf{discrete} manipulations, $\Delta\mathrm{FN/img}$ is the arithmetic mean
across severity levels:
\begin{equation}
\Delta\mathrm{FN/img} = \frac{1}{|\mathcal{S}_m|} \sum_{s \in \mathcal{S}_m} \delta_s.
\end{equation}

For \textbf{continuous} compatibility-based evaluations, values are aggregated over
$K$ bins via trapezoidal integration along the compatibility axis:
\begin{equation}
\Delta\mathrm{FN/img} =
\frac{1}{x_K - x_1}
\sum_{k=1}^{K-1}
\frac{\delta_k + \delta_{k+1}}{2}\,(x_{k+1} - x_k),
\end{equation}
where $x_k$ is the center of bin $k$.
The same aggregation is applied analogously to $\Delta\mathrm{FP/img}$ and
$\Delta\mathrm{pred/img}$.

\section{Evaluation}
\label{sec:appendix_evaluation_protocol}

\subsection{Detailed Evaluation Protocol}

This appendix provides the full evaluation protocol used throughout \textsc{ContextShift}. Our goal is to define a reusable and standardized procedure for measuring detector behavior under controlled context manipulation, while keeping the main paper focused on the core benchmark design and empirical findings.

\subsection{Evaluation Modes}
\label{section:evaluation_modes}

We define two complementary evaluation modes.

\paragraph{Manipulated-object mode.}
Metrics are computed only on the manipulated target object. This mode isolates the direct effect of the manipulation on the instance whose object--context relationship has been changed. It is the primary mode used for attribution, since it avoids dilution by unmodified objects elsewhere in the image.

\paragraph{Full-image mode.}
Metrics are computed over all annotated objects in the image. This mode captures the overall effect of the manipulation on the detector's full prediction behavior and is intended to reflect end-to-end detection performance in the modified scene.

Unless explicitly stated otherwise, both modes should be reported when applicable. In practice, manipulated-object mode is most informative for understanding the causal effect of the intervention, while full-image mode provides complementary information about broader scene-level consequences.

\subsection{Inference Settings}
\label{appendix:inference_settings}

To ensure comparability across detectors, all evaluations use fixed inference settings unless explicitly stated otherwise:
\begin{itemize}
    \item confidence threshold $\tau = 0.25$,
    \item model-default post-processing and non-maximum suppression,
    \item fixed input resolution per model,
    \item no test-time augmentation.
\end{itemize}

Any deviation from these settings should be stated explicitly. In particular, confidence-threshold sweeps used for behavioral analysis are treated as diagnostic analyses and are reported separately from the main benchmark results.

\subsection{Manipulation Protocol}
\label{appendix:manipulation_protocol}

\paragraph{Implicit context manipulations.}
For implicit manipulations, each family is evaluated over a predefined set of severity levels corresponding to increasing geometric deviation from the original object placement or appearance. These include rotation, scaling (shrink and enlarge), and translation (offset). 

\paragraph{Synthetic background substitutions.}
For synthetic backgrounds, each image is evaluated under a fixed set of generated variants, including solid colors, smooth gradients, and low-frequency noise. These manipulations preserve object appearance while progressively removing natural scene semantics and perturbing global image statistics.

\paragraph{Natural background substitutions.}
For compatibility-driven natural background substitution, evaluation is performed over $K=30$ bins along the compatibility axis. For each source object, one background is sampled from each compatibility bin, producing controlled variation from strongly mismatched to highly compatible contexts. The composition procedure, blending pipeline, and default parameters are described in Appendix~\ref{sec:appendix_bg_blending}.

All evaluations use the same manipulation definitions, filtering criteria, and composition pipeline to ensure consistency across models and experiments.
\subsection{Reported Metrics}
\label{appendix:reported_metrics}

We report both standard performance metrics and behavior-oriented metrics designed
to capture failure modes under context shift.

\paragraph{Standard performance metrics.}
\begin{itemize}
    \item \textbf{AP@0.5}: average precision at IoU threshold 0.5, measured as the
    area under the precision--recall curve where a predicted box is a true positive
    only if its intersection-over-union with a ground-truth box exceeds 0.5.
    \item \textbf{Mean IoU}: mean intersection-over-union over matched true-positive
    predictions, reflecting localization quality independent of recall.
    \item \textbf{rAUC}: relative area under the severity (or compatibility) curve,
    aggregating AP@0.5 across manipulation levels and normalizing by clean-baseline
    performance. This enables comparison across detectors with different absolute accuracy.
\end{itemize}

\paragraph{Behavioral metrics.}
To characterize prediction dynamics under context shift, we additionally report:
\begin{itemize}
    \item $\Delta$\textbf{FN/img}: relative change in false negatives per image,
    \item $\Delta$\textbf{FP/img}: relative change in false positives per image,
    \item $\Delta$\textbf{pred/img}: relative change in total prediction volume per image.
\end{itemize}
Together, these metrics distinguish suppression (FN$\uparrow$, pred$\downarrow$,
FP$\approx$) from confusion (FN$\uparrow$, FP$\uparrow$) and calibration drift.
Formal definitions are provided in Appendix~\ref{sec:appendix_metrics}.

\subsection{Behavioral Analysis Protocol}
\label{appendix:behavioral_analysis}

This section describes the step-by-step procedure for conducting behavioral analysis
of a detector under context shift, as instantiated in this paper.
The goal is to distinguish between two competing failure hypotheses that aggregate
metrics such as AP cannot separate: (i)~\emph{confidence degradation}, where the
model continues to produce candidates but assigns them lower scores, and
(ii)~\emph{prediction suppression}, where the model fails to generate viable
candidates altogether.
We provide the protocol as a reusable diagnostic for researchers applying
\textsc{ContextShift} to new models or extending it to new manipulation families.

\paragraph{Step 1: Establish the suppression signature.}
For each manipulation family and severity level (or compatibility bin), compute
$\Delta$FN/img, $\Delta$FP/img, and $\Delta$pred/img relative to the clean baseline
(formal definitions in Appendix~\ref{sec:appendix_metrics}).
A suppression signature is present when false negatives increase, prediction volume
decreases, and false positives remain stable or decline.
This pattern distinguishes suppression from confusion (FP$\uparrow$, FN$\uparrow$)
and from calibration drift (FP$\approx$, FN$\approx$, confidence shift only).
If false positives increase substantially alongside false negatives, the failure mode
is more likely appearance disruption or score redistribution rather than prediction
suppression, and the analysis should proceed accordingly.

\paragraph{Step 2: Test the confidence hypothesis.}
Compute the distribution of maximum prediction score per ground-truth instance ---
i.e., for each GT box, the highest IoU-matched prediction score, or zero if no
prediction achieves IoU $\geq 0.5$.
A confidence-degradation failure would manifest as a rightward-to-leftward shift
of mass in the high-confidence region.
If instead mass shifts directly to zero (no overlapping prediction) with minimal
redistribution in the intermediate range, this rules out the confidence-degradation
hypothesis.

Complement this with a confidence-threshold sweep: evaluate recall at thresholds
$\tau \in [0.01, 0.50]$ for both clean and manipulated conditions.
Define \emph{recoverable recall gap} as the residual difference at $\tau \to 0$.
A large unrecoverable gap --- where substantial recall loss persists even at very
low thresholds --- confirms that missed detections are absent from the model's
output rather than merely suppressed by thresholding.
We recommend reporting the gap at $\tau = 0.01$ as the practical lower bound.

\paragraph{Step 3: Measure candidate formation directly.}
Compute prediction candidate existence rate $\mathcal{E}$: the fraction of
ground-truth instances for which at least one predicted box achieves
IoU $\geq 0.5$ and score $\geq 0.01$, independent of the inference threshold.
Compute the conditional score $\mathcal{S}$: the mean maximum candidate
confidence, averaged over instances where a candidate exists in \emph{both}
the clean and manipulated conditions.

Report $\Delta\mathcal{E}$ and $\Delta\mathcal{S}$ jointly.
If $|\Delta\mathcal{E}| \gg |\Delta\mathcal{S}|$, the primary mechanism is
candidate formation failure.
If both degrade proportionally, the failure reflects a broader collapse of
both localization and scoring.
If $\Delta\mathcal{S}$ dominates, a confidence-recalibration or temperature-based
mitigation may be appropriate.

\paragraph{Step 4: Stratify by object properties.}
Repeat the candidate existence analysis stratified by object size (COCO thresholds:
$<\!32^2$\,px, $32^2$--$96^2$\,px, $\geq\!96^2$\,px) and, where applicable,
by object category and compatibility bin.
Scale sensitivity is a strong discriminator between manipulation families: shrinkage
predominantly affects small objects, while compatibility-driven context shift operates
more uniformly across scales.
Disproportionate small-object degradation in $\Delta\mathcal{E}$ is consistent with
feature resolution collapse; uniform degradation across scales points to a global
contextual signal.

\paragraph{Step 5: Choose the appropriate evaluation mode.}
Use \emph{manipulated-object mode} (metrics on the focal instance only) to attribute
effects causally to the manipulation.
Use \emph{full-image mode} (all annotated objects) to measure scene-level
consequences and avoid underestimating real-world impact.
Both modes should be reported; manipulated-object mode is primary for diagnostic
attribution, and full-image mode for deployment-oriented robustness estimates
(cf.\ Appendix~\ref{section:evaluation_modes}).

\paragraph{Interpretation summary.}
Table~\ref{tab:behavioral_decision} provides a diagnostic decision table.

\begin{table}[htbp]
\centering
\caption{Diagnostic decision table for characterizing detector failure mode under
context shift.}
\label{tab:behavioral_decision}
\small
\setlength{\tabcolsep}{5pt}
\begin{tabular}{llll}
\toprule
$\Delta\mathcal{E}$ & $\Delta\mathcal{S}$ & Recoverable gap & Failure mode \\
\midrule
large $\downarrow$ & small           & large           & Prediction suppression \\
small             & large $\downarrow$ & small           & Confidence degradation \\
large $\downarrow$ & large $\downarrow$ & moderate        & Compound failure \\
small             & small           & small           & Threshold artifact only \\
\bottomrule
\end{tabular}
\end{table}

\paragraph{Statistical reliability.}
For discrete manipulations, report 95\% confidence intervals under a conservative
coefficient of variation bound (e.g., $\mathrm{CV} = 1$), using the formula
$\mathrm{CI}_{95} = 1.96\,\bar{x}/\sqrt{N}$.
Effects should be considered reliable only when substantially larger than this bound.
For continuous compatibility-based evaluations, stability can be assessed by
repeating the analysis with alternative window sizes or stride values and verifying
that the qualitative shape of the performance curve is preserved.

\FloatBarrier

\subsection{Evaluated Models.}
\label{sec:appendix_models}
To ensure a comprehensive and balanced evaluation, we select a diverse set of object detectors that span the primary paradigms in modern object detection, including two-stage, one-stage, and transformer-based approaches. This selection covers both well-established baselines and recent high-performing models, allowing us to assess whether robustness trends are consistent across fundamentally different design principles. The chosen models differ in their reliance on region proposals, dense prediction, and attention-based global reasoning, as well as in their trade-offs between accuracy, efficiency, and contextual dependence.

\paragraph{Faster R-CNN (ResNet50-FPN).}
Faster R-CNN is a canonical two-stage detector in which a Region Proposal Network first generates candidate object regions, followed by classification and box refinement on these proposals~\cite{ren2015fasterrcnn}. It represents the classical proposal-based detection paradigm and remains a strong, widely used reference point for localization-oriented detection. Including Faster R-CNN allows us to compare modern context sensitivity against a detector whose inference is explicitly structured around candidate regions rather than fully end-to-end set prediction. 

\paragraph{YOLO (YOLO26M).}
YOLO26M represents the one-stage real-time detection family, where detections are produced in a single forward pass with an emphasis on deployment efficiency~\cite{redmon2016yolo,jocher2023ultralytics}. Ultralytics describes YOLO26 as a faster, simpler, end-to-end, NMS-free detector optimized for edge and low-power settings, making it a useful counterpoint to heavier two-stage and transformer-based models. Its inclusion is important for \textsc{ContextShift} because one-stage detectors are commonly used in practical real-time systems, and evaluating such a model helps determine whether context sensitivity persists in detectors designed primarily around speed and streamlined inference. 

\paragraph{Deformable DETR (ResNet50).}
Deformable DETR is a foundational transformer-based detector that addresses the original DETR optimization bottlenecks by introducing deformable attention, which attends to a sparse set of relevant sampling locations and naturally supports multi-scale feature aggregation~\cite{zhu2021deformabledetr}. It is especially relevant here because it serves as a canonical DETR-style model: if robustness trends appear in Deformable DETR as well as in CNN-based detectors, they are less likely to be artifacts of a single architectural family. It also provides a strong test of whether attention-based global reasoning reduces or merely changes dependence on contextual configuration.

\paragraph{RF-DETR (Large).}
RF-DETR is a recent real-time detection transformer designed to achieve strong accuracy--latency trade-offs through neural architecture search and specialist adaptation~\cite{roboflow_rf_detr}. The authors position it as a state-of-the-art NAS-based method for fine-tuning end-to-end object detectors for diverse datasets and hardware targets, and report strong results on COCO and Roboflow100-VL. It is therefore a relevant choice for \textsc{ContextShift} because it represents a modern high-performance DETR-style detector optimized not only for accuracy but also for transferability and deployment, allowing us to test whether context sensitivity remains present in newer real-time transformer designs.

\paragraph{D-FINE (Large).}
D-FINE is a recent high-performing real-time detector that reformulates box regression in DETRs through Fine-grained Distribution Refinement and Global Optimal Localization Self-Distillation~\cite{dfine2024}. Its authors report that D-FINE achieves state-of-the-art real-time accuracy--efficiency performance and substantially improves localization precision while preserving practical speed. This makes it particularly suitable for our benchmark: because \textsc{ContextShift} studies how manipulations affect both detection success and localization behavior, D-FINE provides a strong recent detector whose design explicitly targets precise box prediction, helping assess whether improved localization machinery also improves robustness to structured object--context changes.

These models span two-stage, one-stage, and transformer-based paradigms, enabling evaluation of whether context sensitivity is consistent across fundamentally different architectures.

\section{Detailed Results}
\label{sec:appendix_detailed_results}

\subsection{Implicit Manipulations}
\label{sec:appendix_results_implicit}

\FloatBarrier

\subsubsection{Per-Manipulation Performance Curves - Manipulated Object Only}
\begin{figure*}[htbp]
    \small
    \centering

    \textbf{Shrinkage}\\
    \includegraphics[width=\textwidth]{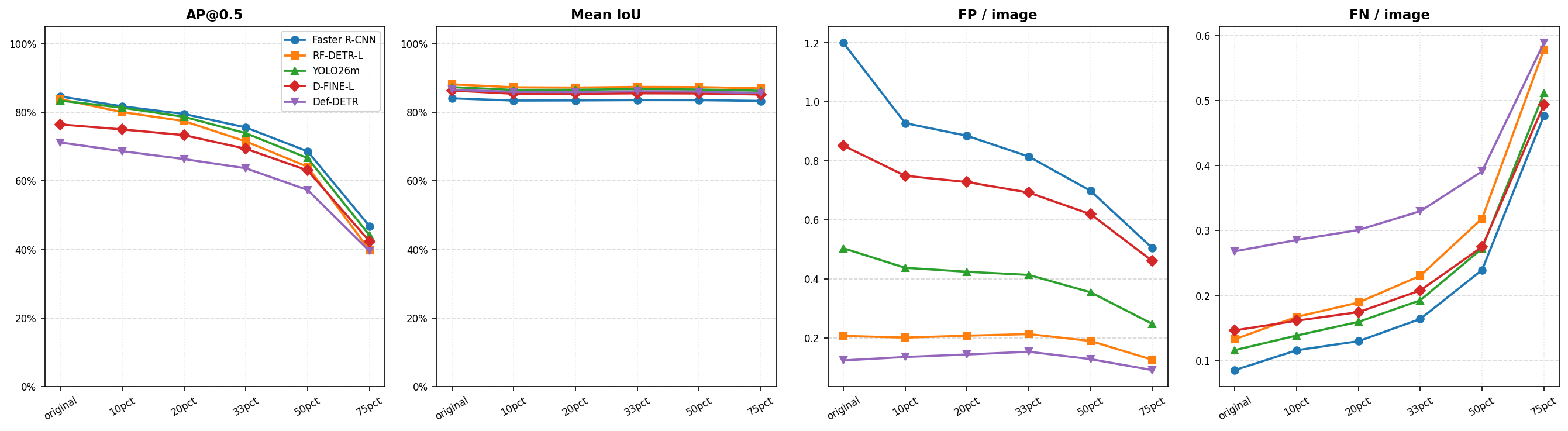}

    \vspace{0.8em}
    \textbf{Rotation}\\
    \includegraphics[width=\textwidth]{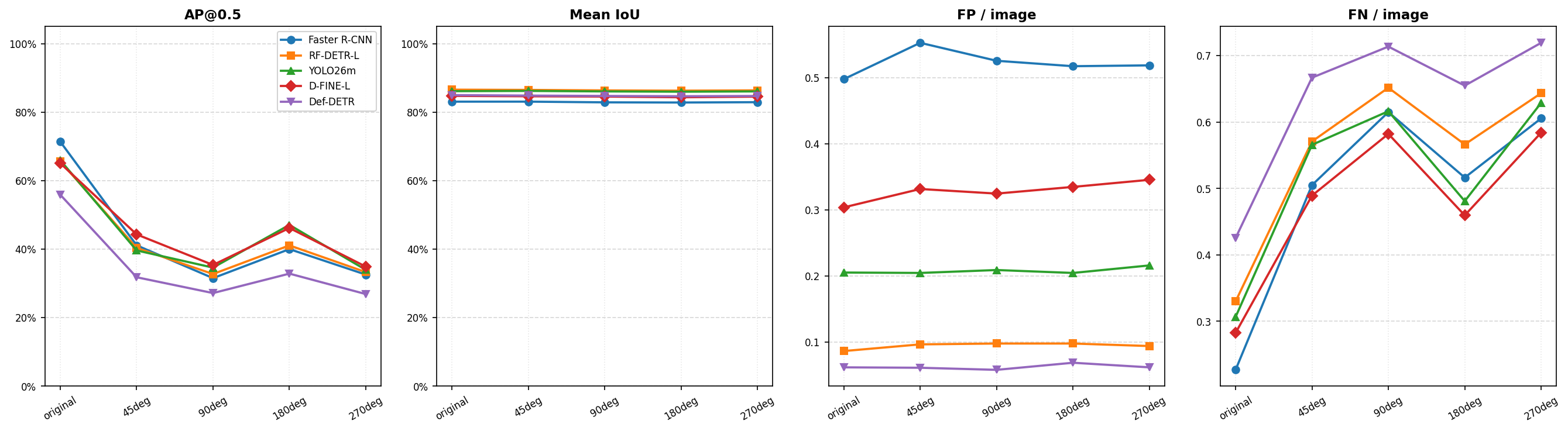}

    \vspace{0.8em}
    \textbf{Enlargement}\\
    \includegraphics[width=\textwidth]{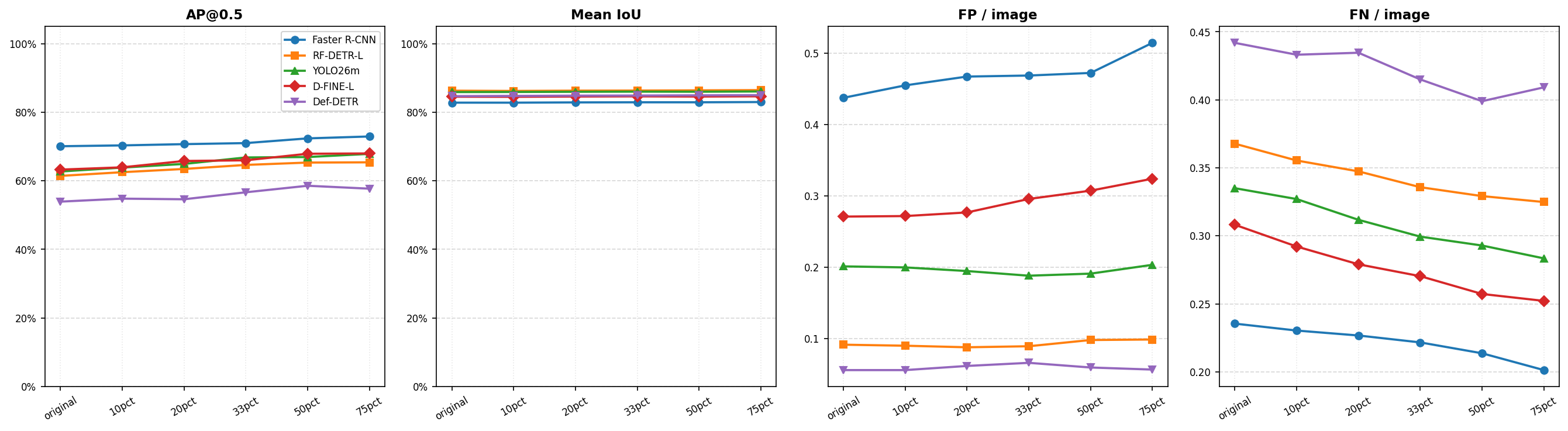}

    \vspace{0.8em}
    \textbf{Translation}\\
    \includegraphics[width=\textwidth]{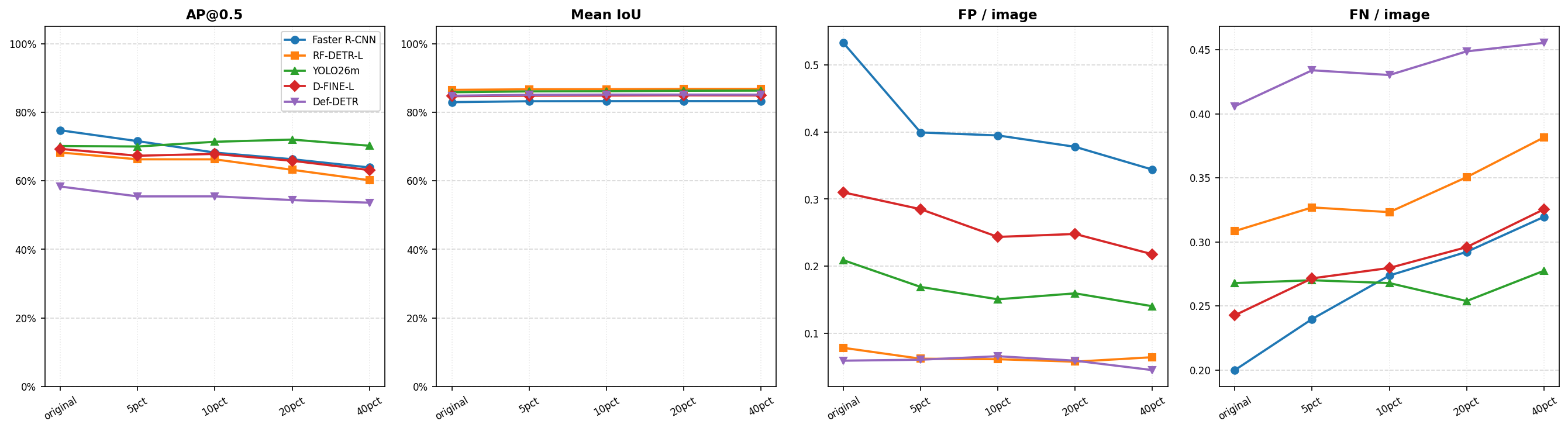}

    \caption{Performance curves under focal-object geometric manipulations. Metrics are computed only on the manipulated object.}
    \label{fig:focal_geometric_curves}
\end{figure*}

Figure~\ref{fig:focal_geometric_curves} shows the performance curves under manipulated-object evaluation.
Each manipulation exhibits a distinct degradation pattern. Shrinkage produces a monotonic decline in AP together with rapidly increasing FN rates, indicating progressive suppression as objects move below the training scale distribution. Enlargement shows the opposite trend: AP slightly improves while FN decreases, suggesting that larger objects align better with the scales most frequently observed during training.

Rotation produces a non-monotonic behavior. Performance degrades sharply at $45^\circ$ and $90^\circ$, partially recovers near $180^\circ$, and destabilizes again toward $270^\circ$. The partial recovery at $180^\circ$ is consistent with the preservation of coarse object structure under inversion, whereas intermediate rotations disrupt the canonical orientation and axis-aligned patterns relied upon during candidate formation. The renewed degradation at $270^\circ$ suggests that robustness is not purely symmetry-driven and remains sensitive to orientation-specific feature statistics learned during training.

Translation produces comparatively mild degradation across all metrics. Because object appearance and scale remain unchanged, the manipulation primarily disrupts local spatial context and positional priors, leading to only moderate increases in FN and relatively stable AP.

\begin{figure*}[htbp]
    \small
    \centering

    \textbf{Shrinkage}\\
    \includegraphics[width=\textwidth]{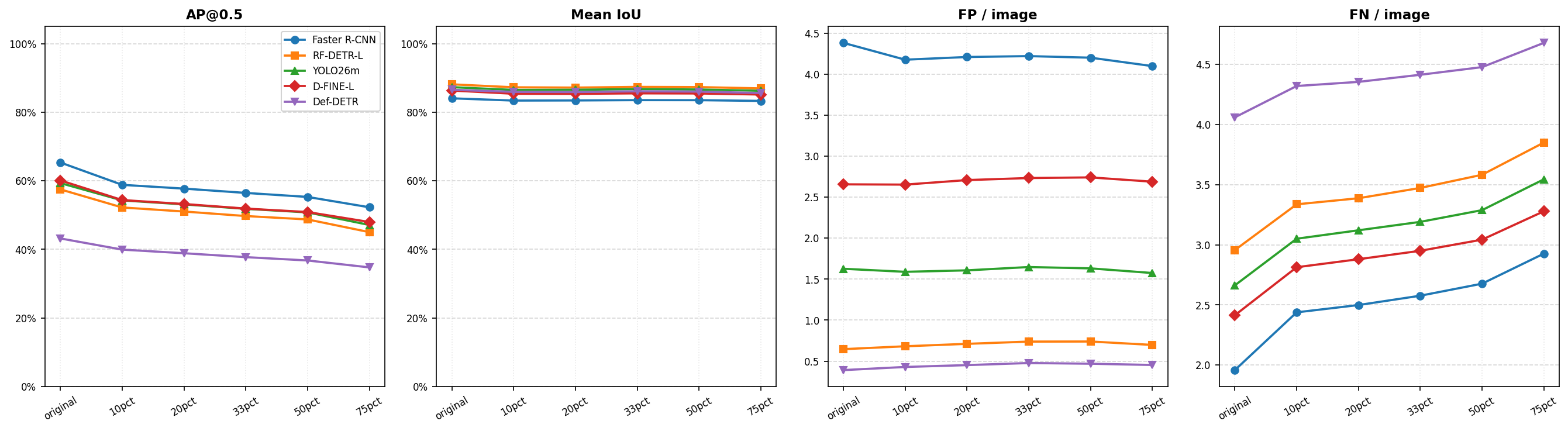}

    \vspace{0.8em}
    \textbf{Rotation}\\
    \includegraphics[width=\textwidth]{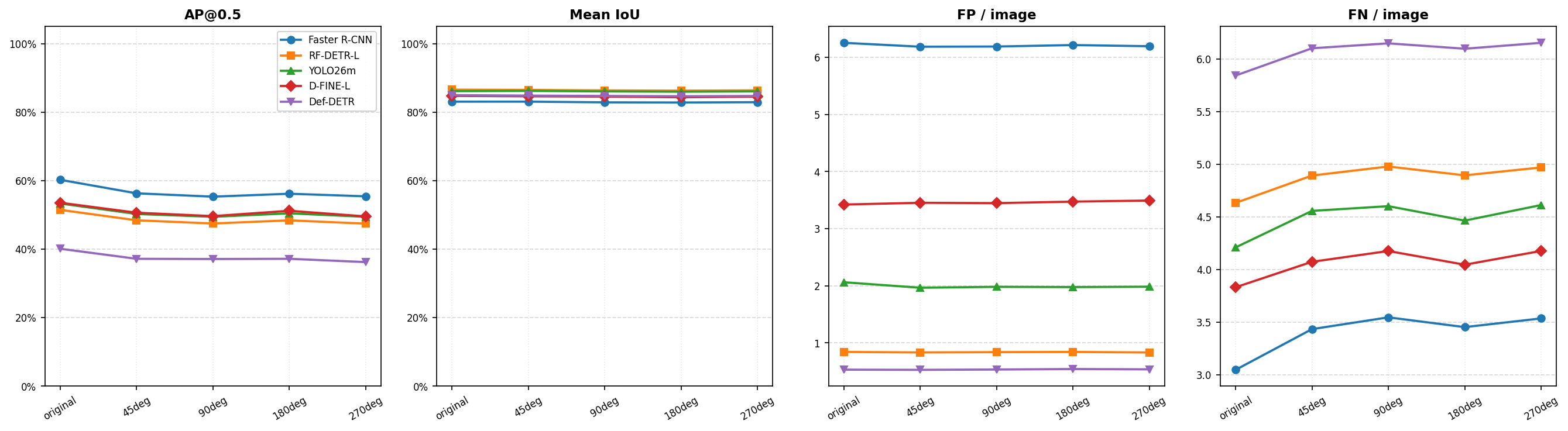}

    \vspace{0.8em}
    \textbf{Enlargement}\\
    \includegraphics[width=\textwidth]{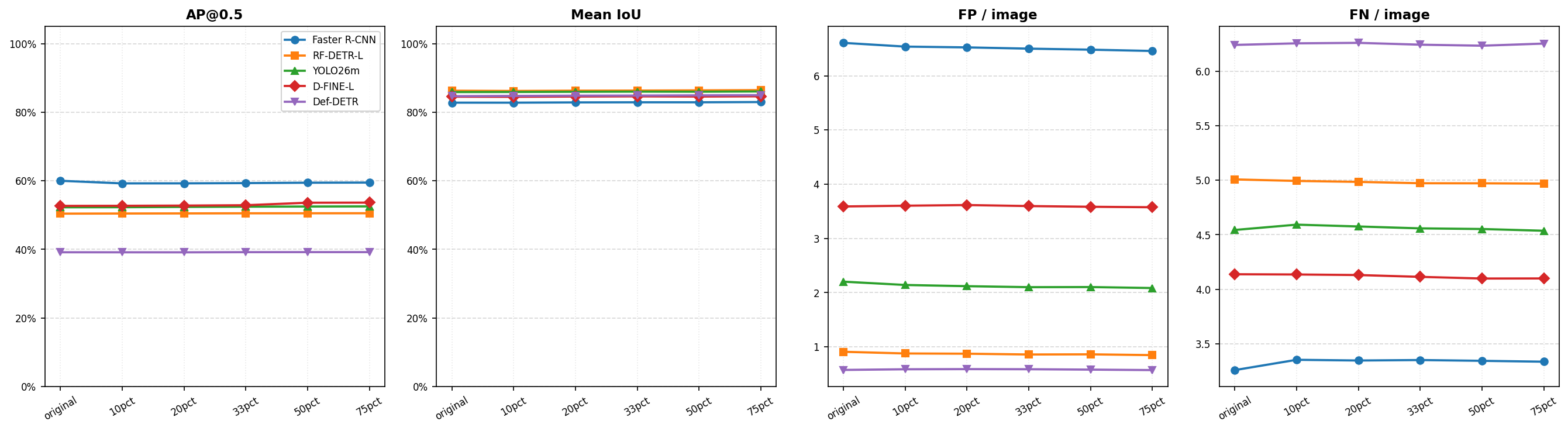}

    \vspace{0.8em}
    \textbf{Translation}\\
    \includegraphics[width=\textwidth]{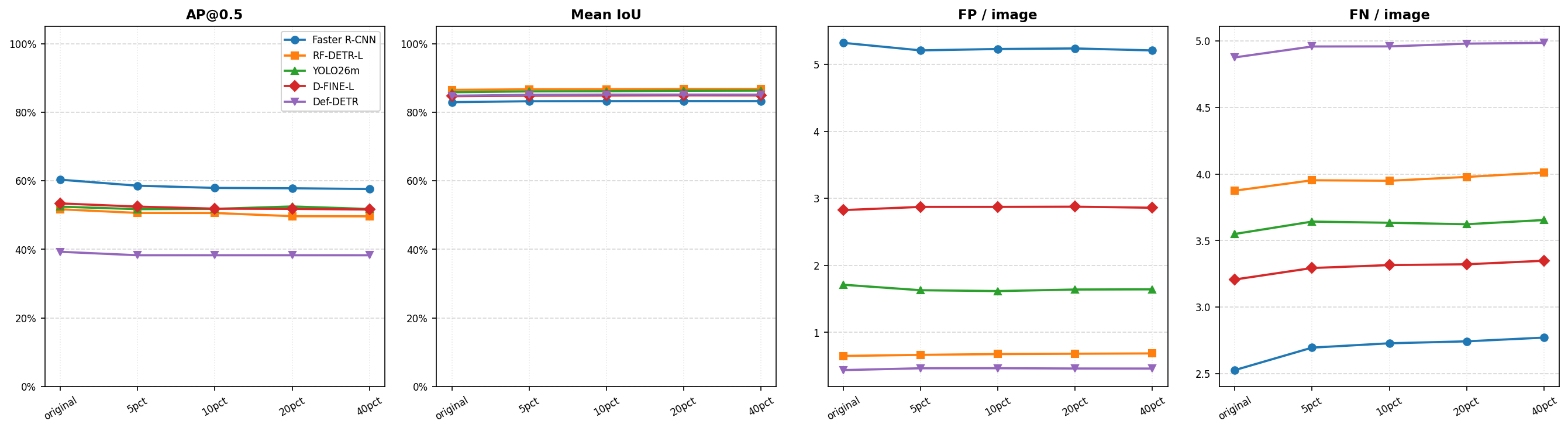}

    \caption{Performance curves under geometric manipulations computed over the full image.}
    \label{fig:geometric_curves_full}
\end{figure*}

\FloatBarrier
\paragraph{Per-Manipulation Performance Curves - Full Object Set}
Figure~\ref{fig:geometric_curves_full} shows the performance curves under full-object evaluation.

\FloatBarrier
\subsubsection{Per-Metric Performance Tables}
Tables~\ref{tab:implicit_models_mega_obj} and~\ref{tab:implicit_models_mega} present the per-model robustness and behavioral metrics.

\begin{table*}[htbp]
\centering
\caption{Per-model robustness and behavioral metrics under implicit context manipulations - computed on manipulated objects only.}
\label{tab:implicit_models_mega_obj}
\small
\setlength{\tabcolsep}{5pt}
\begin{tabular}{llccccc}
\toprule
\textbf{Manipulation} & \textbf{Metric} & \textbf{Deformable DETR} & \textbf{D-FINE} & \textbf{Faster R-CNN} & \textbf{RF-DETR} & \textbf{YOLO26M} \\
\midrule
\multirow{5}{*}{Rotation}
& rAUC              & 0.5342 & 0.6223 & 0.5056 & 0.5631 & 0.6058 \\
& $\Delta$FN/img    & +61.89\% & +87.14\% & +147.01\% & +84.29\% & +86.91\% \\
& $\Delta$FP/img    & +1.02\%  & +9.98\%  & +6.19\%   & +11.68\% & +1.62\% \\
& $\Delta$pred/img  & -41.26\% & -21.13\% & -23.82\%  & -35.42\% & -29.27\% \\
\midrule
\multirow{5}{*}{Shrinkage}
& rAUC              & 0.8124 & 0.8274 & 0.8150 & 0.7720 & 0.8064 \\
& $\Delta$FN/img    & +41.37\% & +79.04\% & +168.11\% & +123.03\% & +119.20\% \\
& $\Delta$FP/img    & +5.37\%  & -23.69\% & -36.17\%  & -9.36\%   & -25.48\% \\
& $\Delta$pred/img  & -12.20\% & -18.63\% & -27.13\%  & -17.05\%  & -19.24\% \\
\midrule
\multirow{5}{*}{Enlargement}
& rAUC              & 1.0564 & 1.0558 & 1.0232 & 1.0521 & 1.0602 \\
& $\Delta$FN/img    & -5.36\% & -12.31\% & -7.10\% & -7.94\% & -9.59\% \\
& $\Delta$FP/img    & +7.27\% & +8.90\%  & +8.70\% & +1.43\% & -2.89\% \\
& $\Delta$pred/img  & +4.52\% & +6.45\%  & +4.56\% & +4.22\% & +3.04\% \\
\midrule
\multirow{5}{*}{Translation}
& rAUC              & 0.9339 & 0.9465 & 0.8885 & 0.9260 & 1.0155 \\
& $\Delta$FN/img    & +8.91\% & +20.74\% & +40.68\% & +12.02\% & -0.21\% \\
& $\Delta$FP/img    & -2.50\% & -19.82\% & -28.88\% & -21.70\% & -25.88\% \\
& $\Delta$pred/img  & -5.76\% & -10.48\% & -17.65\% & -7.02\% & -5.69\% \\
\bottomrule
\end{tabular}
\end{table*}

\begin{table*}[htbp]
\centering
\caption{Per-model robustness and behavioral metrics under implicit context manipulations - full annotated object set.}
\label{tab:implicit_models_mega}
\small
\setlength{\tabcolsep}{5pt}
\begin{tabular}{llccccc}
\toprule
\textbf{Manipulation} & \textbf{Metric} & \textbf{Deformable DETR} & \textbf{D-FINE} & \textbf{Faster R-CNN} & \textbf{RF-DETR} & \textbf{YOLO26M} \\
\midrule
\multirow{5}{*}{Rotation}
& rAUC              & 0.9217 & 0.9402 & 0.9260 & 0.9315 & 0.9363 \\
& $\Delta$FN/img    & +4.84\% & +7.48\% & +14.61\% & +6.50\% & +8.29\% \\
& $\Delta$FP/img    & +0.77\% & +1.29\% & -0.93\% & -0.60\% & -4.12\% \\
& $\Delta$pred/img  & -6.05\% & -2.55\% & -3.83\% & -4.99\% & -5.58\% \\
\midrule
\multirow{5}{*}{Shrinkage}
& rAUC              & 0.8617 & 0.8525 & 0.8523 & 0.8509 & 0.8595 \\
& $\Delta$FN/img    & +9.59\%  & +23.98\% & +33.99\% & +19.27\% & +21.64\% \\
& $\Delta$FP/img    & +16.66\% & +1.83\%  & -4.58\%  & +10.44\% & -1.02\% \\
& $\Delta$pred/img  & -8.84\%  & -7.00\%  & -8.87\%  & -9.99\%  & -9.41\% \\
\midrule
\multirow{5}{*}{Enlargement}
& rAUC              & 1.0006 & 1.0103 & 0.9895 & 1.0019 & 1.0036 \\
& $\Delta$FN/img    & +0.13\% & -0.54\% & +2.68\% & -0.58\% & +0.43\% \\
& $\Delta$FP/img    & +1.52\% & +0.15\% & -1.62\% & -4.98\% & -4.21\% \\
& $\Delta$pred/img  & +0.01\% & +0.28\% & -1.40\% & -0.25\% & -1.37\% \\
\multirow{5}{*}{Translation}
& rAUC              & 0.9743 & 0.9704 & 0.9588 & 0.9659 & 0.9932 \\
& $\Delta$FN/img    & +1.93\% & +3.51\% & +8.22\% & +2.53\% & +2.50\% \\
& $\Delta$FP/img    & +5.82\% & +1.61\% & -1.88\% & +4.44\% & -4.62\% \\
& $\Delta$pred/img  & -1.81\% & -0.85\% & -2.79\% & -1.38\% & -2.62\% \\
\bottomrule
\end{tabular}
\end{table*}

\FloatBarrier

\subsection{Synthetic Backgrounds}
\label{sec:synthetic_results_appendix}
\subsubsection{Per-Manipulation Performance Curves}
\begin{figure*}[htbp]
    \small
    \centering

    \textbf{Solid Color}\\
    \includegraphics[width=\textwidth]{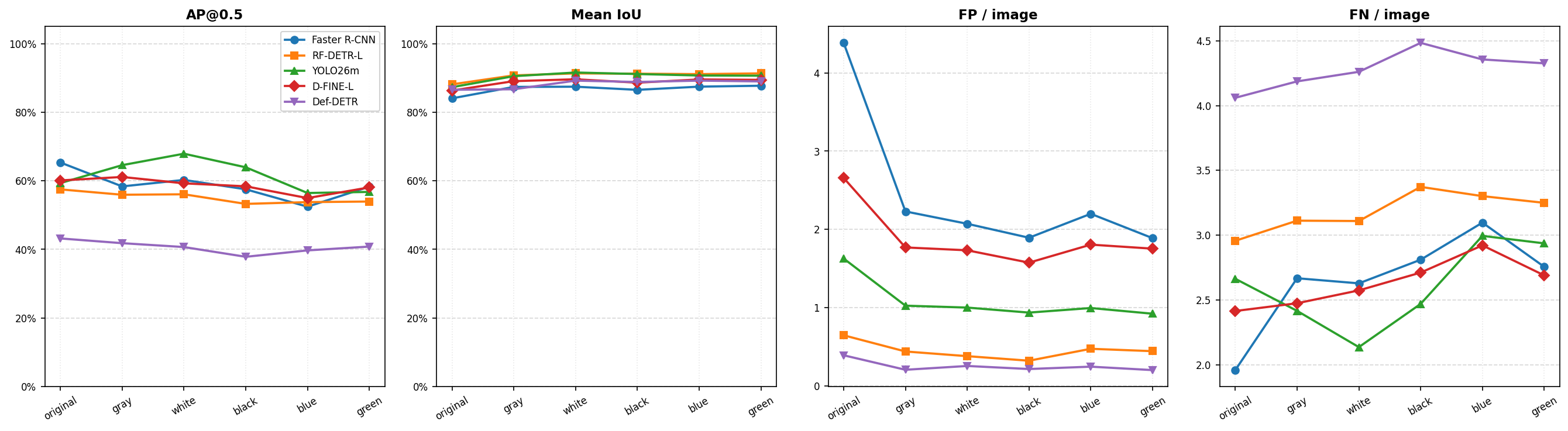}

    \vspace{0.8em}
    \textbf{Smooth Gradient}\\
    \includegraphics[width=\textwidth]{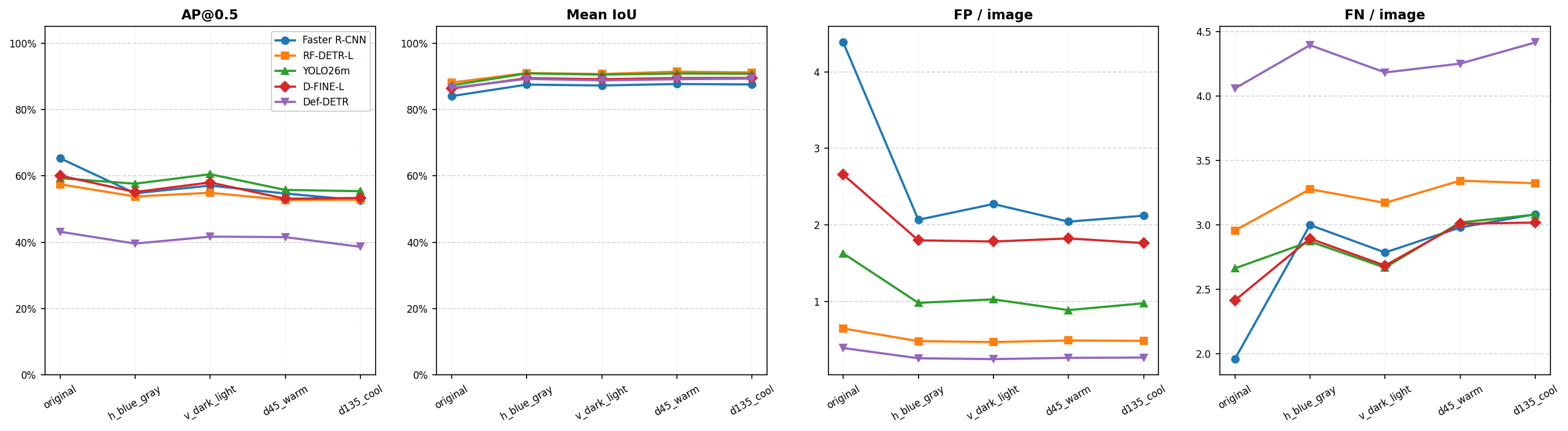}

    \vspace{0.8em}
    \textbf{Low-Frequency Noise}\\
    \includegraphics[width=\textwidth]{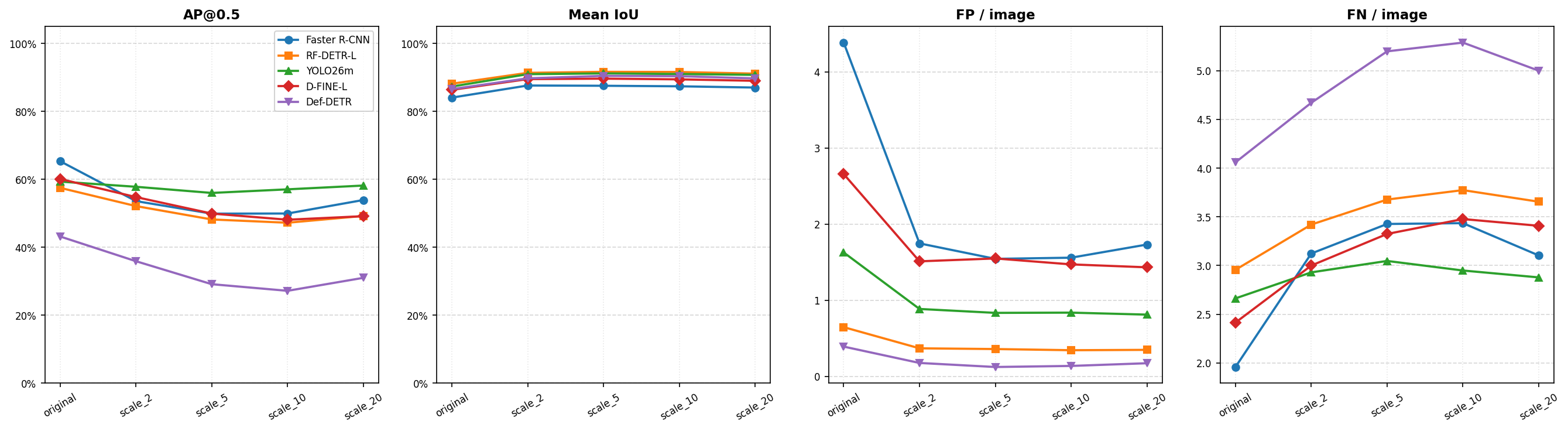}

    \caption{Performance curves under synthetic background replacement.}
    \label{fig:synthetic_bg_curves}
\end{figure*}

Figure~\ref{fig:synthetic_bg_curves} shows the performance curves under background replacement.
Synthetic background manipulations exhibit consistent prediction suppression behavior, with degradation severity depending on the structure of the substituted background. Solid-color backgrounds produce the mildest effect, causing only modest AP degradation and relatively stable IoU, suggesting that detectors can partially ignore highly simplified backgrounds despite reduced contextual support.

Smooth gradients induce slightly stronger degradation, particularly in FN rates, indicating that even weak spatial structure can begin to interfere with candidate formation. Low-frequency noise produces the strongest effect overall, yielding the largest AP reduction and FN increase across models. Unlike solid colors or gradients, low-frequency noise introduces broad spatial patterns and texture-like structure that may compete with or disrupt the contextual statistics learned during training.

Across all synthetic manipulations, localization quality remains comparatively stable while FN increases and prediction volume decreases, further supporting prediction suppression as the dominant failure mode rather than confidence or localization degradation.

\FloatBarrier

\FloatBarrier
\subsubsection{Per-Manipulation Metric Analysis}
Table~\ref{tab:bg_models_mega_transposed} presents the per-model robustness and behavioral metrics under synthetic background manipulations.

\begin{table*}[htbp]
\centering
\caption{Per-model robustness and behavioral metrics under synthetic background manipulations.}
\label{tab:bg_models_mega_transposed}
\small
\setlength{\tabcolsep}{5pt}
\begin{tabular}{llccccc}
\toprule
\textbf{Manipulation} & \textbf{Metric} & \textbf{Deformable DETR} & \textbf{D-FINE} & \textbf{Faster R-CNN} & \textbf{RF-DETR} & \textbf{YOLO26M} \\
\midrule
\multirow{5}{*}{Low Freq Noise}
& AP rAUC      & 0.6812 & 0.8212 & 0.7859 & 0.8419 & 0.9640 \\
& $\Delta$FP/img   & -61.14\% & -43.91\% & -62.48\% & -45.15\% & -48.28\% \\
& $\Delta$FN/img   & +24.15\% & +36.80\% & +67.13\% & +22.86\% & +10.83\% \\
& $\Delta$pred/img & -33.24\% & -27.11\% & -41.52\% & -19.25\% & -17.04\% \\
\midrule
\multirow{5}{*}{Smooth Gradient}
& AP rAUC      & 0.9400 & 0.8988 & 0.8328 & 0.9243 & 0.9515 \\
& $\Delta$FP/img   & -33.97\% & -32.59\% & -51.52\% & -25.59\% & -40.55\% \\
& $\Delta$FN/img   & +6.22\%  & +20.18\% & +51.29\% & +10.92\% & +9.27\% \\
& $\Delta$pred/img & -10.50\% & -17.85\% & -33.42\% & -9.71\%  & -14.39\% \\
\midrule
\multirow{5}{*}{Solid Color}
& AP rAUC      & 0.9234 & 0.9658 & 0.8745 & 0.9481 & 1.0489 \\
& $\Delta$FP/img   & -42.60\% & -35.01\% & -53.10\% & -36.32\% & -40.04\% \\
& $\Delta$FN/img   & +6.47\%  & +10.75\% & +42.56\% & +9.21\%  & -2.73\% \\
& $\Delta$pred/img & -11.70\% & -15.70\% & -32.38\% & -10.08\% & -9.18\% \\
\bottomrule
\end{tabular}
\end{table*}
\FloatBarrier

\subsection{Natural Background Substitution}
\label{sec:appendix_natural_bg_full_dataset}

First, we present the 30 classes selected for the single-object evaluation in the main text in Table~\ref{tab:coco30}

\begin{table}[htbp]
\centering
\caption{The 30 COCO object categories used in the focal NPMI analysis.}
\label{tab:coco30}
\small
\begin{tabular}{clcl}
\toprule
\# & Category & \# & Category \\
\midrule
1  & person        & 16 & cow          \\
2  & bicycle       & 17 & elephant     \\
3  & car           & 18 & zebra        \\
4  & motorcycle    & 19 & giraffe      \\
5  & bus           & 20 & backpack     \\
6  & train         & 21 & umbrella     \\
7  & truck         & 22 & bottle       \\
8  & boat          & 23 & cup          \\
9  & traffic light & 24 & chair        \\
10 & bench         & 25 & couch        \\
11 & bird          & 26 & potted plant \\
12 & cat           & 27 & dining table \\
13 & dog           & 28 & laptop       \\
14 & horse         & 29 & sink         \\
15 & sheep         & 30 & book         \\
\bottomrule
\end{tabular}
\end{table}
\FloatBarrier

\subsubsection{Multi-Object Evaluation}
We evaluate compatibility-driven substitution in a \emph{multi-object setting}, preserving scene (context) structure while measuring the effect on a single, randomly selected focal instance($4{,}952$ images, full 80-class NPMI matrix). Table~\ref{tab:robustness_summary_natural_bg_multi} shows consistent suppression across models: FN increases and prediction volume decreases, while FP remains stable or declines.

\begin{table}[ht]
\centering
\caption{Robustness summary across natural background substitutions—multi-object focal analysis (NPMI-based compatibility, focal object evaluated only).}
\label{tab:robustness_summary_natural_bg_multi}
\footnotesize
\begin{tabular}{lccccc}
\toprule
\textbf{Metric} & \textbf{Def-DETR} & \textbf{D-FINE-L} & \textbf{Faster R-CNN} & \textbf{RF-DETR-L} & \textbf{YOLO26M} \\
\midrule
AP rAUC        & \textbf{0.75} & 0.79 & 0.78 & \textbf{0.75} & 0.92 \\
$\Delta$FN/img & $+27.1\%$ & $+57.0\%$ & $\mathbf{+71.8\%}$ & $+51.4\%$ & $+23.5\%$ \\
$\Delta$FP/img & $-18.0\%$ & $-19.5\%$ & $-23.5\%$ & $-16.9\%$ & $\mathbf{-15.8\%}$ \\
$\Delta$pred/img & $-19.2\%$ & $-18.8\%$ & $\mathbf{-21.7\%}$ & $-18.2\%$ & $-13.2\%$ \\
\bottomrule
\end{tabular}
\par\vspace{4pt}
\begin{minipage}{\linewidth}
\footnotesize
Metrics are averaged over image–background pairs ($N{=}4{,}952 \times 30 = 148{,}560$). Under a conservative bound ($\mathrm{CV}{=}1$), 95\% CI half-widths are $\leq{\pm}2.0$pp for $\Delta$FN/img and $\leq{\pm}0.4$pp for $\Delta$FP/img, far smaller than observed effects, indicating statistical reliability.
\end{minipage}
\end{table}

\FloatBarrier
\begin{figure}[htbp]
\centering
\includegraphics[width=\textwidth]{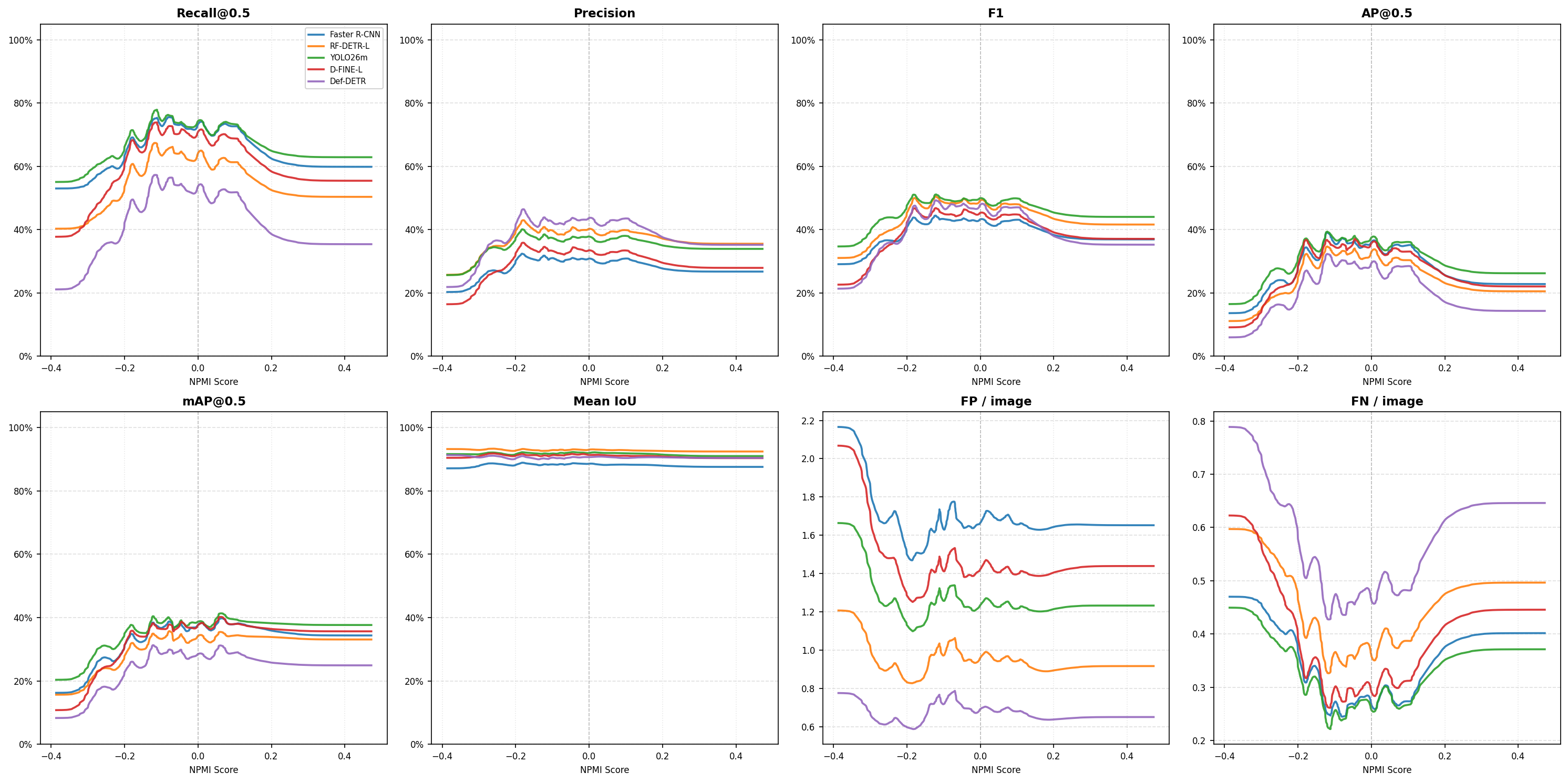}
\caption{Detection performance as a function of object--background compatibility (NPMI) in the multi-object setting.}
\label{fig:npmi_curves}
\end{figure}

\paragraph{Analysis.}
Figure~\ref{fig:npmi_curves} shows a non-monotonic relationship between NPMI-based compatibility and detection performance: performance peaks near $\text{NPMI} \approx 0$, degrades at higher compatibility, and recovers on the original distribution exhibiting the same qualitative suppression pattern as the single-object setting.
\FloatBarrier
\subsubsection{Detection Candidate Analysis}
\label{sec:appendix_candidate_analysis}

We define \emph{prediction candidate existence} as the presence of at least one predicted box
with IoU$\,\geq\,0.5$ to a ground-truth object, independent of confidence thresholding.
This detector-agnostic proxy isolates whether the model produces any viable localization,
abstracting over architecture-specific mechanisms (region proposals, grid predictions, decoder
queries) to enable consistent cross-architecture comparison.
For natural background substitution, existence is computed in the multi-object setting,
restricted to the manipulated object.

\paragraph{Object size definition.}
We follow standard COCO pixel thresholds: small ($<32^2$), medium ($32^2$--$96^2$),
and large ($\geq96^2$), using ground-truth bounding box area.

\FloatBarrier
\begin{figure*}[htbp]
\centering
\setlength{\tabcolsep}{3pt}
\begin{tabular}{cc}
\includegraphics[width=0.48\textwidth]{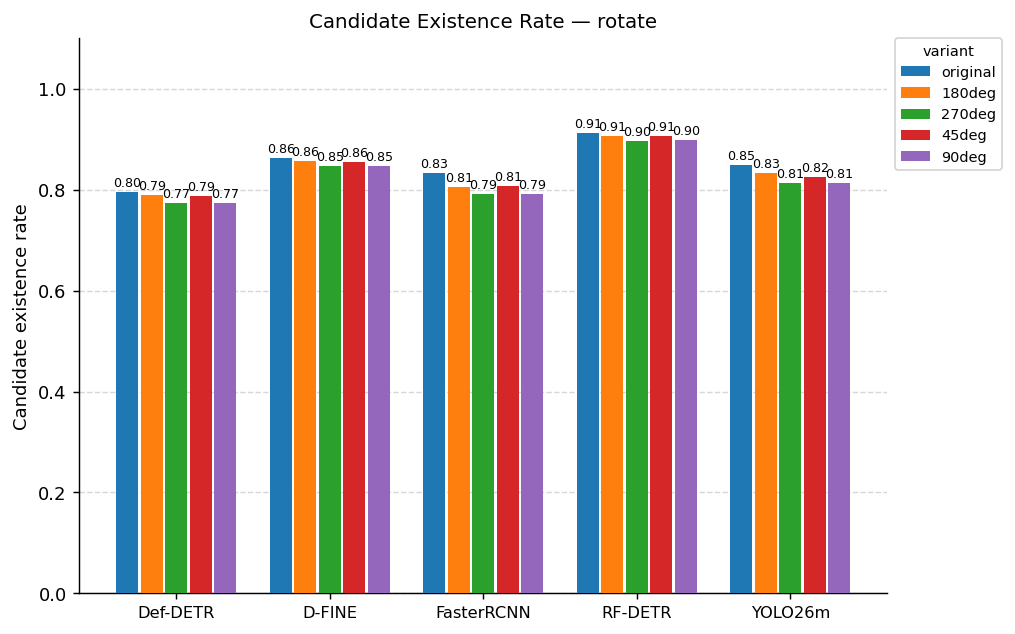} &
\includegraphics[width=0.48\textwidth]{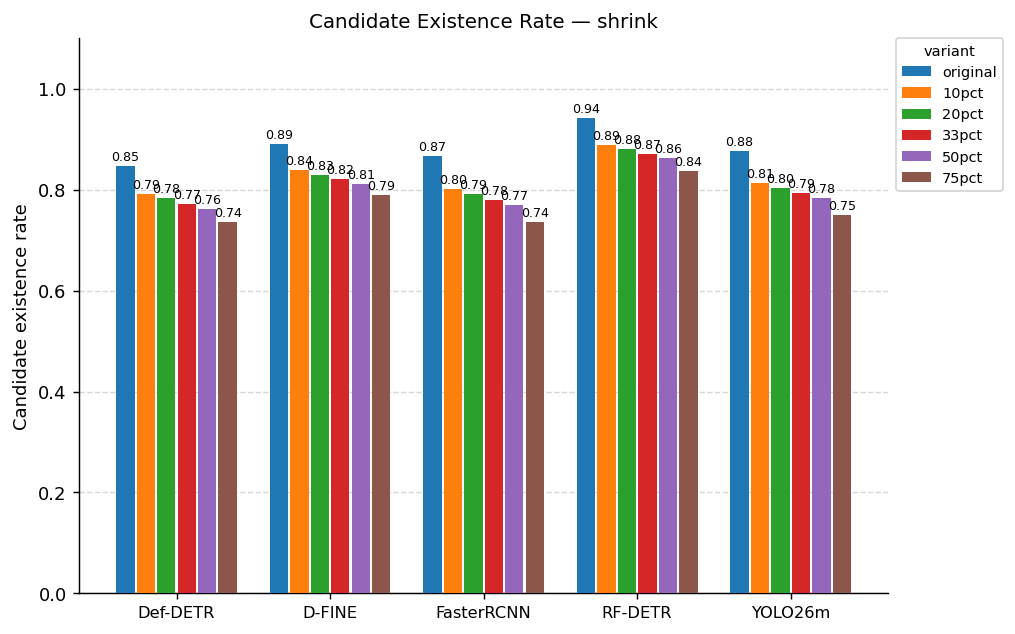} \\
\includegraphics[width=0.48\textwidth]{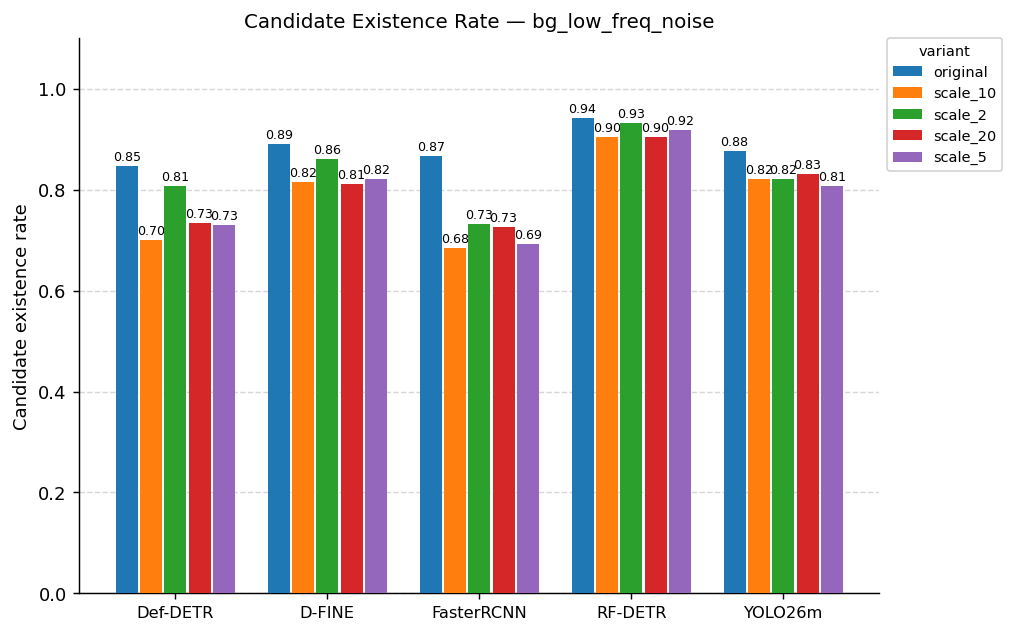} &
\includegraphics[width=0.48\textwidth]{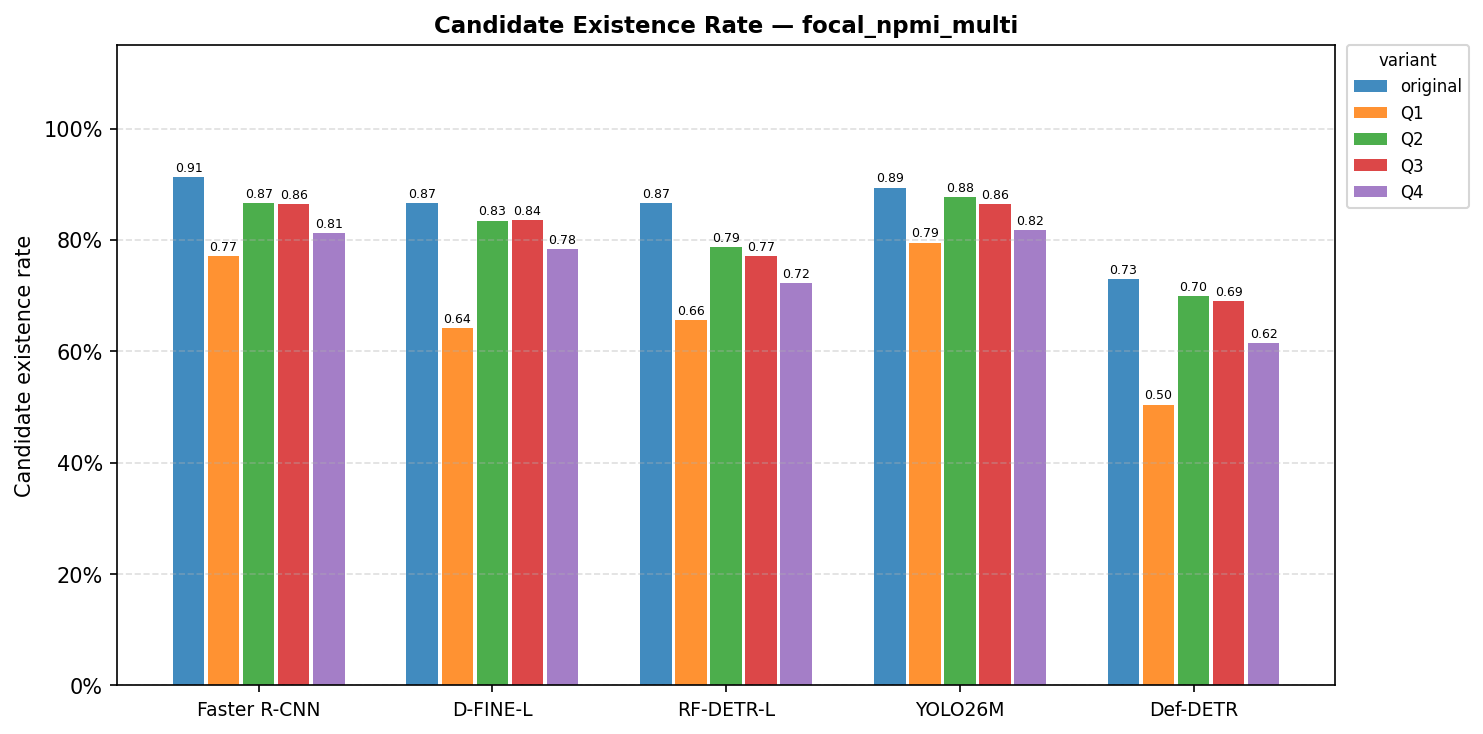}
\end{tabular}
\caption{Prediction candidate existence rate across representative manipulations.}
\label{fig:candidate_existence_all_manips}
\end{figure*}

\begin{figure*}[htbp]
\centering
\begin{tabular}{c}
\includegraphics[width=\textwidth]{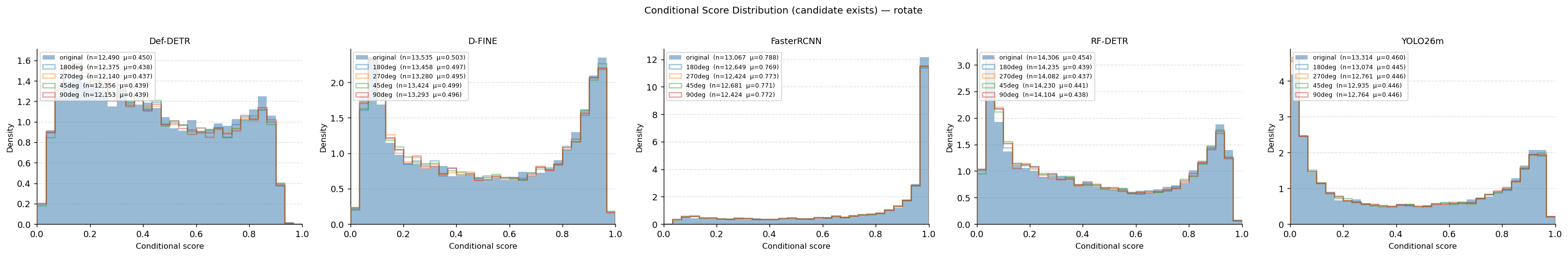} \\
\includegraphics[width=\textwidth]{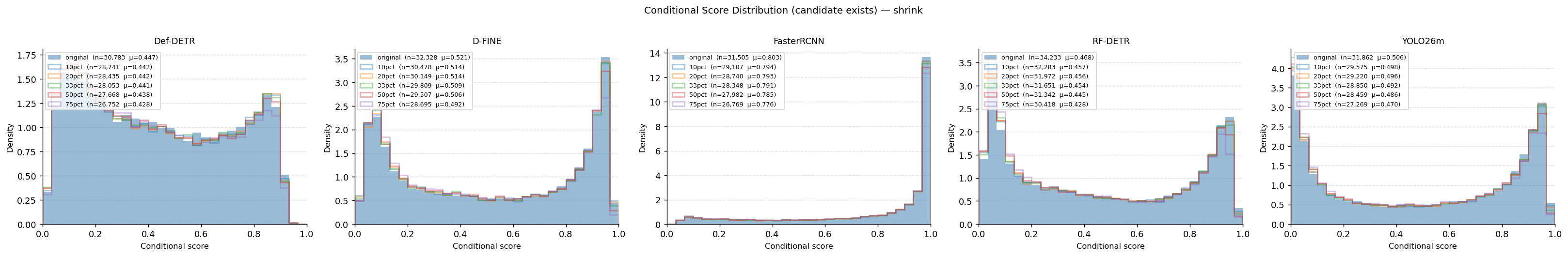} \\
\includegraphics[width=\textwidth]{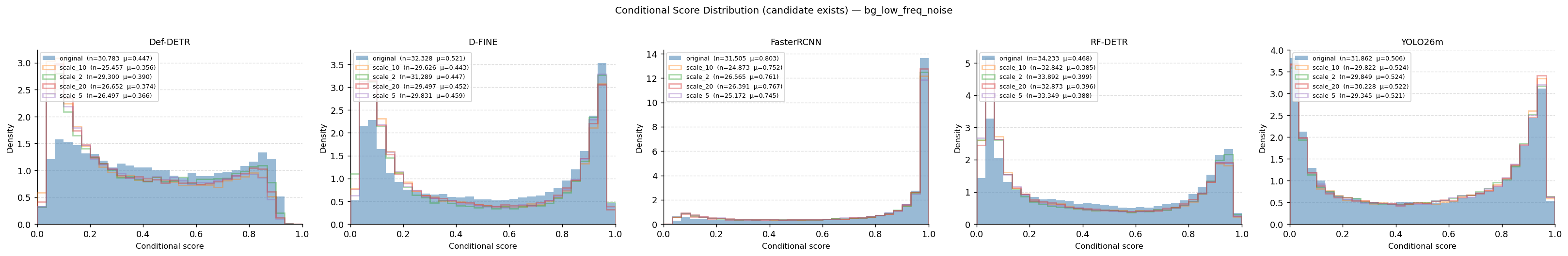} \\
\includegraphics[width=\textwidth]{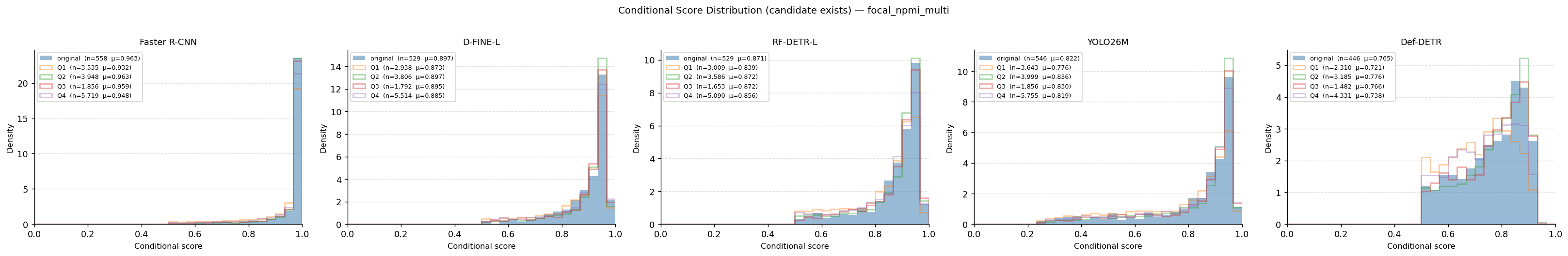}
\end{tabular}
\caption{Conditional score distributions given prediction candidate existence.
  Distributions remain largely stable across manipulations, indicating that
  confidence is preserved once a candidate is formed.}
\label{fig:conditional_score_all_manips}
\end{figure*}
\FloatBarrier

\paragraph{Aggregate Analysis.}
Tables~\ref{tab:geometric}--\ref{tab:backgrounds} and Table~\ref{tab:npmi_quintiles_focal}
(main text) summarize candidate behavior across manipulation families.
$\Delta\mathcal{E}$ is the mean percentage change in existence rate relative to the native
baseline; $\Delta\mathcal{S}$ is the mean percentage change in conditional score---the maximum
candidate confidence---restricted to instances where a candidate exists in both conditions.
\emph{Worst-case} $\Delta\mathcal{E}$ reports the largest single existence drop across all
model--variant pairs; \emph{small-object} $\Delta\mathcal{E}$ restricts to boxes with area
$<32^2$\,px.

\textbf{Geometric manipulations} (Table~\ref{tab:geometric}).
Shrinkage dominates: existence drops $-7.7\%$ on average versus only $-1.2\%$ in score,
with a pronounced small-object penalty of $-12.6\%$---consistent with the loss of
low-resolution signal rather than a change in classifier confidence.
Rotation induces moderate but balanced degradation ($-1.6\%$ existence, $-1.4\%$ score),
while translation and enlargement have negligible impact on both metrics.

\begin{table}[htbp]
\centering
\caption{Effect of geometric context shifts on candidate formation across five detectors.}
\label{tab:geometric}
\small
\begin{tabular}{lcccc}
\toprule
Manipulation & $\Delta\mathcal{E}$ & $\Delta\mathcal{S}$ &
  Worst-case $\Delta\mathcal{E}$ & Small-object $\Delta\mathcal{E}$ \\
\midrule
Shrinkage   & $-7.7\%$ & $-1.2\%$ & $-11.9\%$ & $-12.6\%$ \\
Rotation    & $-1.6\%$ & $-1.4\%$ & $-3.6\%$  & $-3.7\%$  \\
Translation & $-0.9\%$ & $-0.2\%$ & $-2.0\%$  & $-2.5\%$  \\
Enlargement & $-0.2\%$ & $-0.2\%$ & $-0.7\%$  & $-0.9\%$  \\
\bottomrule
\end{tabular}
\end{table}
\FloatBarrier

\textbf{Synthetic backgrounds} (Table~\ref{tab:backgrounds}).
Low-frequency noise produces the steepest degradation ($-8.4\%$ existence,
$-6.3\%$ score), with a severe small-object penalty ($-22.0\%$).
Unlike geometric shifts, existence and score degrade at comparable magnitudes across all
three background types, indicating that synthetic textures corrupt both candidate formation
and candidate quality.

\begin{table}[htbp]
\centering
\caption{Effect of synthetic background substitutions on candidate formation.}
\label{tab:backgrounds}
\small
\begin{tabular}{lcccc}
\toprule
Background & $\Delta\mathcal{E}$ & $\Delta\mathcal{S}$ &
  Worst-case $\Delta\mathcal{E}$ & Small-object $\Delta\mathcal{E}$ \\
\midrule
Low-freq.\ noise & $-8.4\%$ & $-6.3\%$ & $-18.7\%$ & $-22.0\%$ \\
Smooth gradient  & $-4.7\%$ & $-4.1\%$ & $-14.2\%$ & $-9.7\%$  \\
Solid color      & $-2.8\%$ & $-3.0\%$ & $-14.7\%$ & $-9.1\%$  \\
\bottomrule
\end{tabular}
\end{table}

\textbf{Natural backgrounds} (Table~\ref{tab:npmi_quintiles_focal}).
Incompatible contexts (Q1) produce the largest existence drop ($-22.5\%$) with modest score
degradation ($-3.3\%$), confirming that semantic incompatibility primarily suppresses
candidate formation.
Degradation partially recovers through Q2--Q3, reaching its minimum at Q3 ($-3.6\%$
existence, $+0.84\%$ score).
This trend is non-monotonic: high-compatibility contexts (Q4--Q5) exhibit renewed
suppression ($-9.9\%$ at Q5), suggesting that both extreme incompatibility and
over-familiarity disrupt detection.
Across all quintiles, existence drops consistently exceed score changes, ruling out
confidence collapse as the primary failure mode.

\paragraph{Summary.}
Across all manipulation families, missed detections are driven predominantly by failure to
form valid candidates rather than by degraded confidence in existing ones
(Figures~\ref{fig:candidate_existence_all_manips}--\ref{fig:conditional_score_all_manips}).
The dominant mechanism is manipulation-specific: shrinkage suppresses formation through
signal loss; synthetic backgrounds degrade both formation and quality; natural backgrounds
produce the largest existence drops under semantically extreme contexts.

\FloatBarrier
\subsubsection{Confidence Sweeps}
\label{sec:appendix_confidence_sweeps}

We examine confidence behavior through two complementary views: the maximum prediction score
assigned to each ground-truth object, and recall recovery under decreasing confidence
thresholds.
For natural background substitutions, results are stratified by NPMI quintile (Q1–Q5,
least to most compatible, equal-sized bins).

\FloatBarrier
\begin{figure*}[htbp]
\centering
\includegraphics[width=\textwidth]{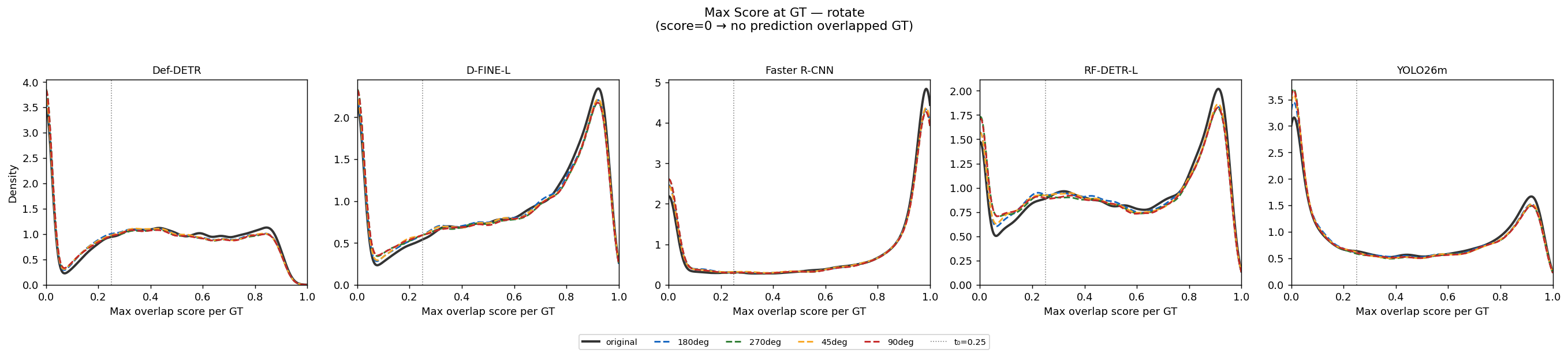}
\includegraphics[width=\textwidth]{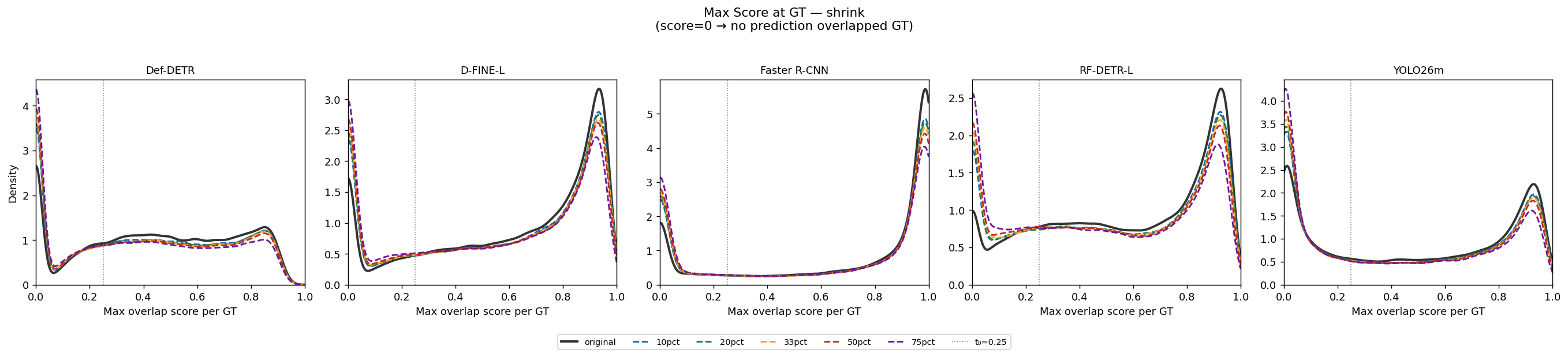}
\includegraphics[width=\textwidth]{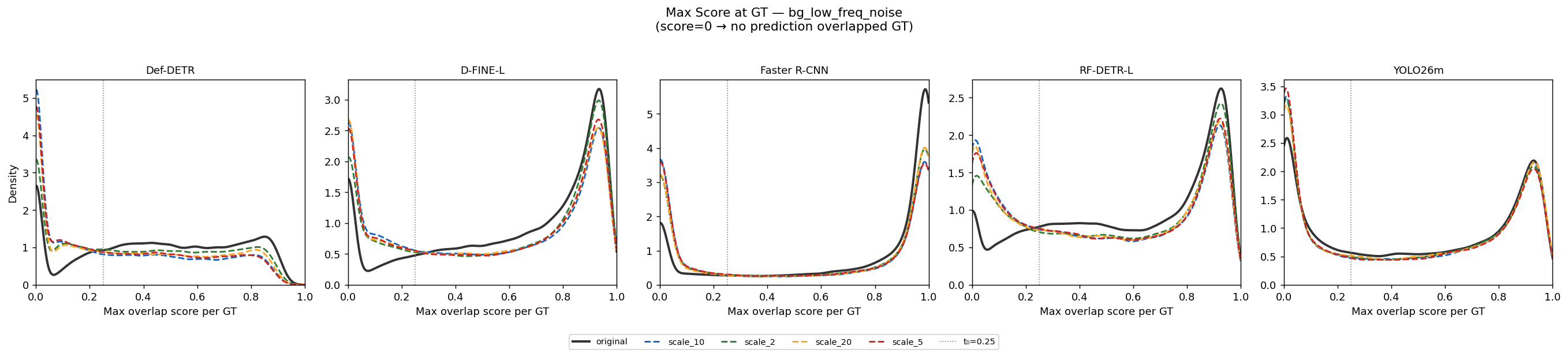}
\includegraphics[width=\textwidth]{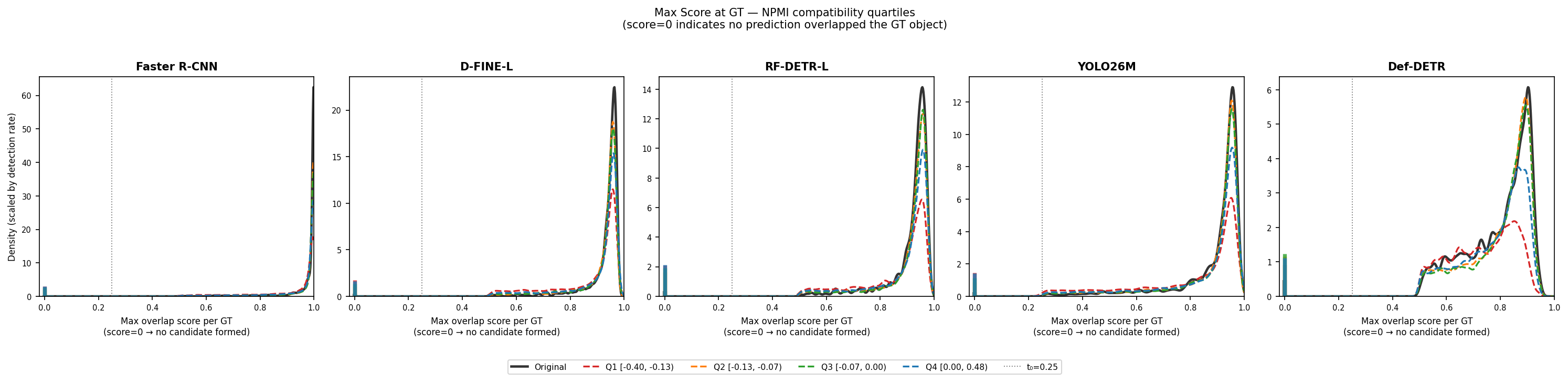}
\caption{Distribution of maximum prediction score per ground-truth object. A score of 0
  indicates no prediction sufficiently overlaps the ground-truth object. Aggregate statistics
  are reported in the tables.}
\label{fig:max_score_all_manips}
\end{figure*}

Figure~\ref{fig:max_score_all_manips} shows a consistent pattern across all manipulation
types and models: under context shift, mass in the distribution migrates directly to zero
rather than redistributing toward intermediate confidence values.
Objects that receive any prediction retain near-identical high-confidence scores to the
clean condition; the manipulation effect is concentrated entirely in the zero bin,
representing objects for which no sufficiently overlapping prediction is produced at all.

\FloatBarrier
\begin{figure*}[htbp]
\centering
\includegraphics[width=\textwidth]{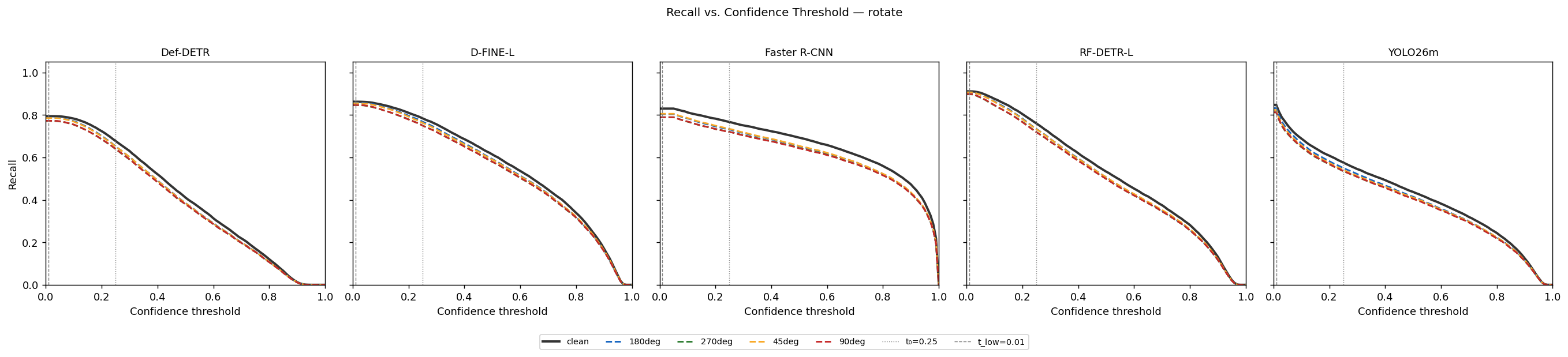}
\includegraphics[width=\textwidth]{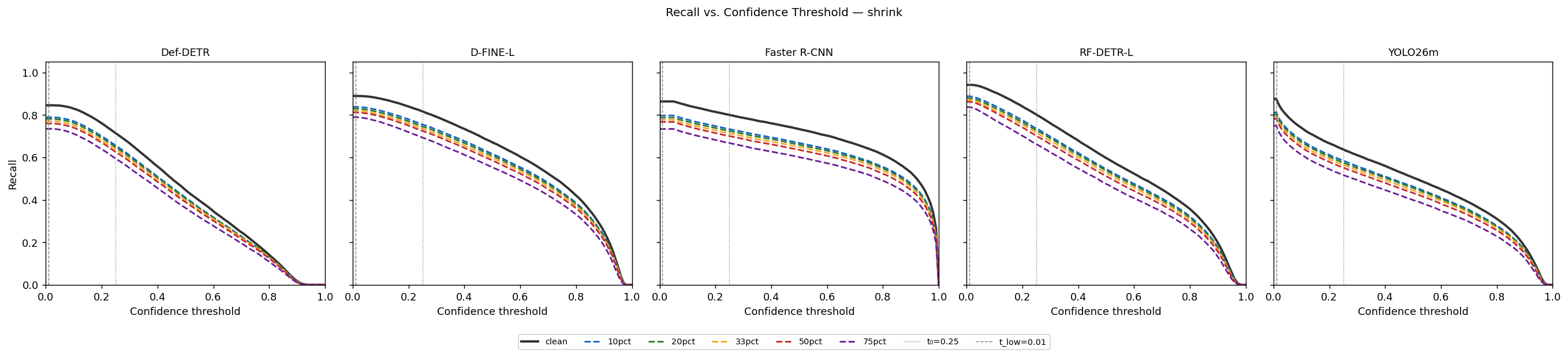}
\includegraphics[width=\textwidth]{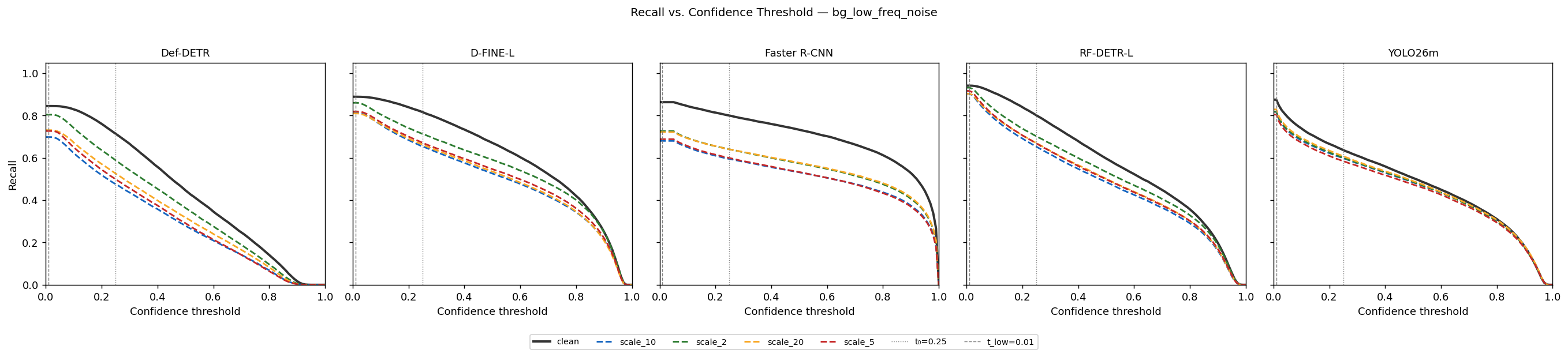}
\includegraphics[width=\textwidth]{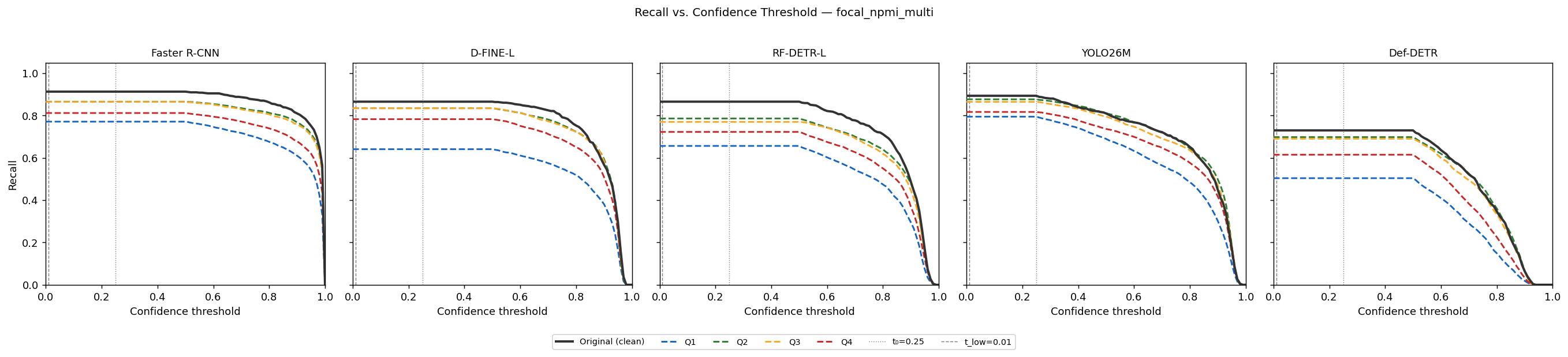}
\caption{Recoverable recall as a function of confidence threshold. Shaded regions indicate
  the recall gap recoverable by lowering the threshold. Aggregate behavior is reported in
  the tables.}
\label{fig:recoverable_recall_all_manips}
\end{figure*}

Figure~\ref{fig:recoverable_recall_all_manips} decomposes the recall gap into two
components.
Lowering the threshold recovers some missed detections, confirming that some fraction of
errors correspond to low-confidence but existing predictions.
However, a substantial gap persists even at near-zero thresholds: these are irreversible
misses attributable to complete prediction suppression, meaning - no prediction candidate was formed whatsoever.
Together, the two figures establish that context shift operates primarily through suppression
of candidate formation, with confidence degradation playing only a secondary role.

\FloatBarrier
\subsubsection{Robustness to Compatibility Definition}
\label{sec:compatibility_robustness}

To assess whether our findings depend on the specific choice of compatibility formulation,
we evaluate two alternative measures: a visually grounded baseline using raw co-occurrence
frequency, and a non-visual control using character-level string similarity.

\paragraph{Image-frequency compatibility.}
We define compatibility as the empirical co-occurrence frequency $P(o, c)$ between object
category $o$ and context label $c$, estimated directly from COCO caption co-occurrence
counts.
Unlike NPMI, this formulation does not normalize for marginal frequencies and is therefore
more sensitive to dataset bias.
However, because $P(o, c)$ and NPMI share the same underlying co-occurrence statistics,
they are monotonically related for fixed marginals: high co-occurrence frequency implies
high NPMI.
This relationship explains why image-frequency compatibility retains some structural signal.

\paragraph{Character $n$-gram surface similarity (non-visual baseline).}
As a deliberate non-visual control, we use character $n$-gram cosine similarity between
the object-in-context prompt and the object-only prompt:
\[
\kappa_{\text{ngram}}(o, c') =
  \cos\!\bigl(\phi(\text{``a photo of a } o \text{ in a } c'\text{''}),\;
              \phi(\text{``a photo of a } o\text{''})\bigr),
\]
where $\phi(\cdot)$ maps a string to its $n$-gram count vector
(\texttt{char\_wb} analyser, $n \in \{2,3,4\}$).
The score depends solely on string overlap, with no grounding in visual data or
object--background relationships.

\paragraph{Evaluation.}
We repeat the natural background substitution experiment using each score to construct
equal-frequency bins, following the same pipeline as the main experiments.

\paragraph{Results.}

\begin{figure*}[htbp]
  \centering
  \includegraphics[width=\textwidth]{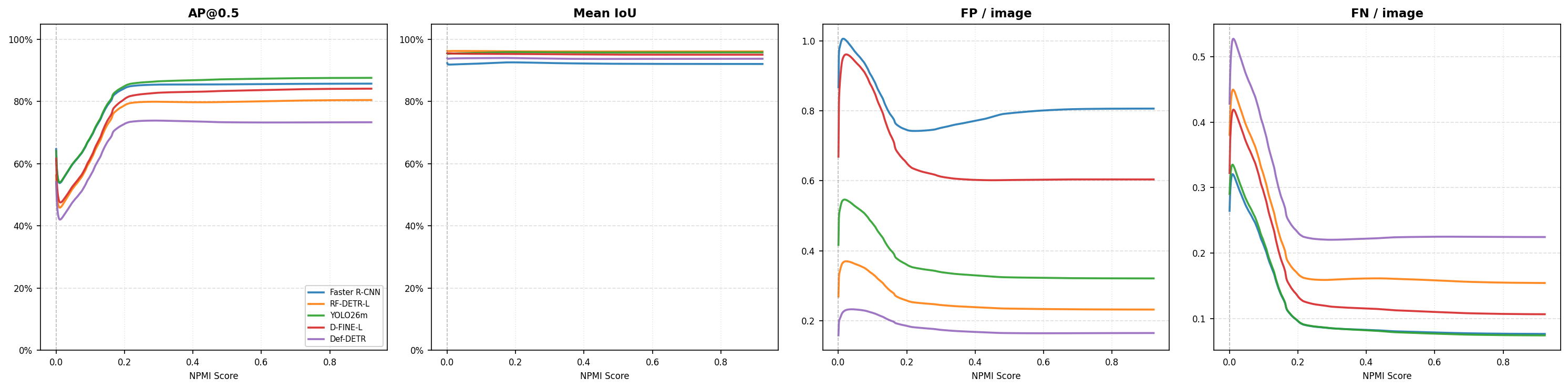}
  \caption{Performance as a function of image-frequency compatibility.
    Curves exhibit a saturation shape: structured degradation is concentrated at
    very low frequency, with performance plateauing beyond $\approx 0.2$.}
  \label{fig:imgfreq_comp_curves}
\end{figure*}

\textbf{Image-frequency compatibility} (Figure~\ref{fig:imgfreq_comp_curves}) produces
structured curves, but with a characteristic saturation shape: AP rises sharply from
frequency~$\approx 0$ to $\approx 0.2$, then plateaus, with FP and FN both declining
over the same interval.
This shape reflects the normalization gap between $P(o,c)$ and NPMI.
At very low frequency, both scores agree that context is incompatible, and performance
degrades accordingly.
In the mid-to-high frequency regime, however, $P(o, c)$ saturates and can no longer
discriminate among contexts that differ meaningfully in normalized compatibility.
Unlike the NPMI curves, FP and FN move together rather than diverging, and the model
ordering is compressed once the saturation point is reached.
Despite this coarser resolution, the suppression effect is consistent: across models,
mean $\Delta$FN/img increases by up to $+23\%$ and mean $\Delta$pred/img decreases by
up to $-39\%$ (Table~\ref{tab:imgfreq_ngram_suppression}).

\begin{figure*}[htbp]
  \centering
  \includegraphics[width=\textwidth]{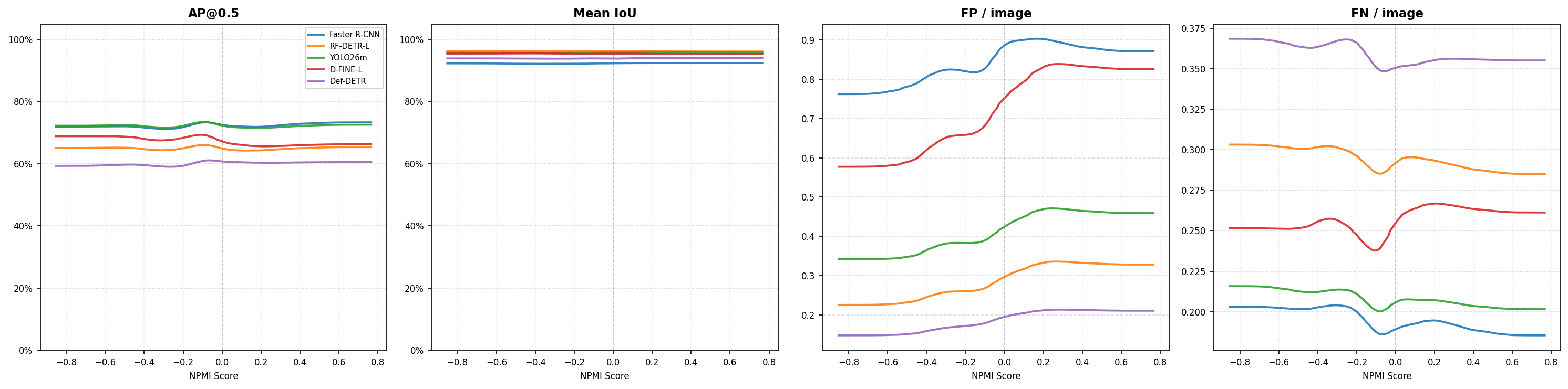}
  \caption{Performance as a function of character $n$-gram surface compatibility.
    Curves are flat and unordered across all metrics, confirming that string-level
    similarity carries no visual compatibility signal.}
  \label{fig:ngram_comp_curves}
\end{figure*}

\FloatBarrier
\textbf{$n$-gram baseline} (Figure~\ref{fig:ngram_comp_curves}) produces flat, unordered
curves across all four metrics.
AP and FN show no monotonic trend; FP exhibits a weak upward drift with increasing
similarity — the opposite of what a meaningful compatibility signal would predict.
The score contains no information about visual detectability, and bins constructed from
it are effectively random context partitions.

Despite this absence of structure, aggregate suppression persists: mean $\Delta$FN/img
reaches $+126\%$ and mean $\Delta$pred/img reaches $-43\%$
(Table~\ref{tab:imgfreq_ngram_suppression}).
These numbers reflect the background swap itself, not any ordering imposed by the score.

\FloatBarrier
\begin{table}[htbp]
\centering
\caption{Mean $\Delta$FN/image and $\Delta$pred/image under image-frequency and $n$-gram
  compatibility, averaged over bins relative to the original-image baseline.
  Negative $\Delta$pred/image reflects reduced prediction volume.
  \textbf{Bold}: worst case per column; \underline{underline}: least suppressed.}
\label{tab:imgfreq_ngram_suppression}
\small
\setlength{\tabcolsep}{5pt}
\begin{tabular}{lcccc}
\toprule
 & \multicolumn{2}{c}{Image-frequency} & \multicolumn{2}{c}{$n$-gram baseline} \\
\cmidrule(lr){2-3}\cmidrule(lr){4-5}
Model & $\Delta$FN/img & $\Delta$pred/img & $\Delta$FN/img & $\Delta$pred/img \\
\midrule
Faster R-CNN  & $+19.0\%$ & $-38.0\%$ & $\mathbf{+125.8\%}$ & $-39.9\%$ \\
RF-DETR-L     & $\mathbf{+23.0\%}$ & $\mathbf{-39.2\%}$ & $+94.8\%$ & $\mathbf{-43.2\%}$ \\
D-FINE-L      & $-11.0\%$ & $-36.5\%$ & $+62.5\%$ & $-37.5\%$ \\
YOLO26m       & $-14.9\%$ & $-38.3\%$ & $+70.7\%$ & $-40.2\%$ \\
Def-DETR      & \underline{$-2.2\%$} & \underline{$-31.9\%$} & \underline{$+40.0\%$} & \underline{$-38.9\%$} \\
\bottomrule
\end{tabular}
\end{table}

\paragraph{Discussion.}
Two conclusions follow from these experiments.

First, prediction suppression under background substitution is robust to the choice of
compatibility formulation: both measures—one structured, one random with respect to
visual context—produce consistent aggregate suppression, confirming the effect is a
property of the substitution itself.

Second, revealing the \emph{structured dependence} of suppression on context requires a
compatibility measure grounded in real object--background co-occurrence.
Image-frequency compatibility, by sharing the same co-occurrence statistics as NPMI,
preserves a coarse compatibility axis, but its unnormalized form saturates early and
conflates high-frequency contexts that differ meaningfully in normalized compatibility.
Character $n$-gram similarity, lacking any visual grounding, imposes no meaningful
ordering and produces curves indistinguishable from random context assignment.
NPMI provides the finest-grained compatibility axis by normalizing for marginal
frequencies, enabling the smooth, monotonic structure observed in the main results.

\FloatBarrier
\subsubsection{Ablation: Blending and composition Parameters}
\label{sec:results_blending_ablation}

To verify that observed robustness effects are attributable to model behaviour rather than blending artifacts, we conduct an ablation over five pipeline configurations, progressively enabling stages from raw copy-paste to aggressive harmonisation. All runs use 1,079 single-object source images with 30 backgrounds each (32,370 pairs), with the NPMI compatibility matrix as the scoring function. Table~\ref{tab:blending_ablation} lists the parameter values for each configuration. This ablation was performed using the single-object setting in order to maximize isolation of the blending effects on the evaluated object.

\begin{table}[htbp]
  \caption{Blending pipeline configurations used in the ablation study. Each row adds or strengthens one or more processing stages relative to the previous configuration.}
  \label{tab:blending_ablation}
  \centering
  \small
  \begin{tabular*}{\columnwidth}{@{\extracolsep{\fill}} lcccccc @{}}
    \toprule
    Configuration
      & \texttt{decontam\_px}
      & \texttt{feather\_px}
      & \texttt{lum\_match}
      & \texttt{lum\_max\_shift}
      & \texttt{blur\_sigma} \\
    \midrule
    C1 --- Raw copy-paste   & 0 & 0 & $\times$   & ---  & 0.0 \\
    C2 --- +Decontamination & 1 & 0 & $\times$   & ---  & 0.0 \\
    C3 --- +Feathering      & 1 & 2 & $\times$   & ---  & 0.0 \\
    C4 --- Default pipeline & 1 & 2 & \checkmark & 0.15 & 0.5 \\
    C5 --- Aggressive       & 2 & 4 & \checkmark & 0.30 & 1.5 \\
    \bottomrule
  \end{tabular*}
\end{table}

\begin{figure*}[htbp]
    \small
    \centering

    \textbf{Raw Copy-Paste}\\
    \includegraphics[width=\textwidth]{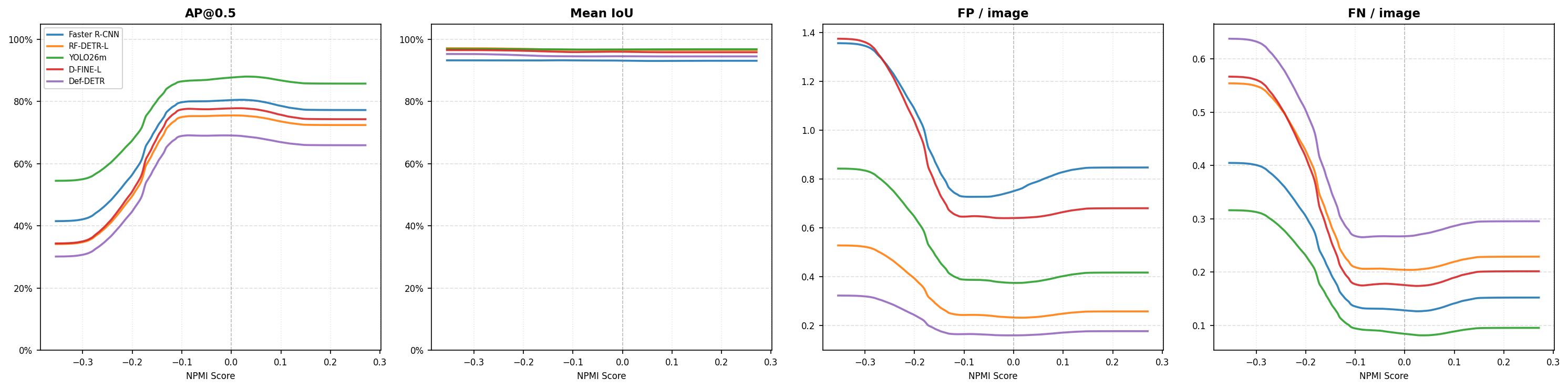}

    \vspace{0.8em}
    \textbf{Decontamination Only}\\
    \includegraphics[width=\textwidth]{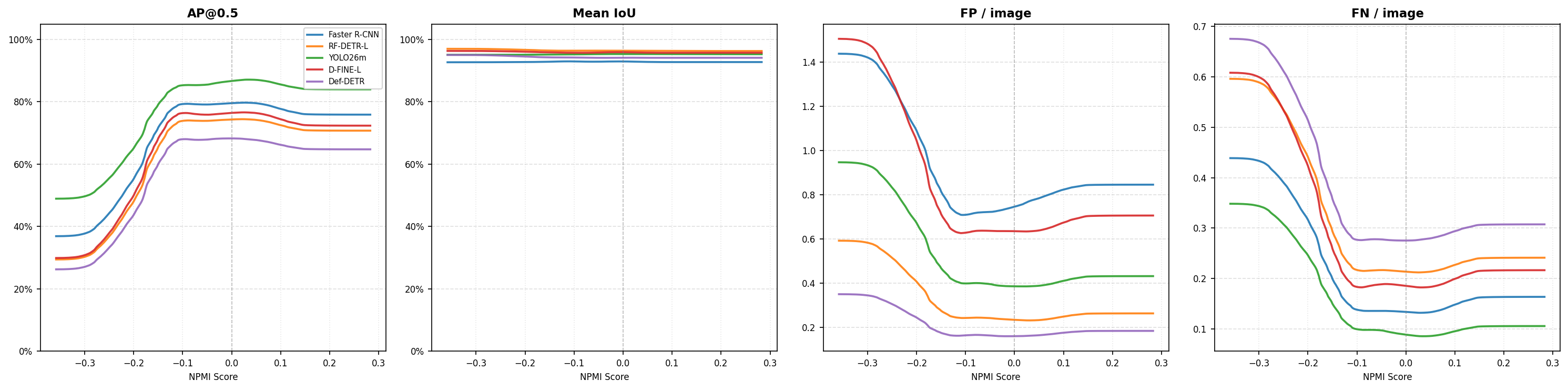}

    \vspace{0.8em}
    \textbf{Decontamination + Feathering}\\
    \includegraphics[width=\textwidth]{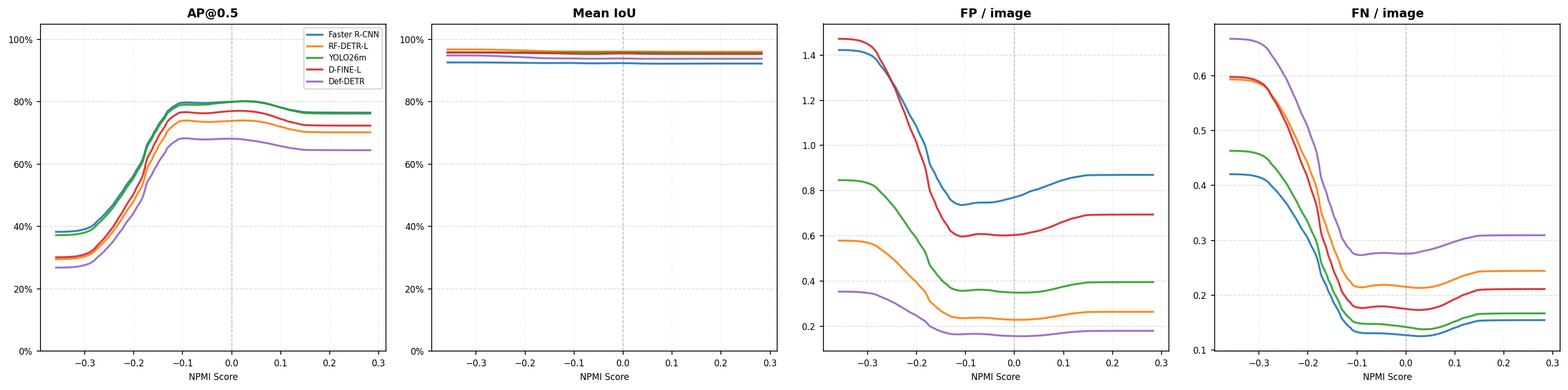}

    \vspace{0.8em}
    \textbf{Default Harmonisation}\\
    \includegraphics[width=\textwidth]{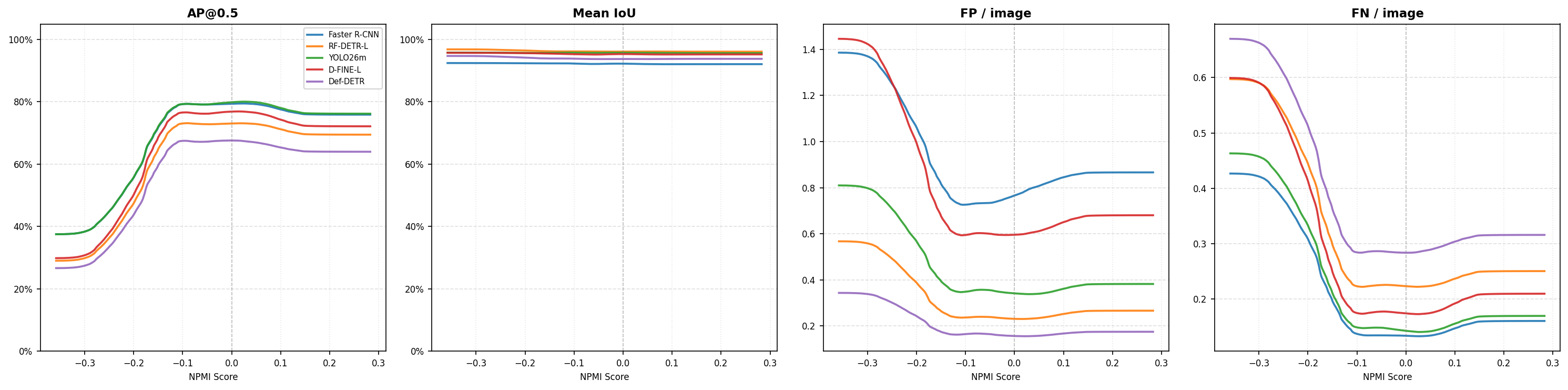}

    \vspace{0.8em}
    \textbf{Aggressive Harmonisation}\\
    \includegraphics[width=\textwidth]{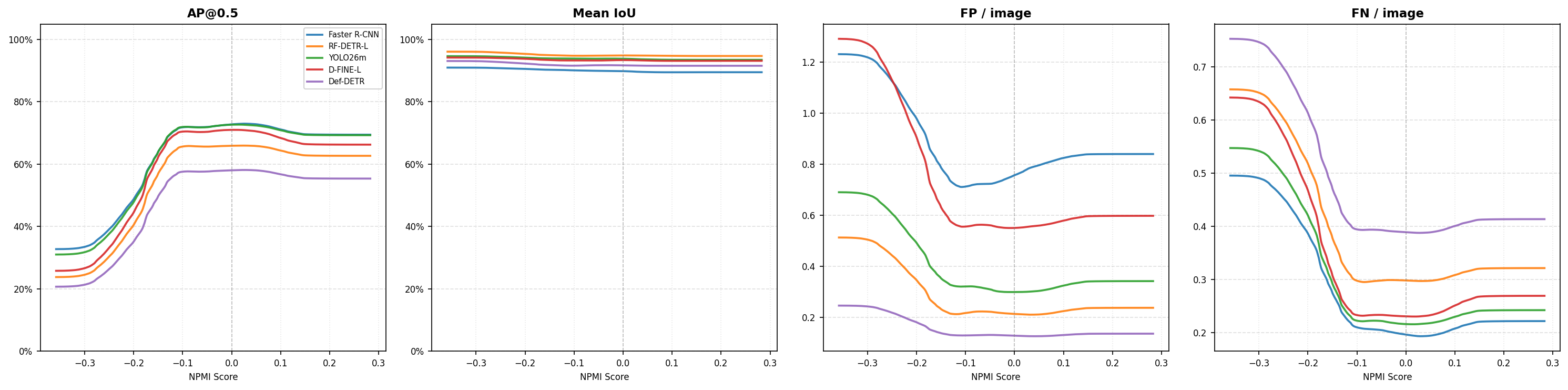}

    \caption{Performance curves under compatibility-driven natural background replacement across five composition settings, ordered from weakest to strongest blending: raw copy-paste, decontamination only, decontamination with feathering, default harmonisation, and aggressive harmonisation. Across settings, compatibility-dependent suppression remains qualitatively consistent, indicating that the effect is not specific to a particular blending configuration.}
    \label{fig:blend_curves_all}
\end{figure*}

\paragraph{Results Across Blending Configurations}
Figure~\ref{fig:blend_curves_all} presents performance curves under natural background replacement with different blending parameters. Across all configurations, the relationship between NPMI compatibility and performance remains qualitatively consistent. Lower compatibility reduces recall, increases false negatives, and decreases total predictions, while false positives remain stable or decrease. Although degradation magnitude varies with blending strength, the overall curve structure is preserved, suggesting that the observed effects are not driven by a specific composition configuration.

\FloatBarrier

\FloatBarrier
\section{Mitigation}
\label{sec:appendix_mitigation}

\subsection{Training Setup}
\label{sec:appendix_mitigation_training}
Across all mitigation runs, we re-train YOLO26M, with the same weights and configuration used in the evaluation pipeline \ref{sec:appendix_evaluation_protocol}, on a fixed split of COCO~train2017 (118k images), concatenated with one augmented slice. Augmented slices are produced by applying a single manipulation family to a subset of train2017 instances. For geometric augmentations, parameter values cycle round-robin across the configured severity grid (shrink/enlarge: $\{10,20,33,50,75\}\%$; rotate: $\{45,90,180,270\}^\circ$; translate: $\{5,10,20,40\}\%$ of image dimension), and the LaMa inpainter fills any uncovered pixels. For bg\_swap, single-object compositions onto Places365 backgrounds are sampled from the same compatibility distribution used in evaluation. Synthetic-background augmentations (\emph{solid}, \emph{gradient}, \emph{noise}) replace background pixels outside polygon segmentations with the corresponding synthetic field; YOLO labels are inherited from the source COCO image. Each augmented slice contains $\sim$11k images ($\sim$9\% of the train set) and is held fixed across runs to remove sampling variance from the comparison. We train each variant for 40 epochs, batch size 64, AMP fp16, with hyperparameters identical to the COCO-only \emph{baseline} retrain.

Critically, the held-out evaluation split (990 COCO val2017 images) is disjoint from \emph{any} image used as a source for training composites, and the scene splits used to source bg\_swap backgrounds are disjoint from the scene splits used at evaluation, ruling out leakage between train and eval composites.

\subsection{Clean-Image Behavior}
\label{sec:appendix_mitigation_clean}
Table~\ref{tab:mitigation_clean_fn} reports clean-image behavior on the 990-image held-out
eval split.
All augmented variants improve over the baseline on every metric: AP@0.5 rises by
$1.1\%$--$2.7\%$, FN/img falls by $-9\%$ to $-22\%$, and prediction volume increases
by $+22\%$ to $+36\%$.
The best overall variant is \emph{aug\_noise} (AP $0.904$, $\Delta$FN $-22.0\%$);
the largest prediction-volume gain comes from \emph{aug\_bg\_swap} ($+36.1\%$).
Background augmentations match or exceed geometric augmentations across all three metrics.

\begin{table}[htbp]
\centering
\small
\caption{Clean-image behavior on the 990-image held-out eval split.}
\label{tab:mitigation_clean_fn}
\begin{tabular}{lcccc}
\toprule
Model        & AP@0.5 & FN/img & $\Delta$FN vs.\ baseline & $\Delta$Preds/img \\
\midrule
baseline     & 0.877 & 0.123 &       \\
aug\_bg\_swap & 0.899 & 0.101 & $-17.7\%$ & $\mathbf{+36.1\%}$ \\
aug\_shrink  & 0.903 & 0.097 & $-21.1\%$ & $+22.3\%$ \\
aug\_enlarge & 0.888 & 0.112 & $-8.9\%$ & $+26.1\%$ \\
aug\_rotate  & 0.899 & 0.101 & $-17.7\%$ & $+29.2\%$\\
aug\_translate  & 0.900 & 0.100 & $-18.7\%$ & $+23.3\%$ \\
\midrule
aug\_solid    & 0.895 & 0.105 & $-14.6\%$ & $+21.9\%$ \\
aug\_gradient & 0.900 & 0.100 & $-18.7\%$ & $+23.1\%$ \\
aug\_noise    & \textbf{0.904} & \textbf{0.096} & $\mathbf{-22.0\%}$ & $+23.0\%$ \\
\bottomrule
\end{tabular}
\end{table}

\subsection{Cross-Shift Robustness Matrix}
\label{sec:appendix_mitigation_cross_shift}
Table~\ref{tab:mitigation_cross_shift} reports per-shift AP rAUC for every (training augmentation, evaluation context-shift) pair. 

\begin{table}[htbp]
\centering
\small
\caption{Per-shift AP rAUC. Rows: training augmentation. Columns: evaluation dataset (context-shifted). Bold = best column, Underlined = second best.}
\label{tab:mitigation_cross_shift}
\begin{tabular}{lccccc|c}
\toprule
Model        & bg\_swap & shrink & enlarge & rotate & translate & mean \\
\midrule
baseline     & 0.346 & 0.815 & \textbf{1.098} & 0.594 & 0.981 & $ 0.767_{\pm{0.302}}$ \\
aug\_bg\_swap& \textbf{0.385} & 0.849 & 1.051 & 0.635 & 0.983 & $ 0.781_{\pm{0.272}}$ \\
aug\_shrink  & 0.356 & \textbf{0.935} & 1.015 & 0.619 & 0.975 & $ 0.776_{\pm{0.284}}$ \\
aug\_enlarge & 0.356 & 0.822 & 1.071 & 0.635 & 0.955 & $ 0.768_{\pm{0.282}}$ \\
aug\_rotate  & 0.348 & 0.831 & 1.057 & \textbf{0.823} & \textbf{1.021} & \textbf{0.816$_{\pm{0.283}}$} \\
aug\_translate  & 0.352 & 0.842 & 1.059 & 0.655 & 1.011 & $ 0.784_{\pm{0.289}}$ \\
\midrule
aug\_solid    & 0.355 & 0.843 & 1.013 & 0.615 & 0.985 & $ 0.762_{\pm{0.277}}$ \\
aug\_gradient & 0.358 & 0.839 & 1.028 & 0.598 & 0.972 & $ 0.759_{\pm{0.279}}$ \\
aug\_noise    & 0.355 & 0.827 & 1.052 & 0.629 & 0.968 & $ 0.766_{\pm{0.280}}$ \\
\bottomrule
\end{tabular}
\end{table}

\FloatBarrier
\subsection{Held-Out Transfer (Leave-Home-Turf-Out)}
\label{sec:appendix_mitigation_transfer}
To isolate performance improvement from matched-context-shift wins, we average per-shift rAUC for each augmented variant over the four shifts that are \emph{not} its matched manipulation (Table~\ref{tab:mitigation_transfer}). \emph{aug\_bg\_swap} is the strongest cross-shift generalizer ($+11.3$ pp held-out vs.\ baseline), suggesting that compatibility-driven background substitution exposes the detector to the broadest distribution of object--context decouplings. Geometric augmentations help mostly on their matched shift and degrade on average elsewhere.

\begin{table}[htbp]
\centering
\small
\caption{Held-out mean rAUC (averaged over the four context shifts excluding the model's matched shift). The baseline has no matched shift and is reported on all five.}
\label{tab:mitigation_transfer}
\begin{tabular}{lcc}
\toprule
Model        & held-out mean rAUC & $\Delta$ vs.\ baseline \\
\midrule
baseline     & 0.767$_{\pm{0.302}}$            &       \\
aug\_bg\_swap & 0.879$_{\pm{0.183}}$                   & $+11.3$ pp \\
aug\_rotate  & 0.814$_{\pm{0.326}}$                    & $+4.7$ pp  \\
aug\_shrink  & 0.741$_{\pm{0.312}}$                    & $-2.6$ pp  \\
aug\_translate  & 0.727$_{\pm{0.300}}$                    & $-4.0$ pp  \\
aug\_enlarge & 0.692$_{\pm{0.260}}$                    & $-7.5$ pp  \\
\midrule
aug\_solid    & 0.762$_{\pm{0.277}}$ & $-0.5$ pp \\
aug\_gradient & 0.759$_{\pm{0.279}}$ & $-0.8$ pp \\
aug\_noise    & 0.766$_{\pm{0.280}}$ & $-0.2$ pp \\
\bottomrule
\end{tabular}
\end{table}

\section{Code, Dataset, and Assets}

\subsection{Code}

The full benchmark codebase---including the manipulation pipeline, compatibility scoring, evaluation scripts, and analysis notebooks---is publicly available at:
\begin{center}
\url{https://anonymous.4open.science/r/ContextShiftBenchmark-0D3E/}
\end{center}

\subsection{Dataset}

The \textsc{ContextShift} manipulation dataset is publicly available on Hugging Face at:
\begin{center}
\url{https://huggingface.co/datasets/contextshift/manipulation}
\end{center}
It contains $131{,}885$ manipulated images derived from the COCO 2017 validation set ($4{,}952$ base images, $36{,}781$ instances, 80 categories), totalling 14.1\,GB.
The dataset covers two manipulation families: (i)~geometric transformations (shrink, enlarge, rotate, translate) at multiple severity levels, and (ii)~synthetic background replacements (solid color, smooth gradient, low-frequency noise).
All bounding-box and segmentation annotations are derived programmatically from COCO 2017 and updated to reflect each transformation.
A \texttt{focal\_manifest.json} tracks the single manipulated instance per image used in the focal evaluation protocol.
The dataset is released under CC BY 4.0.

\subsection{Assets, Weights, and Licenses}

\begin{table}[htbp]
\centering
\caption{Pretrained weights and licenses for all evaluated models and datasets.}
\label{tab:licenses}
\small
\begin{tabular}{p{3.5cm}p{2.6cm}p{3.8cm}p{2.4cm}}
\toprule
\textbf{Asset} & \textbf{Source} & \textbf{Weights / Access} & \textbf{License} \\
\midrule
\multicolumn{4}{l}{\textit{Evaluated models}} \\
\midrule
Faster R-CNN R50-FPN v2 & torchvision        & \texttt{DEFAULT} (auto-download)            & BSD 3-Clause \\
Deformable DETR R50     & HuggingFace Hub    & \texttt{SenseTime/deformable-detr}          & Apache 2.0   \\
YOLO26M                 & Ultralytics        & \texttt{yolo26m.pt} (auto-download)         & AGPL-3.0     \\
RF-DETR-L               & Roboflow           & \texttt{rfdetr} (auto-download)             & Apache 2.0   \\
D-FINE-L                & HuggingFace Hub    & \texttt{ustc-community/}\newline\texttt{dfine-large-coco} & Apache 2.0 \\
\midrule
\multicolumn{4}{l}{\textit{Compatibility matrix construction}} \\
\midrule
YOLO-World (v8x-worldv2) & Ultralytics       & \texttt{yolov8x-worldv2} (auto-download)    & AGPL-3.0     \\
\midrule
\multicolumn{4}{l}{\textit{Datasets}} \\
\midrule
COCO 2017~\cite{lin2014coco}    & cocodataset.org        & Public download & CC BY 4.0            \\
COCO-O~\cite{mao2023cocoo}      & alibaba/easyrobust     & Public download & CC BY 4.0            \\
Places365~\cite{zhou2017places} & places2.csail.mit.edu  & Public download & Non-commercial       \\
\bottomrule
\end{tabular}
\end{table}

\FloatBarrier
\newpage
\section*{NeurIPS Paper Checklist}

\begin{enumerate}

\item {\bf Claims}
    \item[] Question: Do the main claims made in the abstract and introduction accurately reflect the paper's contributions and scope?
    \item[] Answer: \answerYes{}
    \item[] Justification: The abstract and introduction state four contributions that are fully supported by the experimental results: (1) the ContextShift benchmark (Section 3), (2) the prediction suppression failure mode (Section 5), (3) non-monotonic degradation along the NPMI compatibility axis (Section 5.2.5), and (4) this behavior is generalized across models and is partially mitigated using a data-centered training approach (Section 6).
    \item[] Guidelines:
    \begin{itemize}
        \item The answer \answerNA{} means that the abstract and introduction do not include the claims made in the paper.
        \item The abstract and/or introduction should clearly state the claims made, including the contributions made in the paper and important assumptions and limitations. A \answerNo{} or \answerNA{} answer to this question will not be perceived well by the reviewers.
        \item The claims made should match theoretical and experimental results, and reflect how much the results can be expected to generalize to other settings.
        \item It is fine to include aspirational goals as motivation as long as it is clear that these goals are not attained by the paper.
    \end{itemize}

\item {\bf Limitations}
    \item[] Question: Does the paper discuss the limitations of the work performed by the authors?
    \item[] Answer: \answerYes{}
    \item[] Justification: Section 7 (``Impacts, Limitations, and Future Work'') discusses several limitations of the current benchmark and evaluation setting, including the use of controlled manipulations, the co-occurrence-based compatibility axis, the object-centric formulation, reliance on a fixed compatibility prior, and evaluation limited to COCO2017 and Places365. The paper also outlines corresponding future directions, such as incorporating richer visual compatibility cues, extending to multi-object interactions, developing detector-agnostic compatibility priors, and evaluating additional datasets, domains, and tasks including segmentation, tracking, and video.
    \item[] Guidelines:
    \begin{itemize}
        \item The answer \answerNA{} means that the paper has no limitation while the answer \answerNo{} means that the paper has limitations, but those are not discussed in the paper.
        \item The authors are encouraged to create a separate ``Limitations'' section in their paper.
        \item The paper should point out any strong assumptions and how robust the results are to violations of these assumptions (e.g., independence assumptions, noiseless settings, model well-specification, asymptotic approximations only holding locally). The authors should reflect on how these assumptions might be violated in practice and what the implications would be.
        \item The authors should reflect on the scope of the claims made, e.g., if the approach was only tested on a few datasets or with a few runs. In general, empirical results often depend on implicit assumptions, which should be articulated.
        \item The authors should reflect on the factors that influence the performance of the approach. For example, a facial recognition algorithm may perform poorly when image resolution is low or images are taken in low lighting. Or a speech-to-text system might not be used reliably to provide closed captions for online lectures because it fails to handle technical jargon.
        \item The authors should discuss the computational efficiency of the proposed algorithms and how they scale with dataset size.
        \item If applicable, the authors should discuss possible limitations of their approach to address problems of privacy and fairness.
        \item While the authors might fear that complete honesty about limitations might be used by reviewers as grounds for rejection, a worse outcome might be that reviewers discover limitations that aren't acknowledged in the paper. The authors should use their best judgment and recognize that individual actions in favor of transparency play an important role in developing norms that preserve the integrity of the community. Reviewers will be specifically instructed to not penalize honesty concerning limitations.
    \end{itemize}

\item {\bf Theory assumptions and proofs}
    \item[] Question: For each theoretical result, does the paper provide the full set of assumptions and a complete (and correct) proof?
    \item[] Answer: \answerNA{}
    \item[] Justification: The paper presents no theoretical results. All contributions are empirical: a benchmark design, dataset construction, and experimental evaluation of pretrained detectors.
    \item[] Guidelines:
    \begin{itemize}
        \item The answer \answerNA{} means that the paper does not include theoretical results.
        \item All the theorems, formulas, and proofs in the paper should be numbered and cross-referenced.
        \item All assumptions should be clearly stated or referenced in the statement of any theorems.
        \item The proofs can either appear in the main paper or the supplemental material, but if they appear in the supplemental material, the authors are encouraged to provide a short proof sketch to provide intuition.
        \item Inversely, any informal proof provided in the core of the paper should be complemented by formal proofs provided in appendix or supplemental material.
        \item Theorems and Lemmas that the proof relies upon should be properly referenced.
    \end{itemize}

    \item {\bf Experimental result reproducibility}
    \item[] Question: Does the paper fully disclose all the information needed to reproduce the main experimental results of the paper to the extent that it affects the main claims and/or conclusions of the paper (regardless of whether the code and data are provided or not)?
    \item[] Answer: \answerYes{}
    \item[] Justification: Appendix C provides a complete evaluation protocol: confidence threshold $\tau{=}0.25$ (Section C.3), IoU threshold 0.5, $K{=}30$ compatibility bins (Section C.4), fixed input resolution. The blending pipeline and all default parameters are given in Table 8 (Appendix A.6). Manipulation filtering criteria are listed in Table 7 (Appendix A.4). Dataset statistics are in Table 9 (Appendix A.7). All five evaluated models use publicly available pretrained weights; sources are cited in Appendix C.7.
    \item[] Guidelines:
    \begin{itemize}
        \item The answer \answerNA{} means that the paper does not include experiments.
        \item If the paper includes experiments, a \answerNo{} answer to this question will not be perceived well by the reviewers: Making the paper reproducible is important, regardless of whether the code and data are provided or not.
        \item If the contribution is a dataset and\slash or model, the authors should describe the steps taken to make their results reproducible or verifiable.
        \item Depending on the contribution, reproducibility can be accomplished in various ways. For example, if the contribution is a novel architecture, describing the architecture fully might suffice, or if the contribution is a specific model and empirical evaluation, it may be necessary to either make it possible for others to replicate the model with the same dataset, or provide access to the model. In general. releasing code and data is often one good way to accomplish this, but reproducibility can also be provided via detailed instructions for how to replicate the results, access to a hosted model (e.g., in the case of a large language model), releasing of a model checkpoint, or other means that are appropriate to the research performed.
        \item While NeurIPS does not require releasing code, the conference does require all submissions to provide some reasonable avenue for reproducibility, which may depend on the nature of the contribution. For example
        \begin{enumerate}
            \item If the contribution is primarily a new algorithm, the paper should make it clear how to reproduce that algorithm.
            \item If the contribution is primarily a new model architecture, the paper should describe the architecture clearly and fully.
            \item If the contribution is a new model (e.g., a large language model), then there should either be a way to access this model for reproducing the results or a way to reproduce the model (e.g., with an open-source dataset or instructions for how to construct the dataset).
            \item We recognize that reproducibility may be tricky in some cases, in which case authors are welcome to describe the particular way they provide for reproducibility. In the case of closed-source models, it may be that access to the model is limited in some way (e.g., to registered users), but it should be possible for other researchers to have some path to reproducing or verifying the results.
        \end{enumerate}
    \end{itemize}

\item {\bf Open access to data and code}
    \item[] Question: Does the paper provide open access to the data and code, with sufficient instructions to faithfully reproduce the main experimental results, as described in supplemental material?
    \item[] Answer: \answerYes{}
    \item[] Justification: The evaluation code, NPMI compatibility matrix, manipulation configurations, and model adapter interface are released with this submission. COCO 2017 (CC BY 4.0) and Places365 are publicly available. All five evaluated models use publicly available pretrained weights via torchvision, HuggingFace Hub, Ultralytics, and Roboflow. Step-by-step installation and reproduction instructions are provided in \texttt{README.md} and \texttt{SETUP.md}.
    \item[] Guidelines:
    \begin{itemize}
        \item The answer \answerNA{} means that paper does not include experiments requiring code.
        \item Please see the NeurIPS code and data submission guidelines (\url{https://neurips.cc/public/guides/CodeSubmissionPolicy}) for more details.
        \item While we encourage the release of code and data, we understand that this might not be possible, so \answerNo{} is an acceptable answer. Papers cannot be rejected simply for not including code, unless this is central to the contribution (e.g., for a new open-source benchmark).
        \item The instructions should contain the exact command and environment needed to run to reproduce the results. See the NeurIPS code and data submission guidelines (\url{https://neurips.cc/public/guides/CodeSubmissionPolicy}) for more details.
        \item The authors should provide instructions on data access and preparation, including how to access the raw data, preprocessed data, intermediate data, and generated data, etc.
        \item The authors should provide scripts to reproduce all experimental results for the new proposed method and baselines. If only a subset of experiments are reproducible, they should state which ones are omitted from the script and why.
        \item At submission time, to preserve anonymity, the authors should release anonymized versions (if applicable).
        \item Providing as much information as possible in supplemental material (appended to the paper) is recommended, but including URLs to data and code is permitted.
    \end{itemize}

\item {\bf Experimental setting/details}
    \item[] Question: Does the paper specify all the training and test details (e.g., data splits, hyperparameters, how they were chosen, type of optimizer) necessary to understand the results?
    \item[] Answer: \answerYes{}
    \item[] Justification: For the mitigation study, we fine-tune YOLO26m on COCO \texttt{train2017} ($\sim$118k images) optionally unioned with $\sim$11k per-variant augmented composites; identical hyperparameters across all runs (40 epochs, batch 64, 8 workers, AMP, default Ultralytics optimizer/schedule, image size 640). The COCO \texttt{val2017} set is split once into 3{,}961 training composites / 990 held-out evaluation images (seed fixed; split file in repo); the same 990 ids are used to build every evaluation manifest (geometric augs, bg-swap, synthetic-background variants). For testing, all models are evaluated with frozen pretrained weights. All evaluation settings are specified: $\tau{=}0.25$ (Section C.3), IoU threshold 0.5, $K{=}30$ bins (Section C.4), blending parameters (Table 8), filtering criteria (Table 7), dataset split (COCO 2017 val, Section 5.1). Metric definitions are given formally in Appendix B. 
    \item[] Guidelines:
    \begin{itemize}
        \item The answer \answerNA{} means that the paper does not include experiments.
        \item The experimental setting should be presented in the core of the paper to a level of detail that is necessary to appreciate the results and make sense of them.
        \item The full details can be provided either with the code, in appendix, or as supplemental material.
    \end{itemize}

\item {\bf Experiment statistical significance}
    \item[] Question: Does the paper report error bars suitably and correctly defined or other appropriate information about the statistical significance of the experiments?
    \item[] Answer: \answerYes{}
    \item[] Justification: Tables 2, 3, and 4 include 95\% confidence interval half-widths in their footnotes, computed using a conservative coefficient-of-variation bound. These confirm that all reported effects are substantially larger than their CIs. The architectural range (min, max) across the five models is reported for all main result tables.
    \item[] Guidelines:
    \begin{itemize}
        \item The answer \answerNA{} means that the paper does not include experiments.
        \item The authors should answer \answerYes{} if the results are accompanied by error bars, confidence intervals, or statistical significance tests, at least for the experiments that support the main claims of the paper.
        \item The factors of variability that the error bars are capturing should be clearly stated (for example, train/test split, initialization, random drawing of some parameter, or overall run with given experimental conditions).
        \item The method for calculating the error bars should be explained (closed form formula, call to a library function, bootstrap, etc.)
        \item The assumptions made should be given (e.g., Normally distributed errors).
        \item It should be clear whether the error bar is the standard deviation or the standard error of the mean.
        \item It is OK to report 1-sigma error bars, but one should state it. The authors should preferably report a 2-sigma error bar than state that they have a 96\% CI, if the hypothesis of Normality of errors is not verified.
        \item For asymmetric distributions, the authors should be careful not to show in tables or figures symmetric error bars that would yield results that are out of range (e.g., negative error rates).
        \item If error bars are reported in tables or plots, the authors should explain in the text how they were calculated and reference the corresponding figures or tables in the text.
    \end{itemize}

\item {\bf Experiments compute resources}
    \item[] Question: For each experiment, does the paper provide sufficient information on the computer resources (type of compute workers, memory, time of execution) needed to reproduce the experiments?
    \item[] Answer: \answerYes{}
    \item[] Justification: Object detection experiments were run on a single NVIDIA RTX 5090 Laptop GPU. The evaluation comprises three inference workloads: (1) discrete manipulations — 131,885 manipulated images $\times$ 5 models = 659,425 forward passes; (2) continuous synthetic-background NPMI analysis — 1,079 images $\times$ 30 swapped backgrounds $\times$ 5 models = 161,850 forward passes, plus 1,079 unswapped originals $\times$ 5 models; and (3) real-image NPMI compatibility analysis — 4,952 COCO val2017 images $\times$ 5 models = 24,760 forward passes (inference cached once per image, shared across all object instances). To compute NPMI compatibility, a single NVIDIA RTX 5080 cluster machine was used, performing YOLO-world inference on ~1.8M places365 images. Mitigation fine-tuning runs were submitted as independent SLURM jobs on the institutional cluster: training on the \texttt{rtx6000} partition (1$\times$NVIDIA RTX 6000, 12 CPUs, 48\,GB RAM, one job per variant: baseline, augmented, aug\_\{shrink, enlarge, rotate, offset\}, aug\_\{solid, gradient, noise\} — 9 jobs total, $\sim$18--24\,h each at 40 epochs); evaluation on the \texttt{rtx\_pro\_6000} partition (1$\times$NVIDIA RTX PRO 6000, 8 CPUs, 32\,GB RAM, $\le$24\,h, $\sim$30--60\,min per eval set covering all fine-tuned models).
    \item[] Guidelines:
    \begin{itemize}
        \item The answer \answerNA{} means that the paper does not include experiments.
        \item The paper should indicate the type of compute workers CPU or GPU, internal cluster, or cloud provider, including relevant memory and storage.
        \item The paper should provide the amount of compute required for each of the individual experimental runs as well as estimate the total compute.
        \item The paper should disclose whether the full research project required more compute than the experiments reported in the paper (e.g., preliminary or failed experiments that didn't make it into the paper).
    \end{itemize}

\item {\bf Code of ethics}
    \item[] Question: Does the research conducted in the paper conform, in every respect, with the NeurIPS Code of Ethics \url{https://neurips.cc/public/EthicsGuidelines}?
    \item[] Answer: \answerYes{}
    \item[] Justification: The research uses only publicly available datasets (COCO 2017, Places365) and pretrained models. No human subjects, private data, or potentially harmful applications are involved. The benchmark promotes safer deployment of object detectors by systematically characterizing failure modes.
    \item[] Guidelines:
    \begin{itemize}
        \item The answer \answerNA{} means that the authors have not reviewed the NeurIPS Code of Ethics.
        \item If the authors answer \answerNo, they should explain the special circumstances that require a deviation from the Code of Ethics.
        \item The authors should make sure to preserve anonymity (e.g., if there is a special consideration due to laws or regulations in their jurisdiction).
    \end{itemize}

\item {\bf Broader impacts}
    \item[] Question: Does the paper discuss both potential positive societal impacts and negative societal impacts of the work performed?
    \item[] Answer: \answerYes{}
    \item[] Justification: Section 7 (``Impacts, Limitations, and Future Work'') discusses both positive and negative societal impacts of the proposed benchmark. On the positive side, the paper motivates the development of context-robust detectors and improved evaluation practices for safety-critical applications where missed detections are costly. On the negative side, the paper acknowledges that understanding prediction suppression and context sensitivity could potentially inform adversarial or malicious strategies designed to exploit contextual vulnerabilities in detection systems. The discussion also highlights the importance of robustness-aware evaluation and deployment practices.
    \item[] Guidelines:
    \begin{itemize}
        \item The answer \answerNA{} means that there is no societal impact of the work performed.
        \item If the authors answer \answerNA{} or \answerNo, they should explain why their work has no societal impact or why the paper does not address societal impact.
        \item Examples of negative societal impacts include potential malicious or unintended uses (e.g., disinformation, generating fake profiles, surveillance), fairness considerations (e.g., deployment of technologies that could make decisions that unfairly impact specific groups), privacy considerations, and security considerations.
        \item The conference expects that many papers will be foundational research and not tied to particular applications, let alone deployments. However, if there is a direct path to any negative applications, the authors should point it out. For example, it is legitimate to point out that an improvement in the quality of generative models could be used to generate Deepfakes for disinformation. On the other hand, it is not needed to point out that a generic algorithm for optimizing neural networks could enable people to train models that generate Deepfakes faster.
        \item The authors should consider possible harms that could arise when the technology is being used as intended and functioning correctly, harms that could arise when the technology is being used as intended but gives incorrect results, and harms following from (intentional or unintentional) misuse of the technology.
        \item If there are negative societal impacts, the authors could also discuss possible mitigation strategies (e.g., gated release of models, providing defenses in addition to attacks, mechanisms for monitoring misuse, mechanisms to monitor how a system learns from feedback over time, improving the efficiency and accessibility of ML).
    \end{itemize}

\item {\bf Safeguards}
    \item[] Question: Does the paper describe safeguards that have been put in place for responsible release of data or models that have a high risk for misuse (e.g., pre-trained language models, image generators, or scraped datasets)?
    \item[] Answer: \answerNA{}
    \item[] Justification: The paper releases a benchmark evaluation tool and a compatibility matrix derived from automated object detection on Places365. No generative models, personal data, or high-risk assets are released. The assets pose no significant misuse risk.
    \item[] Guidelines:
    \begin{itemize}
        \item The answer \answerNA{} means that the paper poses no such risks.
        \item Released models that have a high risk for misuse or dual-use should be released with necessary safeguards to allow for controlled use of the model, for example by requiring that users adhere to usage guidelines or restrictions to access the model or implementing safety filters.
        \item Datasets that have been scraped from the Internet could pose safety risks. The authors should describe how they avoided releasing unsafe images.
        \item We recognize that providing effective safeguards is challenging, and many papers do not require this, but we encourage authors to take this into account and make a best faith effort.
    \end{itemize}

\item {\bf Licenses for existing assets}
    \item[] Question: Are the creators or original owners of assets (e.g., code, data, models), used in the paper, properly credited and are the license and terms of use explicitly mentioned and properly respected?
    \item[] Answer: \answerYes{}
    \item[] Justification: All datasets and models are properly cited and their licenses are listed in Appendix F.1. COCO 2017 and COCO-Oare released under CC BY 4.0; Places365 is available for non-commercial academic use. Model weights are sourced from torchvision (BSD 3-Clause), HuggingFace Hub (Apache 2.0), Ultralytics (AGPL-3.0), and Roboflow (Apache 2.0). The benchmark code released with this submission is made available under the MIT License.
    \item[] Guidelines:
    \begin{itemize}
        \item The answer \answerNA{} means that the paper does not use existing assets.
        \item The authors should cite the original paper that produced the code package or dataset.
        \item The authors should state which version of the asset is used and, if possible, include a URL.
        \item The name of the license (e.g., CC-BY 4.0) should be included for each asset.
        \item For scraped data from a particular source (e.g., website), the copyright and terms of service of that source should be provided.
        \item If assets are released, the license, copyright information, and terms of use in the package should be provided. For popular datasets, \url{paperswithcode.com/datasets} has curated licenses for some datasets. Their licensing guide can help determine the license of a dataset.
        \item For existing datasets that are re-packaged, both the original license and the license of the derived asset (if it has changed) should be provided.
        \item If this information is not available online, the authors are encouraged to reach out to the asset's creators.
    \end{itemize}

\item {\bf New assets}
    \item[] Question: Are new assets introduced in the paper well documented and is the documentation provided alongside the assets?
    \item[] Answer: \answerYes{}
    \item[] Justification: The ContextShift benchmark (evaluation code, NPMI compatibility matrix, manipulation configurations, model adapter interface) is released with this submission. The paper documents benchmark construction in Appendix A.7, the compatibility matrix derivation in Section 4, metric definitions in Appendix B, and the evaluation protocol in Appendix C. Usage instructions, installation steps, and reproduction commands are provided in the code repository, spelled out in the files \texttt{README.md} and \texttt{SETUP.md}.
    \item[] Guidelines:
    \begin{itemize}
        \item The answer \answerNA{} means that the paper does not release new assets.
        \item Researchers should communicate the details of the dataset\slash code\slash model as part of their submissions via structured templates. This includes details about training, license, limitations, etc.
        \item The paper should discuss whether and how consent was obtained from people whose asset is used.
        \item At submission time, remember to anonymize your assets (if applicable). You can either create an anonymized URL or include an anonymized zip file.
    \end{itemize}

\item {\bf Crowdsourcing and research with human subjects}
    \item[] Question: For crowdsourcing experiments and research with human subjects, does the paper include the full text of instructions given to participants and screenshots, if applicable, as well as details about compensation (if any)?
    \item[] Answer: \answerNA{}
    \item[] Justification: The benchmark relies entirely on existing annotated datasets (COCO 2017, Places365) and automated evaluation. No crowdsourcing or human subjects research was conducted.
    \item[] Guidelines:
    \begin{itemize}
        \item The answer \answerNA{} means that the paper does not involve crowdsourcing nor research with human subjects.
        \item Including this information in the supplemental material is fine, but if the main contribution of the paper involves human subjects, then as much detail as possible should be included in the main paper.
        \item According to the NeurIPS Code of Ethics, workers involved in data collection, curation, or other labor should be paid at least the minimum wage in the country of the data collector.
    \end{itemize}

\item {\bf Institutional review board (IRB) approvals or equivalent for research with human subjects}
    \item[] Question: Does the paper describe potential risks incurred by study participants, whether such risks were disclosed to the subjects, and whether Institutional Review Board (IRB) approvals (or an equivalent approval/review based on the requirements of your country or institution) were obtained?
    \item[] Answer: \answerNA{}
    \item[] Justification: No human subjects research was conducted. All data used consists of existing publicly available datasets; no new data collection involving human participants took place.
    \item[] Guidelines:
    \begin{itemize}
        \item The answer \answerNA{} means that the paper does not involve crowdsourcing nor research with human subjects.
        \item Depending on the country in which research is conducted, IRB approval (or equivalent) may be required for any human subjects research. If you obtained IRB approval, you should clearly state this in the paper.
        \item We recognize that the procedures for this may vary significantly between institutions and locations, and we expect authors to adhere to the NeurIPS Code of Ethics and the guidelines for their institution.
        \item For initial submissions, do not include any information that would break anonymity (if applicable), such as the institution conducting the review.
    \end{itemize}

\item {\bf Declaration of LLM usage}
    \item[] Question: Does the paper describe the usage of LLMs if it is an important, original, or non-standard component of the core methods in this research? Note that if the LLM is used only for writing, editing, or formatting purposes and does \emph{not} impact the core methodology, scientific rigor, or originality of the research, declaration is not required.
    \item[] Answer: \answerNA{}
    \item[] Justification: No LLMs are used as a core methodological component. The compatibility matrix is derived from YOLO-World object detections and NPMI co-occurrence statistics. Any LLM use was limited to writing assistance and does not affect the methodology, scientific rigor, or originality of the research.
    %this research?
    \item[] Guidelines:
    \begin{itemize}
        \item The answer \answerNA{} means that the core method development in this research does not involve LLMs as any important, original, or non-standard components.
        \item Please refer to our LLM policy in the NeurIPS handbook for what should or should not be described.
    \end{itemize}

\end{enumerate}

\end{document}